\documentclass[10pt,journal,compsoc]{IEEEtran}
\ifCLASSOPTIONcompsoc
  \usepackage[nocompress]{cite}
\else
  \usepackage{cite}
\fi
\ifCLASSINFOpdf
\else
\fi

\usepackage{ifpdf}

\usepackage[pdftex]{graphicx}
\usepackage{epsfig} 
\usepackage{amsthm}
\usepackage{amsmath}
\usepackage{amsfonts}
\usepackage{amssymb}
\usepackage{bm}
\usepackage{cases}
\usepackage{multirow}

\usepackage{multicol}
\usepackage{float,subfig}
\usepackage{caption}
\usepackage{cite}
\usepackage{url}
\usepackage{here}
\usepackage{stfloats} 
\usepackage{braket}
\usepackage[linesnumbered,ruled,vlined]{algorithm2e}
\usepackage{algorithmic}

\usepackage{multicol}
\usepackage[export]{adjustbox}

\usepackage{array}
\newcolumntype{C}[1]{>{\hfil}m{#1}<{\hfil}}

\usepackage{lscape} 

\begin{document}
%
\title{Multi-label Classification via Adaptive Resonance Theory-based Clustering}
%
%
%
%

\author{
	Naoki~Masuyama,~\IEEEmembership{Member,~IEEE,}
	Yusuke~Nojima,~\IEEEmembership{Member,~IEEE,}
    Chu~Kiong~Loo,~\IEEEmembership{Senior Member,~IEEE,}
    and~Hisao~Ishibuchi,~\IEEEmembership{Fellow,~IEEE}
\IEEEcompsocitemizethanks{\IEEEcompsocthanksitem N. Masuyama, and Y. Nojima are with the Graduate School of Engineering, Osaka Prefecture University, 1-1 Gakuen-cho Naka-ku, Sakai-Shi, Osaka 599-8531, Japan.\protect\\
E-mails: \{masuyama, nojima\}@cs.osakafu-u.ac.jp
\IEEEcompsocthanksitem C. K. Loo is with the Faculty of Computer Science and Information Technology, University of Malaya, 50603 Kuala Lumpur, Malaysia.\protect\\
E-mail: ckloo.um@um.edu.my
\IEEEcompsocthanksitem H. Ishibuchi is with the Guangdong Provincial Key Laboratory of Brain-inspired Intelligent Computation, Department of Computer Science and Engineering, Southern University of Science and Technology, Shenzhen 518055, China.\protect\\
E-mail: hisao@sustech.edu.cn
}
\thanks{Corresponding author: Hisao Ishibuchi (e-mail: hisao@sustech.edu.cn).}
\thanks{Manuscript received April 19, 2005; revised August 26, 2015.}}

%
%

\markboth{Journal of \LaTeX\ Class Files,~Vol.~14, No.~8, August~2015}%
{Shell \MakeLowercase{\textit{et al.}}: Bare Demo of IEEEtran.cls for Computer Society Journals}
%



\IEEEtitleabstractindextext{%
\begin{abstract}
This paper proposes a multi-label classification algorithm capable of continual learning by applying an Adaptive Resonance Theory (ART)-based clustering algorithm and the Bayesian approach for label probability computation. The ART-based clustering algorithm adaptively and continually generates prototype nodes corresponding to given data, and the generated nodes are used as classifiers. The label probability computation independently counts the number of label appearances for each class and calculates the Bayesian probabilities. Thus, the label probability computation can cope with an increase in the number of labels. Experimental results with synthetic and real-world multi-label datasets show that the proposed algorithm has competitive classification performance to other well-known algorithms while realizing continual learning.
\end{abstract}

}

\maketitle

\IEEEdisplaynontitleabstractindextext

%
\IEEEpeerreviewmaketitle

\IEEEraisesectionheading{\section{Introduction}\label{sec:introduction}}

%
%
%
%
\IEEEPARstart{T}{hanks} to the recent advances of IoT technology, we can easily obtain a wide variety of data and utilize them to machine learning algorithms. Thus, the importance of continual learning is increasing for machine learning algorithms in order to efficiently utilize the data \cite{parisi19}. The requirement for continual learning is to handle both sequential learning and class-incremental learning without destroying the learned knowledge. In general, sequential learning is defined as a method that learns the data instance by instance, not in batches. Class-incremental learning is defined as a method that can deal with the situation where the number of classes (labels) increases during the learning process.

Since real-world phenomena and objects are complex and may have multiple semantics in nature, multi-label classification attracts a great deal of attention from machine learning and related fields such as web mining \cite{zhang18}, rule mining \cite{han19}, and information retrieval \cite{he19, zhang21}. In regard to multi-label classification algorithms, the sequential learning has realized by stream multi-label classification algorithms \cite{zheng19}. For those algorithms, however, data pre-processing such as normalization and standardization is often required. In addition, it is necessary for learning process to define the number of classes in advance. The class-incremental learning is theoretically feasible with the Bayesian approach for label probability computation in Multi-Label $ k $-Nearest Neighbor (ML-$ k $NN) \cite{zhang07}. The label learning process independently counts the number of label appearances for each class and calculates the Bayesian probabilities. Thus, the label probability computation can cope with an increase in the number of labels. However, ML-$ k $NN cannot cope with the sequential learning because $ k $-NN requires the entire data before the learning process. Multi-label Learning with Emerging New Labels (MuENL) \cite{zhu18a} has realized continual learning by constructing two classifiers for classifying instances and for detecting new labels. However, MuENL cannot perform continual learning in the non-stationary environment, i.e., the situation where new data distributions are sequentially provided. 

In the case of single-label classification algorithms, several types of clustering-based classifiers capable of continual learning have been proposed \cite{furao06, shen08, masuyama18, masuyama19a, masuyama19b, masuyamaFTCA}. In particular, classifiers designed by an Adaptive Resonance Theory (ART)-based clustering algorithm have shown comparable classification performance to typical classification algorithms such as SVM and $ k $-NN. The main feature of the above ART-based clustering algorithms is the use of the Correntropy-Induced Metric (CIM) \cite{liu07} to a similarity threshold, which makes the self-organizing process fast and stable \cite{masuyama18, masuyama19a, masuyama19b, masuyamaFTCA}.

In this paper, in order to realize a multi-label classification algorithm capable of continual learning, we propose Multi-Label CIM-based ART (MLCA) by integrating the Bayesian approach for label probability computation into the ART-based clustering with the CIM. Furthermore, we also propose two variants of MLCA by modifying the calculation method of the CIM to improve the classification performance of MLCA.

The contributions of this paper are summarized as follows:
\begin{itemize}
	\vspace{-1mm}
	\item[(i)] A multi-label classification algorithm, called MLCA, is proposed by integrating a CIM-based ART and the Bayesian approach for label probability computation. MLCA computes the prior probability and likelihood by using nodes (representative points of training data) which have accumulated label counts while ML-$ k $NN computes them by directly referencing to training instances.
	
	\item[(ii)] A new CIM-based ART is proposed by introducing an efficient node generation process and a bandwidth adaptation method for a kernel function in CIM. As a result, MLCA does not require any data pre-processing method such as normalization or scaling.
	
	\item[(iii)] Empirical studies show that MLCA and its variants have competitive classification performance to recent multi-label classification algorithms.
	
	\item[(iv)] The continual learning ability of MLCA is analyzed from multiple perspectives, and its usefulness and superiority are clarified.
\end{itemize}

The paper is organized as follows. Section \ref{sec:literaturereview} presents literature review for clustering algorithms and multi-label classification algorithms. Section \ref{sec:mlca} describes details of the proposed algorithm and modifications of the calculation method of the CIM. Section \ref{sec:experiment} presents extensive simulation experiments to evaluate the continual learning ability and classification performance of MLCA. Section \ref{sec:conclusion} concludes this paper.

\section{Literature Review}
\label{sec:literaturereview}

\subsection{Clustering Algorithm}
\label{sec:LR_clustering}
Cluster analysis is one of the widely applied approaches to extract hidden relation from data. Typical types of clustering algorithms are the Gaussian mixture model \cite{mclachlan19}, $ k $-means \cite{lloyd82}, and Self-Organizing Map (SOM) \cite{kohonen82}. Although the above algorithms are quite simple and highly adaptable, the number of classes and network architectures are specified in advance. Growing Neural Gas (GNG) \cite{fritzke95} and Self-Organizing Incremental Neural Network (SOINN) \cite{furao06} are well-known growing self-organizing clustering algorithms that can overcome the drawbacks of the typical types of clustering algorithms. GNG and SOINN can adaptively generate topological networks corresponding to given data. However, since these algorithms permanently insert new nodes into their networks for memorizing new knowledge, they have a potential to forget learned knowledge (i.e., catastrophic forgetting). This trade-off is called the plasticity-stability dilemma \cite{carpenter88}. A variant of GNG, called Grow When Required (GWR) \cite{marsland02} can avoid the plasticity-stability dilemma by adding nodes whenever the state of the current network does not sufficiently match the instance. One problem of GWR is that as the number of nodes in the network increases, the cost of calculating a threshold for each node increases, and thus the learning efficiency decreases.

A successful approach to avoid the plasticity-stability dilemma is the ART-based algorithms \cite{grossberg87}. Because the ART-based algorithms realize sequential and class-incremental learning without the catastrophic forgetting, a number of the ART-based algorithms and their improvements are proposed in both supervised learning \cite{tan14, matias18, matias21} and unsupervised learning \cite{carpenter91b, vigdor07, wang19, da20}. In the ART-based algorithms, a criterion of a new category (node) generation, i.e., a similarity measurement between a node and an instance, has a great impact on the classification/clustering performance. Previous studies have shown that algorithms with the CIM \cite{liu07} as a similarity measurement are capable of faster and more stable learning than other self-organizing clustering algorithms \cite{masuyama18, masuyama19a, masuyama19b, masuyamaFTCA}.

\subsection{Multi-label Classification}
\label{sec:LR_multilabel}

The multi-label classification algorithms are categorized into two approaches, namely, a problem transformation approach and an algorithm adaptation approach \cite{zhang13}. The problem transformation approach transforms a multi-label classification problem into multiple single-label classification problems. The problem transformation approach is further divided into two methods, namely, the Binary Relevance (BR) \cite{boutell04} and the Label Powerset (LP) \cite{tsoumakas10}. The BR transforms a multi-label classification problem into multiple binary classification problems by decomposing multi-labels into multiple single labels. The LP transforms a multi-label classification problem into a multi-class classification problem by merging multi-labels into a single label. Various single-label classification algorithms have been used in the problem transformation approach thanks to its simplicity and applicability. 

The algorithm adaptation approach extends existing single-label classification algorithms for handling multi-label classification problems. Various types of algorithm adaptation approach have been introduced based on $ k $-NN \cite{zhang07, liu16}, decision tree \cite{clare01}, regression \cite{osojnik17}, Support Vector Machine (SVM) \cite{elisseeff01}, and feed-forward neural networks \cite{zhang16_multi, zhang20}. In order to achieve high classification performance, recent studies consider label distributions and their correlations in a label learning process \cite{zhu18, tan20}. These algorithms, however, cannot cope with the situation where new label information is sequentially provided. ML-$ k $NN \cite{zhang07} is a well-known algorithm adaptation method that integrates $ k $-NN and the Bayesian approach for label probability computation. ML-$ k $NN counts the number of relevant labels of neighbors for each instance in training data. Based on the counts of relevant labels, the likelihood and posterior probability of each label are computed by the Bayesian approach. Here, the computation of the Bayesian probability is individually performed in each label. Therefore, in theory, the number of labels to be learned can be increased/decreased during the label probability computation. One disadvantage of ML-$ k $NN is that it cannot efficiently cope with the situation where new training instances are sequentially provided. Multi-Label Self-Adjusting $ k $ Nearest Neighbors (MLSA-$ k $NN) \cite{roseberry21} is capable of handling a stream multi-label classification by employing a self-adjusting window to detect a concept drift and to adaptively control a parameter $ k $ in $ k $NN. MuENL \cite{zhu18a} generates two classifiers for classifying instances and for detecting new labels. By using the two classifiers, MuENL realizes sequential learning and class-incremental learning, simultaneously. Although MuENL is capable of continual learning, MuENL requires a batch learning process in the initial stage of learning for constructing two classifiers for instances and labels. Moreover, MuENL cannot deal with the situation where new data distributions are provided in the non-stationary environment.

\begin{table*}[bp]
	\centering
	\caption{Summary of notations}
	\renewcommand{\arraystretch}{1.2}
	\label{tab:notations}
	\begin{tabular}{ll}
		\hline\hline
		Notation & Description \\
		\hline
		$ \bm{X} =$  $ \left( \bm{x}_{1}, \bm{x}_{2},\ldots, \bm{x}_{n}, \ldots, \bm{x}_{\infty} \right) $   & A set of training instances \\
		$ \bm{x}_{n} = $ $ \left( x_{n,1}, x_{n,2}, \ldots, x_{n,d} \right) $  &  $ d $-dimensional training instance (the $ n $th instance)  \\ 
		$ \bm{L} = \left( \bm{l}_{1}, \bm{l}_{2}, \ldots, \bm{l}_{n}, \ldots, \bm{l}_{\infty} \right) $ & A set of relevant label sets for $ \bm{X} $ \\
		$ \bm{l}_{n} = ( l_{n,1}, l_{n,2}, \ldots, l_{n, N_{l}} ) $ & A set of relevant label for $ \bm{x}_{n} $ \\
		$ N_{l} $ & Dimension of the relevant label set for $ \bm{X} $ \\
		$ \bm{Y} = (\bm{y}_{1}, \bm{y}_{2}, \ldots, \bm{y}_{K}) $  &  A set of prototype nodes (the $ k $th node) \\ 
		$ \bm{y}_{k} = $ $ \left( y_{1}, y_{2}, \ldots, y_{d} \right) $  & $ d $-dimensional prototype node \\ 
		$ \bm{S} = ( \sigma_{1},\sigma_{2},\ldots, \sigma_{K}) $  & A set of bandwidths for a kernel function \\ 
		$ \kappa_{\sigma} $  & Kernel function with a bandwidth $ \sigma $ \\ 
		CIM  &  Correntropy-Induced Metric\\ 
		$ k_{1} $, $ k_{2} $  & Indexes of the 1st and 2nd winner nodes  \\
		$ y_{k_{1}} $, $ y_{k_{2}} $ & The 1st and 2nd winner nodes \\
		$ V_{k_{1}} $, $ V_{k_{2}} $  & Similarities between an instance $ x_{n} $ and winner nodes ($ y_{k_{1}} $ and $ y_{k_{2}} $) \\
		$ V $  &  Predefined similarity threshold \\
		$ \alpha_{k_{1}} $  &  The number of instances that have accumulated by the node $ y_{k_{1}} $ \\
		$ \beta_{k_{1}} $  &  The number of labels that have accumulated by the node $ y_{k_{1}} $ \\
		$ \bm{y}_{k_{1}}^{+}  $  & Neighbor node of $ y_{k_{1}} $  \\
		$ N_{y} $  & The predefined number of neighbor nodes for $ y_{k_{1}} $  \\
		$ \lambda $ & Interval for adapting $ \sigma $ \\
		$ H_{i}^{+} $  &  Event that an instance has the $ i $th label \\
		$ H_{i}^{-} $  &  Event that an instance does not have the $ i $th label \\
		$ E_{i} $ & Event that the label $ l_{i} $ has the highest frequency among the $ N_{y} $ neighbor nodes of $ y_{k_{1}} $ \\
		$ P(E|H) $  & Likelihood for a label probability computation \\
		$ P(H) $  & Prior probability for a label probability computation \\
		$ P(H|E) $  & Posterior probability for a label probability computation \\
		$ c $  &  Label counter \\
		$ \bm{x}^{*} =$  $ \left( x_{1}^{*}, x_{2}^{*}, \ldots, x_{d}^{*} \right) $  &  $ d $-dimensional testing instance \\
		$ \bm{l}^{*} $  & Predicted label vector of the test instance $ \bm{x}^{*} $ \\
		\hline\hline
	\end{tabular}
\end{table*}

Several studies have realized multi-label classification by using a clustering algorithm. A typical type of clustering-based multi-label classification algorithm utilizes SOM \cite{colombini17, benyettou17}. Although the learning process of SOM is performed in an unsupervised learning manner, the convergence of the SOM network is significantly slow and unstable. The online semi-supervised GNG for multi-label classification utilizes GNG as a base classifier \cite{boulbazine18}. Unlike SOM-based algorithms, this method adaptively generates a topological network corresponding to the given instances. However, as mentioned in the literature review for clustering in the previous subsection, GNG-based algorithms do not satisfy the requirements of continual learning. Multi-label Classification via Incremental Clustering (MCIC) \cite{nguyen19} has realized continual learning by applying a continual clustering to a learning process. MCIC extracts and accumulates information from data by nodes through the continual clustering process that takes into account the arrival time of data. In addition, the continual clustering process also constructs and updates a distribution of labels in each node for a label estimation process. Some studies have employed Fuzzy ARTMAP \cite{carpenter92} to a base classifier of the multi-label classification \cite{benites10, benites17, yuan17, park18}. ARTMAP is composed of two ART architecture to realize an explicit supervised learning process. Although Fuzzy ARTMAP has various advantages, there is a well-known problem, i.e., high sensitivity to statistical overlapping between the generated categories \cite{marriott95}. This sensitivity problem results in category proliferation (i.e., disordered generation of categories), which leads to a high computational cost and deterioration in the classification performance. The recent ART-based algorithms in \cite{benites10, benites17, yuan17, park18} potentially have this problem.

\section{Proposed Algorithm}
\label{sec:mlca}
In this section, first the theoretical background of the CIM is briefly described. Next, the proposed algorithm, namely, MLCA, is explained in detail. Then, two variants of MLCA are introduced by modifying the calculation method of the CIM. Table \ref{tab:notations} summarizes the main notations used in this paper.

\subsection{Correntropy and Correntropy-Induced Metric}
\label{sec:cimDefinition}
Correntropy \cite{liu07} provides a generalized similarity measure between two arbitrary instances $ \bm{x} = (x_{1},x_{2},\ldots,x_{d}) $ and $ \bm{y} = (y_{1},y_{2},\ldots,y_{d}) $ as follows:
\begin{equation}
	C(\bm{x}, \bm{y}) = \textbf{E} \left[ \kappa_{\sigma} \left( \bm{x} - \bm{y} \right) \right],
\end{equation}
where $ \textbf{E} \left[ \cdot \right] $ is the expectation operation, and $ \kappa_{\sigma} \left( \cdot \right) $ denotes a positive definite kernel with a kernel bandwidth $ \sigma $. The correntropy can be defined as follows:
\begin{equation}
	\hat{C}(\bm{x}, \bm{y}) = \frac{1}{d} \sum_{i=1}^{d} \kappa_{\sigma} \left( x_{i} - y_{i} \right).
	\label{eq:correntropy}
\end{equation}

In this paper, we use the following Gaussian kernel in the correntropy:
\begin{equation}
	\kappa_{\sigma} \left( x_{i} - y_{i} \right) = \exp \left[ - \frac{\left( x_{i} - y_{i} \right)^{2} }{2 \sigma^{2}} \right].
	\label{eq:gaussian}
\end{equation}

A nonlinear metric called CIM is derived from the correntropy \cite{liu07}. The CIM quantifies the similarity between two instances as follows:
\begin{equation}
	\mathrm{CIM}\left(\bm{x}, \bm{y}, \sigma \right) = \left[ \kappa_{\sigma} (0) - \hat{C}(\bm{x}, \bm{y}) \right]^{\frac{1}{2}},
	\label{eq:defcim}
\end{equation}
where $ \kappa_{\sigma} (0) = 1 $ from (\ref{eq:gaussian}). Here, thanks to the Gaussian kernel without a coefficient $ \frac{1}{\sqrt{2\pi}\sigma} $ as defined in (\ref{eq:gaussian}), a range of the CIM is limited to $ \left[0,1 \right] $.

\subsection{Learning Procedure}
\label{sec:learningProcedure}
We use the following notations: Training instances are $ \bm{X} =$  $ \left( \bm{x}_{1}, \bm{x}_{2},\ldots, \bm{x}_{n}, \ldots, \bm{x}_{\infty} \right) $ where $ \bm{x}_{n} = $ $ \left( x_{n,1}, x_{n,2}, \ldots, x_{n,d} \right) $ is a $ d $-dimensional feature vector. Relevant label sets for $ \bm{X} $ are $ \bm{L} = \left( \bm{l}_{1}, \bm{l}_{2}, \ldots, \bm{l}_{n}, \ldots, \bm{l}_{\infty} \right) $ where $ \bm{l}_{n} = ( l_{n,1}, l_{n,2}, \ldots,$  $ l_{n,\infty} ) $ is a binary vector used to show a set of relevant labels for $ \bm{x}_{n} $. The dimension of $ \bm{l}_{n} $ increases with the number of label types (classes) in the learned training instances. Note that MLCA is capable of continual learning, the algorithm can accept any number of training instances and labels. A set of prototype nodes in MLCA at the point of the presentation of instance $ \bm{x}_{n} $ is $ \bm{Y} = (\bm{y}_{1}, \bm{y}_{2}, \ldots, \bm{y}_{K}) $ ($ K \in \mathbb{Z}^{+} $) where a node $ \bm{y}_{k} = $ $ \left( y_{1}, y_{2}, \ldots, y_{d} \right) $ has the same dimension as $ \bm{x}_{n} $. Furthermore, each node $ \bm{y}_{k} $ has an individual bandwidth $ \sigma $ for the CIM, i.e., $ \bm{S} = ( \sigma_{1},\sigma_{2},\ldots, \sigma_{K}) $.

The learning procedure of MLCA is divided into five parts: 1) initialization process for a bandwidth of a kernel function in the CIM, 2) winner node selection, 3) vigilance test, 4) node learning, and 5) label probability computation. Each of them is explained in the following subsections.

\subsubsection{Initialization Process for a Bandwidth of a Kernel Function in the CIM}
\label{sec:initKernel}
Similarity measurement between an instance and a node has a large impact on the performance of clustering algorithms. MLCA uses the CIM as a similarity measure. As defined in (\ref{eq:defcim}), the state of the CIM is controlled by a bandwidth $ \sigma $ of a kernel function which is a data-dependent parameter.

In general, the bandwidth of a kernel function can be estimated from $ \lambda $ instances belonging to a certain distribution \cite{henderson12}, which is defined as follows:
\begin{equation}
\bm{\Sigma} = U(F_{\nu})  \bm{\Gamma} \lambda^{-\frac{1}{2\nu+d}},
\label{eq:sigmaEst1}
\end{equation}
\begin{equation}
U(F_{\nu}) = \left( \frac{\pi ^{d/2} 2^{d+\nu-1}(\nu !)^{2}R(F)^{d}}{\nu \kappa_{\nu}^{2}(F)\left[(2\nu)!!+(d-1)(\nu!!)^{2}\right]} \right)^{\frac{1}{2\nu+d}},
\label{eq:sigmaEst2}
\end{equation}
where $ \bm{\Gamma} $ denotes a rescale operator ($ d $-dimensional vector) which is defined by a standard deviation of the $ d $ attributes among $ \lambda $ instances. $ \nu $ is the order of a kernel. $ R(F) $ is a roughness function. $ \kappa_{\nu}(F) $ is the moment of a kernel. In this paper, we utilize the Gaussian kernel for the CIM. Therefore, $ \nu = 2 $, $ R(F) = (2\sqrt{\pi})^{-1} $, and $ \kappa_{\nu}^{2}(F) = 1 $ are derived. The details of the derivation of (\ref{eq:sigmaEst1}) and (\ref{eq:sigmaEst2}) can be found in \cite{henderson12}.

In MLCA, the initial state of $ \sigma $ in the CIM is defined by training instances. When a new node $ \bm{y}_{K+1} $ is generated from $ \bm{x}_{n} $, a bandwidth $ \sigma_{K+1} $ is estimated from the past $ \lambda $ instances, i.e., $  \left( \bm{x}_{n-\lambda},\ldots, \bm{x}_{n-2}, \bm{x}_{n-1} \right) $, by using (\ref{eq:sigmaEst1}) and (\ref{eq:sigmaEst2}) with $ \nu = 2 $, $ R(F) = (2\sqrt{\pi})^{-1} $, and $ \kappa_{\nu}^{2}(F) = 1 $, as follows: 
\begin{equation}
	\bm{\Sigma} = \left( \frac{4}{2+d} \right)^{\frac{1}{4+d}}  \bm{\Gamma}  \lambda^{-\frac{1}{4+d}},
	\label{eq:SIGMA}
\end{equation}
where $ \bm{\Gamma} $ denotes a rescale operator ($ d $-dimensional vector) which is defined by a standard deviation of the $ d $ attributes among the past $ \lambda $ training instances of MLCA. Here, $ \bm{\Sigma} $ contains the bandwidth of each attribute. 

In this paper, the median of $ \bm{\Sigma} $ is selected as a representative bandwidth for the new node $ \bm{y}_{K+1} $, i.e.,
\begin{equation}
	\sigma_{K+1} = \mathrm{median} \left( \bm{\Sigma} \right).
	\label{eq:sigma}
\end{equation}


\subsubsection{Winner Node Selection}
Once an instance $ \bm{x}_{n} $ is presented to MLCA, two nodes which have a similar state to the instance $ \bm{x}_{n} $ are selected, namely, winner nodes $ \bm{y}_{k_{1}} $ and $ \bm{y}_{k_{2}} $. The winner nodes are determined based on the state of the CIM as follows:
\begin{equation}
	k_{1} = \arg \min_{k \in K}\left[ \text{CIM}\left(\bm{x}_{n}, \bm{y}_{k}, \text{mean}(\bm{S}) \right) \right],
	\label{eq:winnerCIM1}
\end{equation}
\vspace{-5pt}
\begin{equation}
	k_{2} = \arg \min_{k \in K \backslash \{k_{1}\}}\left[ \text{CIM}\left(\bm{x}_{n}, \bm{y}_{k}, \text{mean}(\bm{S}) \right) \right],
	\label{eq:winnerCIM2}
\end{equation}
\noindent where $ k_{1} $ and $ k_{2} $ denote indexes of the 1st and 2nd winner nodes i.e., $ \bm{y}_{k_{1}} $ and $ \bm{y}_{k_{2}} $, respectively. $ \bm{S} $ is a bandwidth for a kernel function of the CIM in each node.

Note that when there is no node in MLCA, the ($ \lambda $+1)th instance becomes the initial node (i.e., $ \bm{y}_{1} = \bm{x}_{\lambda+1} $). In the case, the bandwidth of $ \bm{y}_{1} $ is estimated from the 1st to $ \lambda $th instances in a set of training instances $ \bm{X} $ by (\ref{eq:sigmaEst1})-(\ref{eq:sigma}), and the next instance is given without vigilance test until the 1st and 2nd winner nodes can be defined.

\subsubsection{Vigilance Test}
\label{sec:vigilance}
Similarities between an instance $ \bm{x}_{n} $ and the 1st and 2nd winner nodes are defined as follows:
\begin{equation}
	V_{k_{1}} =  \mathrm{CIM}\left(\bm{x}_{n}, \bm{y}_{k_{1}}, \text{mean}(\bm{S}) \right),
	\label{eq:cim1}
\end{equation}
\vspace{-5pt}
\begin{equation}
	V_{k_{2}} = \mathrm{CIM}\left(\bm{x}_{n}, \bm{y}_{k_{2}}, \text{mean}(\bm{S}) \right).
	\label{eq:cim2}
\end{equation}

The vigilance test classifies the relationship between an instance and a node into three cases by using a predefined similarity threshold $ V $.

\begin{itemize}
		[ 
		\setlength{\IEEElabelindent}{\dimexpr-\labelwidth-\labelsep}
		\setlength{\itemindent}{\dimexpr\labelwidth+\labelsep}
		\setlength{\listparindent}{\parindent}
		] 
	
	\item Case I \\
	\indent The similarity between an instance $ \bm{x}_{n} $ and the 1st winner node $ \bm{y}_{k_{1}} $ is larger (i.e., less similar) than the similarity threshold $ V $, namely:
	\begin{equation}
		V_{k_{1}} > V.
		\label{eq:condition1}
	\end{equation}
	
	If (\ref{eq:condition1}) is satisfied, $ V_{k_{2}} > V $ is also satisfied since $ V_{k_{2}} > V_{k_{1}} > V $. Thus, a new node is defined as $ \bm{y}_{K+1} = \bm{x}_{n} $, and the bandwidth $ \sigma_{K+1} $ is defined by (\ref{eq:sigma}). 
	
	Moreover, the following two counters $ \alpha $ and $ \beta $ are updated. One counter $ \alpha $ is the number of instances that have been accumulated by the node $ \bm{y}_{k_{1}} $, which is updated as follows:
	\begin{equation}
		\alpha_{k_{1}} \leftarrow \alpha_{k_{1}} + 1.
		\label{eq:countNode}
	\end{equation}
	
	The other counter $ \beta $ is the number of labels that have been accumulated by the node $ \bm{y}_{k_{1}} $, which is updated as follows:
	\begin{equation}
		\bm{\beta}_{k_{1}} \leftarrow \bm{\beta}_{k_{1}} + \bm{l}_{k_{1}}.
		\label{eq:countLabel}
	\end{equation}
	
	\vspace{0.5mm}
	
	\item Case I\hspace{-.1em}I \\
	\indent A similarity between an instance $ \bm{x}_{n} $ and the 1st winner node $ \bm{y}_{k_{1}} $ is smaller (i.e., more similar) than the similarity threshold $ V $, and a similarity between the instance $ \bm{x}_{n} $ and the 2nd winner node $ \bm{y}_{k_{2}} $ is larger (i.e., less similar) than the similarity threshold $ V $, namely:
	\begin{equation}
		V_{k_{1}} \leq V \text{, and } V_{k_{2}} > V.
		\label{eq:condition2}
	\end{equation}
	
	If (\ref{eq:condition2}) is satisfied, node learning is performed. In addition, counters $ \alpha_{k_{1}} $ and $ \bm{\beta}_{k_{1}} $ are updated by (\ref{eq:countNode}) and (\ref{eq:countLabel}), respectively.
	
	\vspace{0.5mm}
	
	\item Case I\hspace{-.1em}I\hspace{-.1em}I \\
	\indent Similarities between an instance $ \bm{x}_{l} $ and the 1st and 2nd winner nodes are both smaller (i.e., more similar) than the similarity threshold $ V $, namely:
	\begin{equation}
		V_{k_{1}} \leq V   \text{, and } V_{k_{2}} \leq V.
		\label{eq:condition3}
	\end{equation}
	
	If (\ref{eq:condition3}) is satisfied, node learning is performed.
	
\end{itemize}

\subsubsection{Node Learning}
\label{sec:nodeLearning}
Different node learning is performed based on the results of the vigilance test.

If Case I\hspace{-.1em}I, the state of the 1st winner node $ \bm{y}_{k_{1}} $ is updated as follows:
\begin{equation}
	\bm{y}_{k_{1}} \leftarrow \bm{y}_{k_{1}} + \frac{1}{\alpha_{k_{1}}} \left( \bm{x}
	_{n} - \bm{y}_{k_{1}} \right).
	\label{eq:updateNodeWeight1}
\end{equation}

When updating the node, the amount of change is divided by $ \alpha_{k_{1}} $, so the larger $ \alpha_{k_{1}} $ becomes, the smaller the node position changes. This is based on the idea that the information around a node, where instances are frequently given, is important and should be held by the node.

If Case I\hspace{-.1em}I\hspace{-.1em}I, the state of the 1st winner node $ \bm{y}_{k_{1}} $ is updated by (\ref{eq:updateNodeWeight1}). In addition, all neighbor nodes of  $ \bm{y}_{k_{1}} $ (i.e., $ \bm{y}_{k_{1}}^{+}  $) are also updated as follows:
\begin{equation}
	\bm{y}_{k_{1}}^{+} \leftarrow \bm{y}_{k_{1}}^{+}  + \frac{1}{N_{y} \alpha_{k_{1}}^{+} } \left( \bm{y}_{k_{1}} - \bm{y}_{k_{1}}^{+}  \right),
	\label{eq:updateNodeWeight2}
\end{equation}
\noindent where $ N_{y} $ is the predefined number of neighbor nodes for $ \bm{y}_{k_{1}} $. $ \alpha_{k_{1}}^{+} $ denotes the number of instances that have been accumulated by the node $ \bm{y}_{k_{1}}^{+} $. 

Equation (\ref{eq:updateNodeWeight2}) has the same concept as (\ref{eq:updateNodeWeight1}), but it should be less affected by the instance than $ \bm{y}_{k_{1}} $ because it is the neighbor node of $ \bm{y}_{k_{1}} $. Thus, $ N_{y}  $ is added as a coefficient.

\subsubsection{Label Probability Computation}
\label{sec:updateProbablity}
Similar to ML-$ k $NN, MLCA employs the Bayesian approach for label probability computation. The prior probability and likelihood are updated if the condition for Case I or Case I\hspace{-.1em}I of the vigilance test is satisfied. Note that ML-$ k $NN computes the prior probability and likelihood by using training instances repeatedly, while MLCA computes the prior probability and likelihood by using nodes which have accumulated label counts. As a result, MLCA realizes continual learning by computing the prior probability and likelihood whenever an instance is given.

To update the prior probability and likelihood, an instance $ \bm{x}_{n} $ with a set of labels $ \bm{l}_{n} $, the 1st winner node $ \bm{y}_{k_{1}} $ and its $ N_{y} $ neighbor nodes are considered. Note that $ N_{y} $ denotes the number of neighbor nodes and is a predefined parameter in MLCA. Here, we consider the situation where $ (n-1) $ instances have been given to MLCA. The likelihood $ P(E | H) $ is computed as follows:
\begin{equation}
	P(E_{n,i} | H_{n,i}^{\phi}) = \frac{\left( s + \bm{c}_{i}^{\phi} \right)}{\left[ s\times(N_{y}+1) + \sum\limits_{j=0}^{N_{y}} \bm{c}_{i,j}^{\phi} \right]},
	\nonumber
\end{equation}
\vspace{-5pt}
\begin{equation}
	\hspace{77pt} (i \in N_{l}, \hspace{1mm} \phi \in \{ +, - \}),
	\label{eq:likelihood1}
\end{equation}
where $ H_{i}^{+} $ is the event that an instance has the $ i $th label (i.e., $ \bm{l}_{i} = 1 $), and $ H_{i}^{-} $ is the event that an instance does not have the $ i $th label (i.e., $ \bm{l}_{i} = 0 $). $ E_{i} $ is the event that the frequency of the label $ l_{i} $ among the $ N_{y} $ neighbor nodes of $ \bm{y}_{k_{1}} $. $ N_{l} $ is a size of a label set. $ s $ is a smoothing parameter. In this paper, $ s $ is set to be 1 which yields the Laplace smoothing. A label counter $ \bm{c} $ is defined as follows:
\begin{equation}
	\begin{cases} 
		\bm{c}_{i, j=g_{i}}^{+} \leftarrow \bm{c}_{i, j=g_{i}}^{+} + 1, & \text{if \hspace{1mm}} l_{n,i} = 1 
		\vspace{1mm} \\  
		\bm{c}_{i, j=g_{i}}^{-} \leftarrow \bm{c}_{i, j=g_{i}}^{-} + 1,   & \text{otherwise}
	\end{cases}.
	\label{eq:labelCounterC}
\end{equation}

Here, $ g_{i} $ is a $ i $th attribute of an $ N_{l} $-dimensional counting vector $ \bm{g} = \left( g_{1},\ldots,g_{N_{l}} \right) $, which is defined as follows:
\begin{equation}
	g_{i} = \sum_{j \in N_{y}} \bm{\beta}_{i, j}, \hspace{2mm} (i \in N_{l}).
	\label{eq:countL1}
\end{equation}

In order to make the maximum value of $ g_{i} $ the same as the number of neighbor nodes $ N_{y} $, the following operation is performed.
\begin{equation}
	g_{i} \leftarrow \mathrm{round} \left[ N_{y} \cdot \frac{  g_{i} }{\mathrm{max}\left( \bm{g} \right)} \right].
	\label{eq:countL2}
\end{equation}

The prior probability $ P(H) $ is computed as follows:
\begin{equation}
	P(H_{i}^{+}) = \frac{\left( s + \sum\limits_{k=1}^{K}\bm{\beta}_{k,i} \right)}{ \left( s \times 2 + n \right) },  \hspace{2mm}(i \in N_{l}),
	\label{eq:prior1}
\end{equation}
\vspace{-5pt}
\begin{equation}
	P(H_{i}^{-}) =  1 - P(H_{i}^{+}), \hspace{2mm} (i \in N_{l}),
	\label{eq:prior2}
\end{equation}

\noindent where $ n $ denotes the number of instances that have been given.

The learning procedure of MLCA is summarized in Algorithm \ref{al:LearnMLCA}.

\begin{algorithm*}[htbp]
	\DontPrintSemicolon
	\caption{Learning procedure of MLCA}
	\label{al:LearnMLCA}
	\KwIn{\\
		\begin{itemize}
			\item[-] a training instance: $ \bm{x}_{n} = $ $ \left( x_{n,1}, x_{n,2}, \ldots, x_{n,d} \right) $, \\
			\item[-] a label set: $ \bm{l}_{n} = ( l_{n,1}, l_{n,2}, \ldots ) $, \\
			\item[-] prototype nodes: $ \bm{Y} = (\bm{y}_{1}, \bm{y}_{2}, \ldots, \bm{y}_{K}) $ ($ K \in \mathbb{Z}^{+} $), \\
			\item[-] bandwidths of $ \bm{Y} $: $ \bm{S} = ( \sigma_{1},\sigma_{2},\ldots, \sigma_{K})  $, \\
			\item[-] the number of instances that have been accumulated by nodes $ \bm{Y} $: $ \bm{\alpha} = (\alpha_{1}, \alpha_{2}, \ldots, \alpha_{K}) $, \\
			\item[-] the number of labels that have been accumulated by nodes $ \bm{Y} $: $ \bm{\beta} = (\bm{\beta}_{1}, \bm{\beta}_{2}, \ldots, \bm{\beta}_{K}) $, \\
			\item[-] a label counter: $ \bm{c} $, \\
			\item[-] the number of neighbor nodes: $ N_{y} $, \\
			\item[-] an interval for adapting $ \sigma $: $ \lambda $, \\
			\item[-] and a similarity threshold: $ V $. \\
		\end{itemize}
	}
	\KwOut{\\
		\begin{itemize}
			\item[-] updated prototype nodes: $ \bm{Y} $, \\
			\item[-] updated bandwidths of $ \bm{Y} $: $ \bm{S} $, \\
			\item[-] the updated number of instances that have been accumulated by the nodes $ \bm{Y} $: $ \bm{\alpha} $, \\
			\item[-] the updated number of labels that have been accumulated by the nodes $ \bm{Y} $: $ \bm{\beta} $, \\
			\item[-] the updated label counter: $ \bm{c} $, \\
			\item[-] a prior probability: $ P(H) $, \\
			\item[-] and a likelihood: $ P(E | H) $.
		\end{itemize}
	}
	\vspace{2mm}
	\SetKwBlock{Begin}{function}{end function}
	\Begin( \text{LearningMLCA($ \bm{x}_{n} $, $ \bm{l}_{n} $, $ \bm{Y} $, $ \bm{S} $, $ \bm{\alpha} $, $ \bm{\beta} $, $ \bm{c} $, $ N_{y} $, $ \lambda $, $ V $)}){
		Input an instance $ \bm{x}_{n} $.\\
		Input a label set $ \bm{l} _{n} $.\\
		\If{\upshape The index $ n $ is multiple of $ \lambda $ }{
			Compute a bandwidth for the CIM by (\ref{eq:sigma}).
		}
		\uIf{ $ K < 1 $ }{
			Generate a new node as $ \bm{y}_{K+1} = \bm{x}_{n} $. \\
			Assign a bandwidth $ \sigma_{K+1} $. \\
			Update $ \alpha_{k+1} $ and $ \bm{\beta}_{k+1} $ by (\ref{eq:countNode}) and (\ref{eq:countLabel}), respectively.
		}\Else{
			Compute the CIM by (\ref{eq:defcim}). \\
			Search indexes of winner nodes $ k_{1} $ and $ k_{2} $ by (\ref{eq:winnerCIM1}) and (\ref{eq:winnerCIM2}), respectively. \\
			\uIf{ \upshape$ \mathrm{CIM}\left(\bm{x}_{n}, \bm{y}_{k_{1}}, \mathrm{mean}(\bm{S}) \right) > V  $ }{
				Generate a new node as $ \bm{y}_{K+1} = \bm{x}_{l} $. \\
				Compute a bandwidth $ \sigma_{K+1} $ which is defined by (\ref{eq:sigma}). \\
				Update $ \alpha_{k+1} $ and $ \bm{\beta}_{k+1} $ by (\ref{eq:countNode}) and (\ref{eq:countLabel}), respectively. \\
				Update similarities between an instance $ \bm{x}_{n} $ and nodes $ \bm{Y} $ by the CIM. \\
				Update a likelihood $ P(E | H) $ by (\ref{eq:likelihood1}).
			}\Else{
				Update a state of $ \bm{y}_{k_{1}} $ by (\ref{eq:updateNodeWeight1}). \\
				Update $ \alpha_{k_{1}} $ and $ \bm{\beta}_{k_{1}} $ by (\ref{eq:countNode}) and (\ref{eq:countLabel}), respectively. \\
				\If{ \upshape$ \mathrm{CIM}\left(\bm{x}_{n}, \bm{y}_{k_{2}}, \mathrm{mean}(\bm{S}) \right) \leq V $ }{
					Update the state of $ N_{y} $ neighbor nodes of $ \bm{y}_{k_{1}} $ by (\ref{eq:updateNodeWeight2}). \\
				}
				Update a likelihood $ P(E | H) $ by (\ref{eq:likelihood1}). \\
			}
			Compute a prior probability by (\ref{eq:prior1}) and (\ref{eq:prior2}). \\
		} 
		\Return{ \upshape $ \bm{Y} $, $ \bm{S} $, $ \bm{\alpha} $, $ \bm{\beta} $, $ \bm{c} $, $ P(E | H) $, and $ P(H) $}.
	}
\end{algorithm*}

\subsection{Label Prediction Procedure}

We use the following notations: A testing instance is $ \bm{x}^{*} =$  $ \left( x_{1}^{*}, x_{2}^{*}, \ldots, x_{d}^{*} \right) $. A set of prototype nodes in MLCA after the learning procedure is $ \bm{Y} = \{\bm{y}_{1}, \bm{y}_{2}, \ldots, \bm{y}_{K}\} $. In addition, each node $ \bm{y}_{k} $ has an individual bandwidth $ \sigma_{k} $ for the CIM, i.e., $ S = ( \sigma_{1},\sigma_{2},\ldots, \sigma_{K}) $. Moreover, the likelihood $ P(E | H) $, the prior probability $ P(H) $, and the label counter $ \bm{\beta} $ are utilized for the Bayesian approach. 

First of all, the winner node $ \bm{y_{k_{1}}} $ for a test instance $ \bm{x}^{*} $ and its $ N_{y} $ neighbor nodes are determined. In the same manner as the learning procedure, a membership counting vector $ \bm{g} $ for the test instance $ \bm{x}^{*} $ is computed by (\ref{eq:countL2}). The posterior probability $ P(H | E) $ for the test instance $ \bm{x}^{*} $ is defined by the Bayes rule as follows:
\begin{equation}
	P(H_{i}^{+} | E_{\bm{g}_{i}}) = \frac{ P(E_{\bm{g}_{i}} | H_{i}^{+}) P(H_{i}^{+}) }{ \sum\limits_{ \phi \in \{ +,- \} }P(E_{\bm{g}_{i}} | H_{i}^{\phi})P(H_{i}^{\phi}) }, \hspace{1mm} (i \in N_{l}).
	\label{eq:posteriorP}
\end{equation}

A predicted label vector $ \bm{l}^{*} $ of the test instance $ \bm{x}^{*} $ is determined by a simple thresholding method as follows:
\begin{equation}
	l_{i}^{*} = 
	\begin{cases} 
		1, & \text{if \hspace{1mm}} P(H_{i}^{+} | E_{\bm{g}_{i}}) >0.5 
		\vspace{1mm} \\  
		0, & \text{otherwise}
	\end{cases}, \hspace{2mm} (i \in N_{l}).
	\label{eq:predLabel}
\end{equation}

Since the label prediction process is a binary classification, it is reasonable to specify the threshold to 0.5. ML-$ k $NN also uses the same value.

The prediction procedure of MLCA is summarized in Algorithm \ref{al:predMLCA}.

\subsection{Attribute Processing for the CIM}
\label{sec:CIM_Individual_Clustering}
As shown in (\ref{eq:correntropy}) and (\ref{eq:gaussian}), the CIM in MLCA uses a common bandwidth $ \sigma $ even if the range of attribute values is different, and the average value as the similarity between an instance $ \bm{x}_{n} $ and a node $ \bm{y}_{k} $. Therefore, if the range of attribute values is significantly different, a specific attribute may have a large impact on the value of the CIM when the common bandwidth $ \sigma $ is not appropriate for that attribute.

In this section, we propose two approaches in order to mitigate the above-mentioned effects: 1) one approach calculates the CIM by using an each individual attribute separately, and the average CIM value is used for similarity measurement, and 2) the other approach applies a clustering algorithm to attribute values, then attributes with similar value ranges are grouped. The CIM is calculated by using an each attribute group, and the average CIM value is used for similarity measurement.

\subsubsection{Individual-based Approach}
\label{sec:CIM_Individual}
In this approach, the CIM is calculated by using an each individual attribute separately, and the average CIM value is used for similarity measurement. The similarity between an instance $ \bm{x}_{n} $ and a node $ \bm{y}_{k} $ is defined by the $ \text{CIM}^{\text{I}} $ as follows:
\begin{equation}
	\mathrm{CIM}^{\text{I}}\left(\bm{x}_{n}, \bm{y}_{k}, \bm{\sigma}_{k} \right) \!=\! \frac{1}{d} \sum_{i=1}^{d} \left[ \kappa_{\sigma_{k, i}} (0) \!-\! C_{\text{I}}(x_{n, i}, y_{k, i}) \right]^{\frac{1}{2}},
	\label{eq:defcim_Individual}
\end{equation}
\begin{equation}
	C_{\text{I}}(x_{n, i}, y_{k, i}) = \kappa_{\sigma_{k, i}} \left( x_{n, i} - y_{k, i} \right),
\end{equation}
where $ \bm{\sigma}_{k} = \left( \sigma_{1}, \sigma_{2},\ldots, \sigma_{d} \right) $ is a bandwidth of a node $ \bm{y}_{k} $. A bandwidth for the $ i $th attribute is defined as follows:
\begin{equation}
	\sigma_{i} = \left( \frac{4}{2+d} \right)^{\frac{1}{4+d}}  \Gamma_{i}  \lambda^{-\frac{1}{4+d}},
	\label{eq:sigma_Individual}
\end{equation}
where $ \Gamma_{i} $ denotes a rescale operator which is defined by a standard deviation of $ i $th attribute values among the $ \lambda $ instances.

In this paper, MLCA with the individual-based approach is called MLCA-Individual (MLCA-I).

\begin{algorithm}[htbp]
	\DontPrintSemicolon
	\caption{Prediction procedure of MLCA}
	\label{al:predMLCA}
	\KwIn{\\
		\begin{itemize}
			\item[-] a testing instance: $ \bm{x}^{*} = $ $ \left( x_{1}^{*}, x_{2}^{*}, \ldots, x_{d}^{*} \right) $, \\
			\item[-] prototype nodes: $ \bm{Y} \!\!=\! (\bm{y}_{1}, \bm{y}_{2}, \ldots, \bm{y}_{K} ) $ ($ K \!\in\! \mathbb{Z}^{+} $), \\
			\item[-] the bandwidths of $ \bm{Y} $: $\! \bm{S} \!=\! ( \sigma_{1},\!\sigma_{2},\ldots, \!\sigma_{K})  $, \\
			\item[-] the number of labels that have been accumulated \\by nodes $ \bm{Y} $: $ \bm{\beta} = (\beta_{1}, \beta_{2}, \ldots, \beta_{K}) $, \\
			\item[-] the number of neighbor nodes: $ N_{y} $, \\
			\item[-] and a similarity threshold: $ V $. \\
		\end{itemize}
	}
	\KwOut{\\
		\begin{itemize}
			\item[-] a predicted label vector: $ \bm{l}^{*} $.
		\end{itemize}
	}
	\vspace{2mm}
	\SetKwBlock{Begin}{function}{end function}
	\Begin( \text{PredictMLCA($ \bm{x}^{*} $, $ \bm{Y} $, $ \bm{S} $, $ \bm{\beta} $, $ N_{y} $, $ V $)}){
		Input an instance $ \bm{x}^{*} $.\\
		Compute similarities between an instance $ \bm{x}^{*} $ and nodes $ \bm{Y} $ by the CIM. \\
		Compute a membership counting vector $ \bm{g}^{*} $ by (\ref{eq:countL2}). \\
		Compute a posterior probability $ P(H | E) $ by (\ref{eq:posteriorP}). \\
		Determine a predicted label vector $ \bm{l}^{*} $ by (\ref{eq:predLabel}). \\
		\Return{ \upshape $ \bm{l}^{*} $}.
	}
\end{algorithm}

\subsubsection{Clustering-based Approach}
\label{sec:CIM_Clustering}
In this approach, for every $ \lambda $ instances, the clustering algorithm presented in Section \ref{sec:learningProcedure} is applied to the attribute values. Each attribute value of $ \lambda $ instances is regarded as a one-dimensional vector and used as the input to the clustering algorithm. As a result, attributes with similar value ranges are grouped together, i.e., an instance $ \bm{x}_{n} = $ $ \left( x_{1}, x_{2}, \ldots, x_{d} \right) $ is transformed into $ \bm{x}_{n}^{\text{C}} = \left( \bm{u}_{n,1}, \bm{u}_{n,2},\ldots \bm{u}_{n,J} \right) $ $\left( J \leq d \right)$ by the clustering algorithm, where $ \bm{u}_{j} $ represents the $ j $th attribute group. The dimensionality of each attribute group is represented by $ \bm{d} = \left( d_{1}, d_{2},\ldots, d_{J} \right) $ where $ d_{j} $ is the dimensionality of the $ j $th attribute group. 

In this approach, the similarity between an instance $ \bm{x}_{n}^{\text{C}} $ and a node $ \bm{y}_{k} $ is defined by the $ \text{CIM}^{\text{C}} $ as follows:
\begin{equation}
	\mathrm{CIM}^{\text{C}} \! \left(\bm{x}_{n}^{\text{C}}, \bm{y}_{k}^{\text{C}}, \bm{\sigma}_{k}^{\text{C}} \right) =\frac{1}{J} \sum_{j=1}^{J} \left[ \kappa_{\sigma_{j}} (0) - \hat{C}_{\text{C}}(\bm{u}_{j}, \bm{v}_{j})  \right]^{\frac{1}{2}},
	\label{eq:defcim_Cluster}
\end{equation}
\begin{equation}
	\hat{C}_{\text{C}}(\bm{u}_{j}, \bm{v}_{j}) = \frac{1}{d_{j}} \sum_{i=1}^{d_{j}} \kappa_{\sigma_{j}} \left( u_{i} - v_{i} \right),
\end{equation}

\noindent where $ \bm{y}_{k}^{\text{C}} = \left( \bm{v}_{1}, \bm{v}_{2},\ldots \bm{v}_{J} \right) $ is a node $ \bm{y}_{k} $, but its attributes are grouped by referencing to the attribute indexes of $ \bm{x}_{n}^{\text{C}} $. A bandwidth $ \sigma_{j} $ is defined as follows:
\begin{equation}
	\sigma_{j} = \frac{1}{d_{j}} \sum_{i=1}^{d_{j}} \left[ \left( \frac{4}{2+d_{j}} \right)^{\frac{1}{4+d_{j}}}  \Gamma_{i}  \lambda^{-\frac{1}{4+d_{j}}} \right],
	\label{eq:sigma_Clustering}
\end{equation}
where $ \Gamma_{i} $ denotes a rescale operator which is defined by the standard deviation of the $ i $th attribute value in the $ j $th attribute group among the $ \lambda $ instances.

In this paper, MLCA with the clustering-based approach is called MLCA-Clustering (MLCA-C).

\begin{figure}[htbp]
	\centering
	\subfloat[General Approach]{
		\includegraphics[width=2.0in]{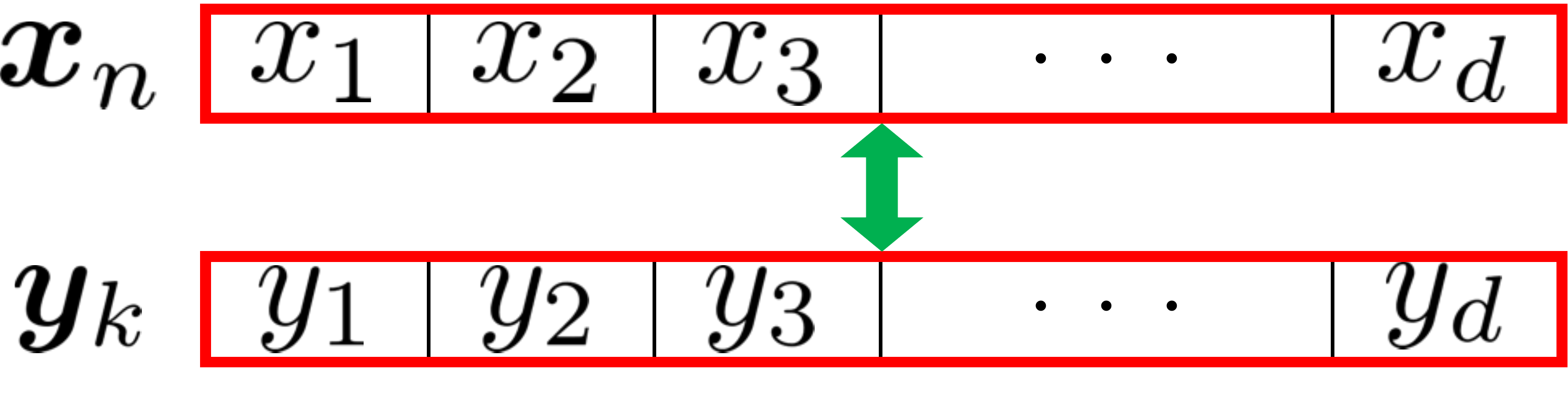}
		\label{fig:cim_original}
	}
	\vspace{5pt}
	\hfil
	\subfloat[Individual-based Approach (MLCA-I)]{
		\includegraphics[width=2.0in]{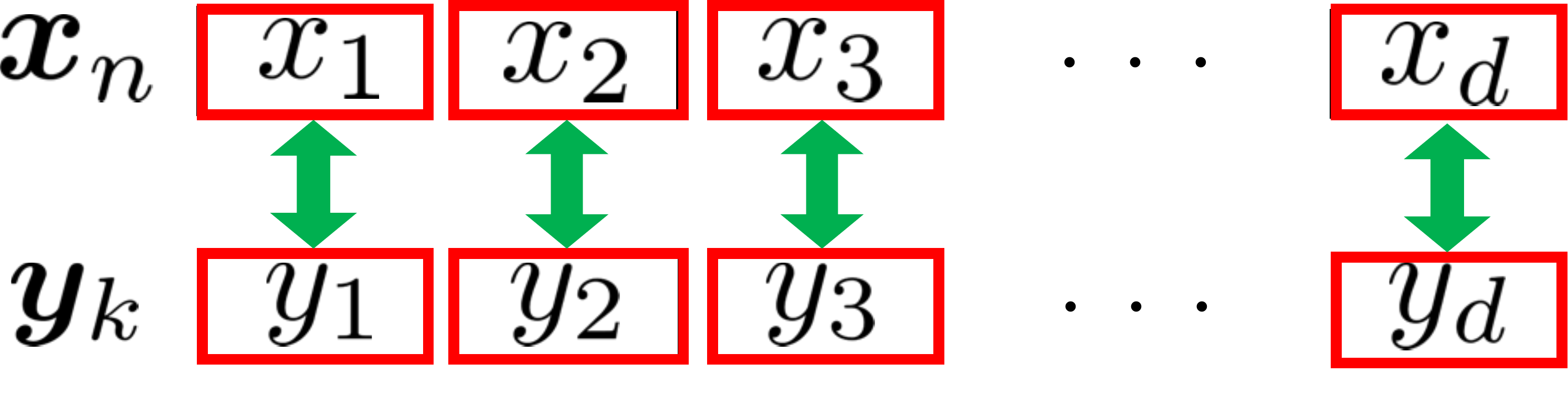}
		\label{fig:cim_individual}
	}
	\\
	\vspace{5pt}
	\subfloat[Clustering-based Approach (MLCA-C)]{
		\includegraphics[width=3.2in]{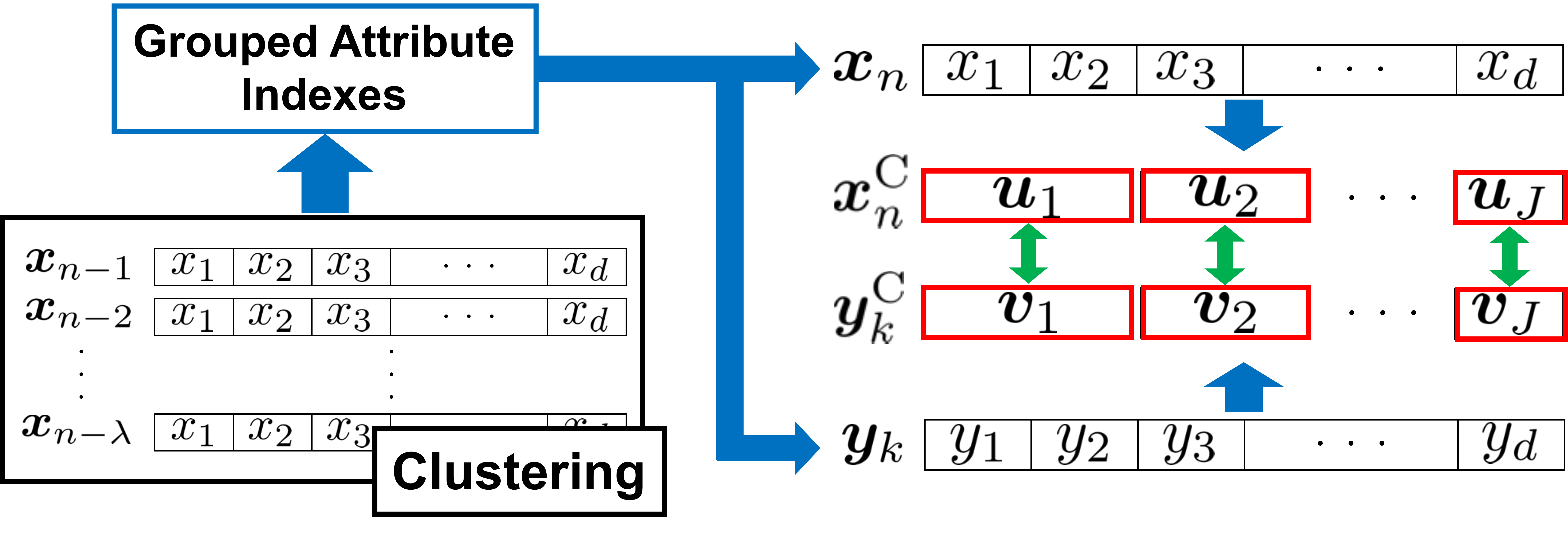}
		\label{fig:cim_clustering}
	}
	\\
	\vspace{5pt}
	\includegraphics[width=1.2in]{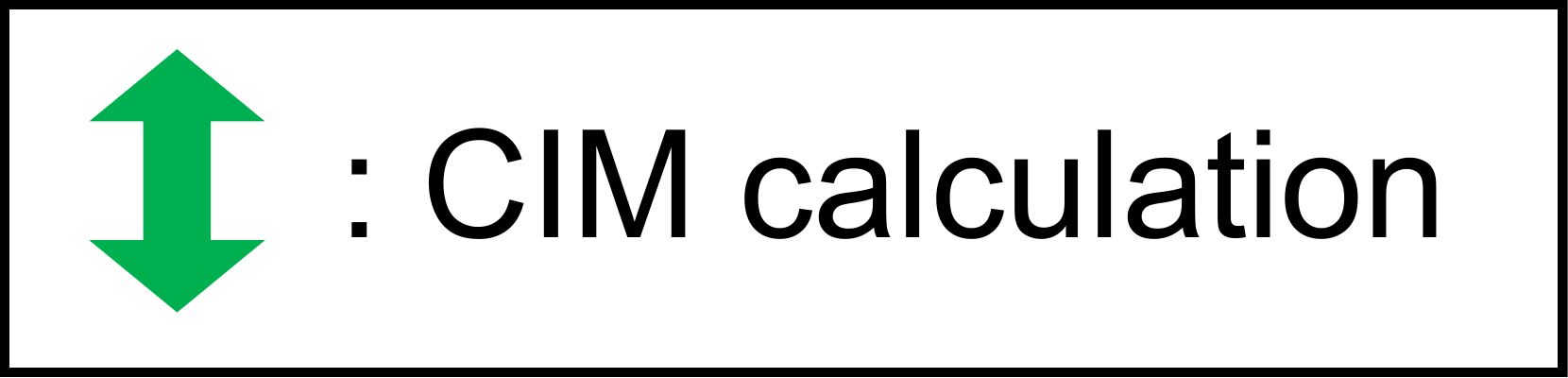}
	\caption{Differences in the CIM calculation.}
	\label{fig:cim_individual_clustering}
\end{figure}

The differences in attribute processing among the general approach, the individual-based approach (MLCA-I), and the clustering-based approach (MLCA-C) are depicted in Fig. \ref{fig:cim_individual_clustering}.

The codes of MLCA, MLCA-I, and MLCA-C are available on GitHub\footnote{https://github.com/Masuyama-lab/MLCA}.

\section{Simulation Experiments}
\label{sec:experiment}

In this section, the ability of MLCA is evaluated from various perspectives. First, the continual learning ability of MLCA is analyzed with a two-dimensional synthetic multi-label dataset under the stationary and non-stationary environments. Next, the classification performance of MLCA is compared with other algorithms by using real-world multi-label datasets. Third, we evaluate the effect of a multi-epoch learning process to MLCA. Finally, we analyze the computational complexity of each algorithm.

\subsection{Evaluation Metrics}
\label{sec:metrics}
We use the six metrics to evaluate the classification performance of multi-label classification algorithms \cite{zhang13}.

\begin{figure*}[htbp]
	\begin{minipage}{0.5\hsize}
		\centering
		\subfloat[Entire Dataset]{
			\includegraphics[height=1.05in]{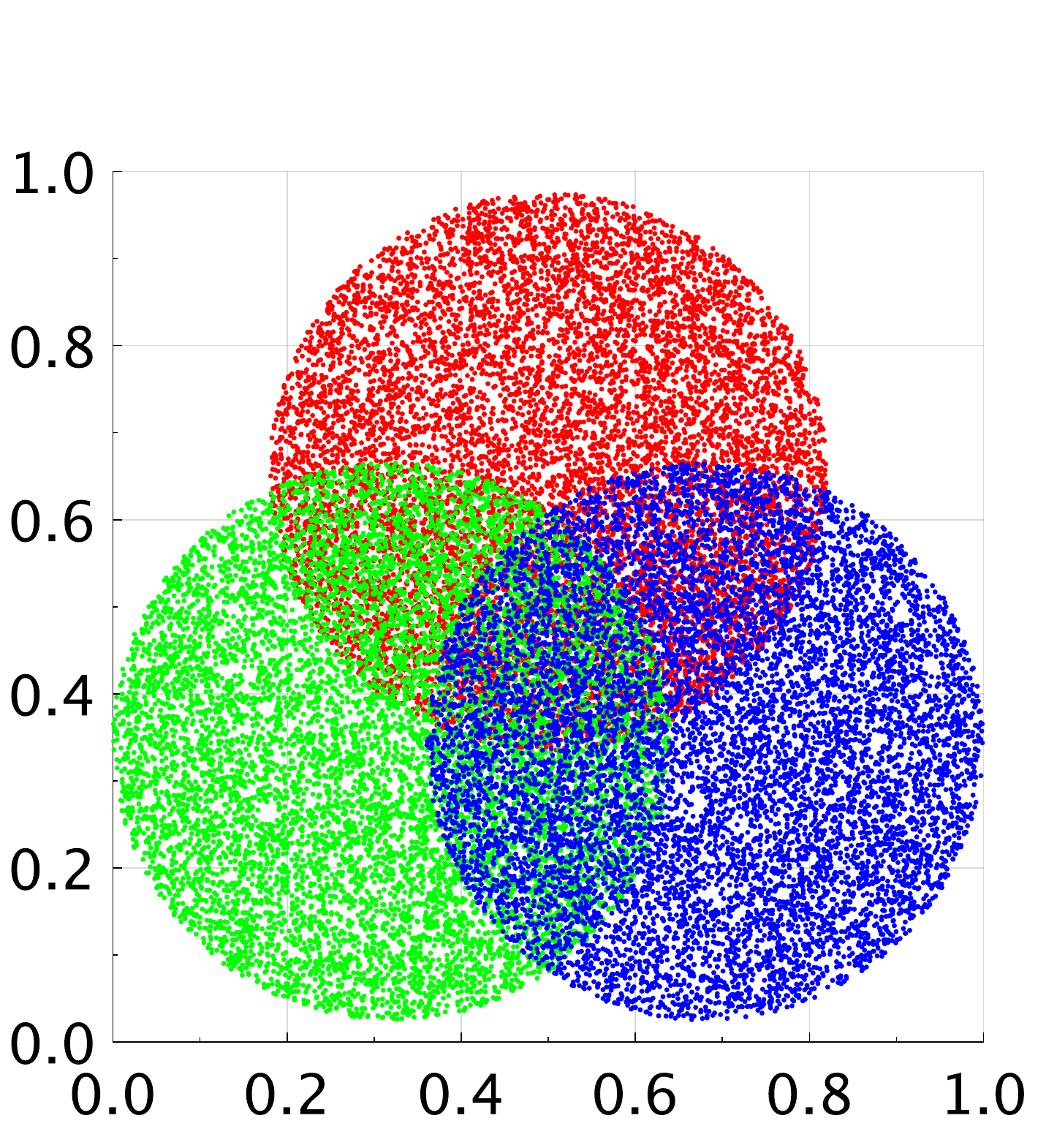}
			\label{fig:entireDataset}
		}
		\hfil
		\subfloat[Label Set]{
			\includegraphics[height=1.05in]{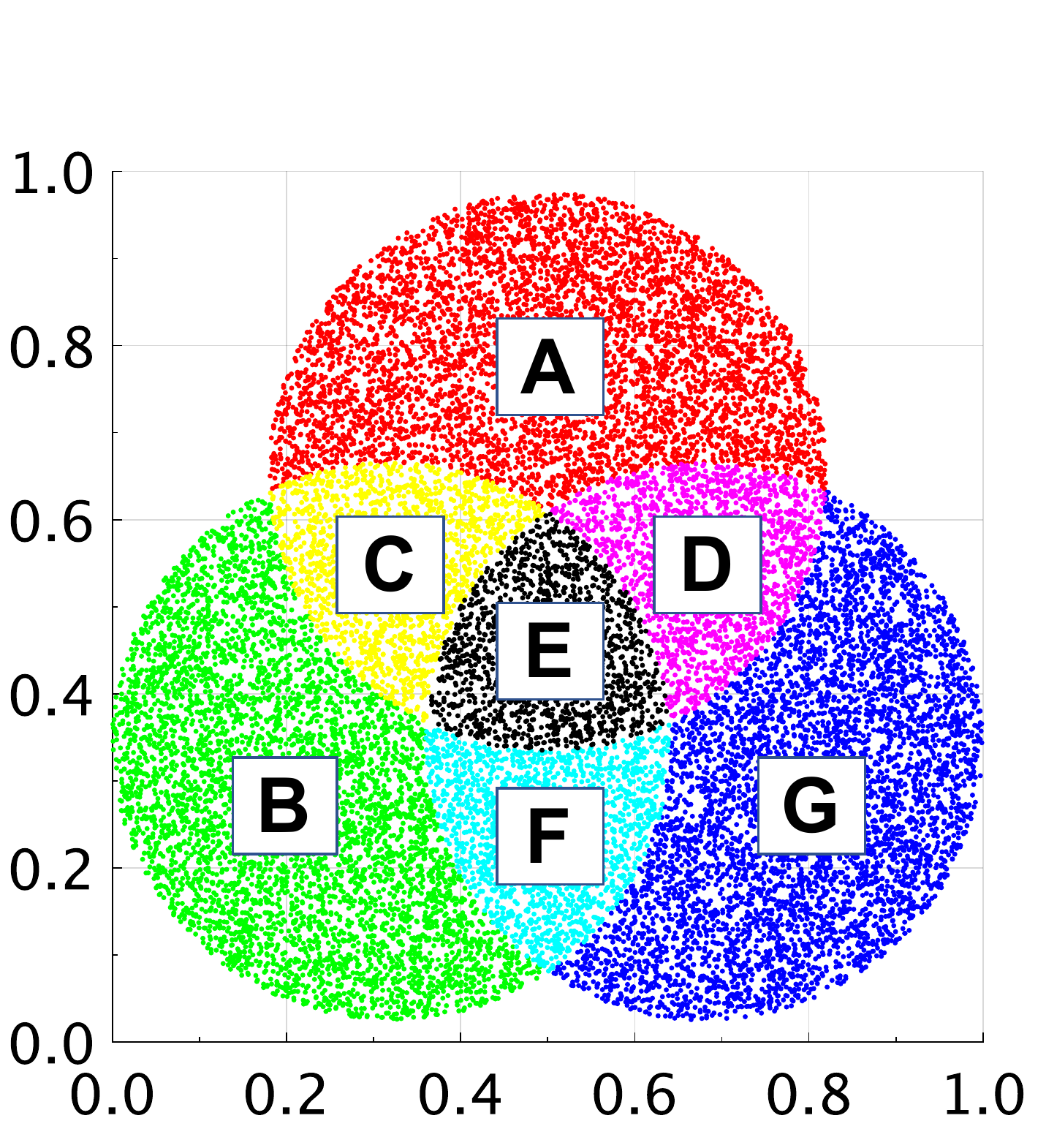}
			\label{fig:twoDimSynthetic_1dist}
		}
		\subfloat{
			\raisebox{-0.5mm}{
				\includegraphics[height=1.0in]{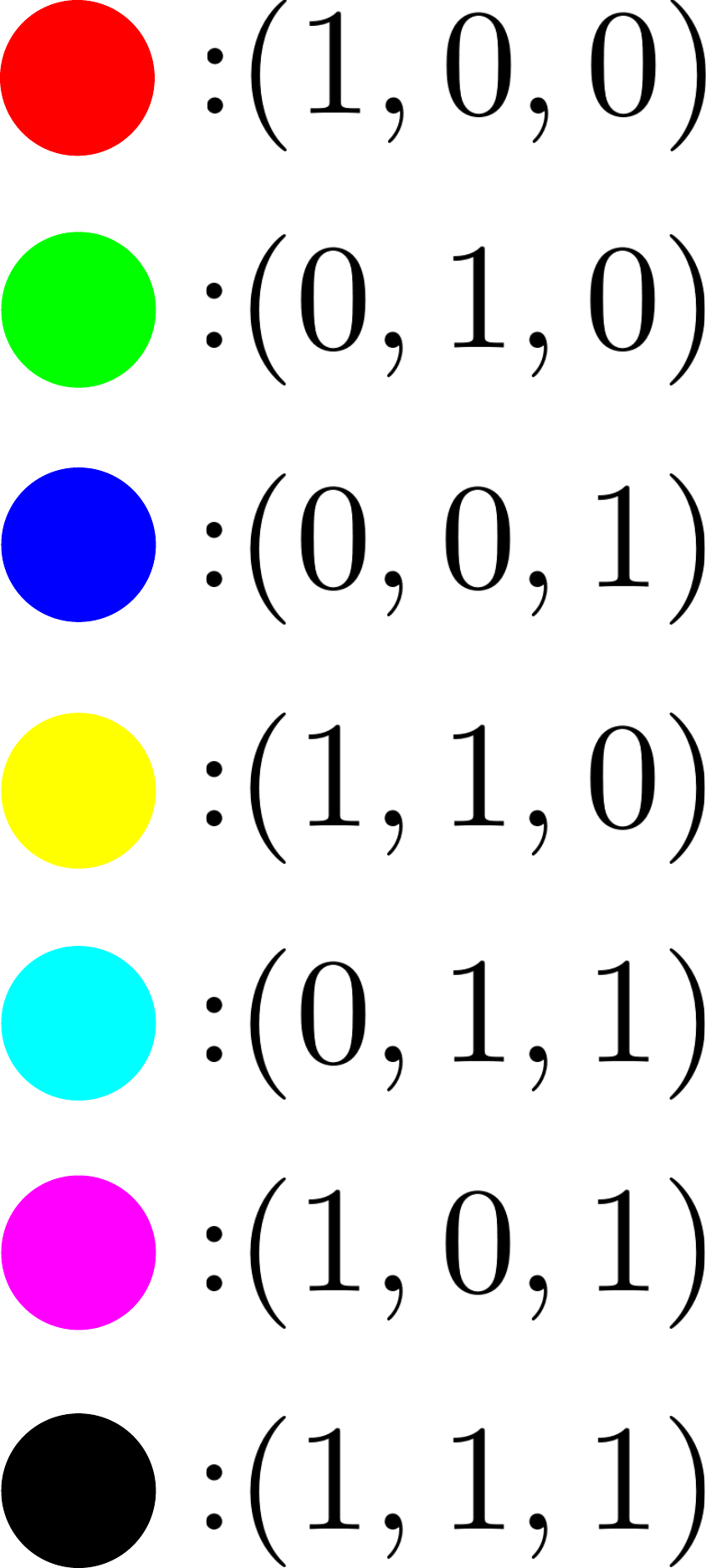}
			}
		}
		\caption{Two-dimensional synthetic multi-label dataset and \\ its label sets.}
		\label{fig:twoDimSynthetic_1dist_dataset}
	\end{minipage}
	\begin{minipage}{0.5\hsize}
		\centering
		\subfloat[Distribution \#1]{
			\includegraphics[height=1.05in]{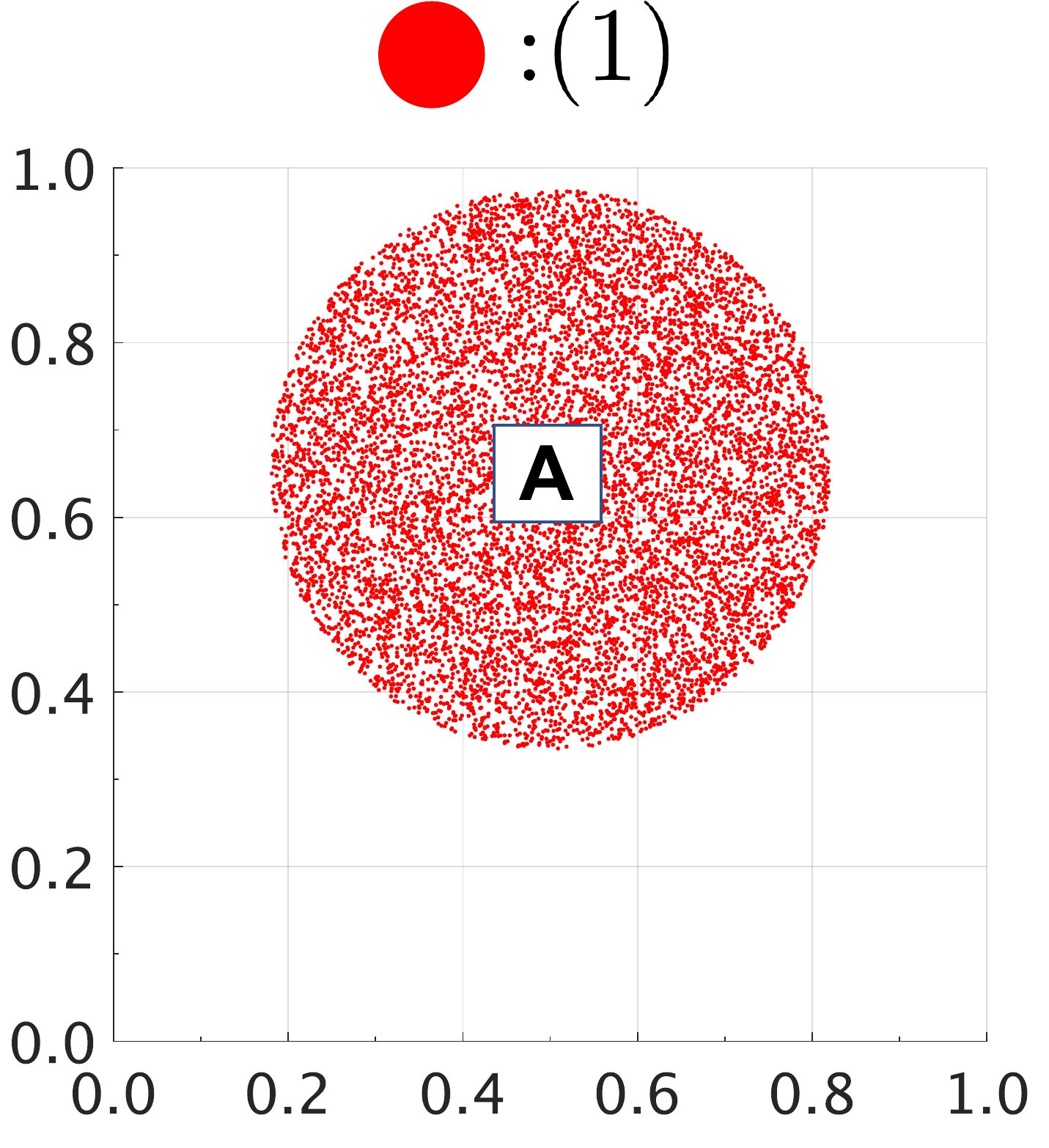}
			\label{fig:twoDimSynthetic_3dist_1}
		}
		\hfil
		\subfloat[Distribution \#2]{
			\includegraphics[height=1.05in]{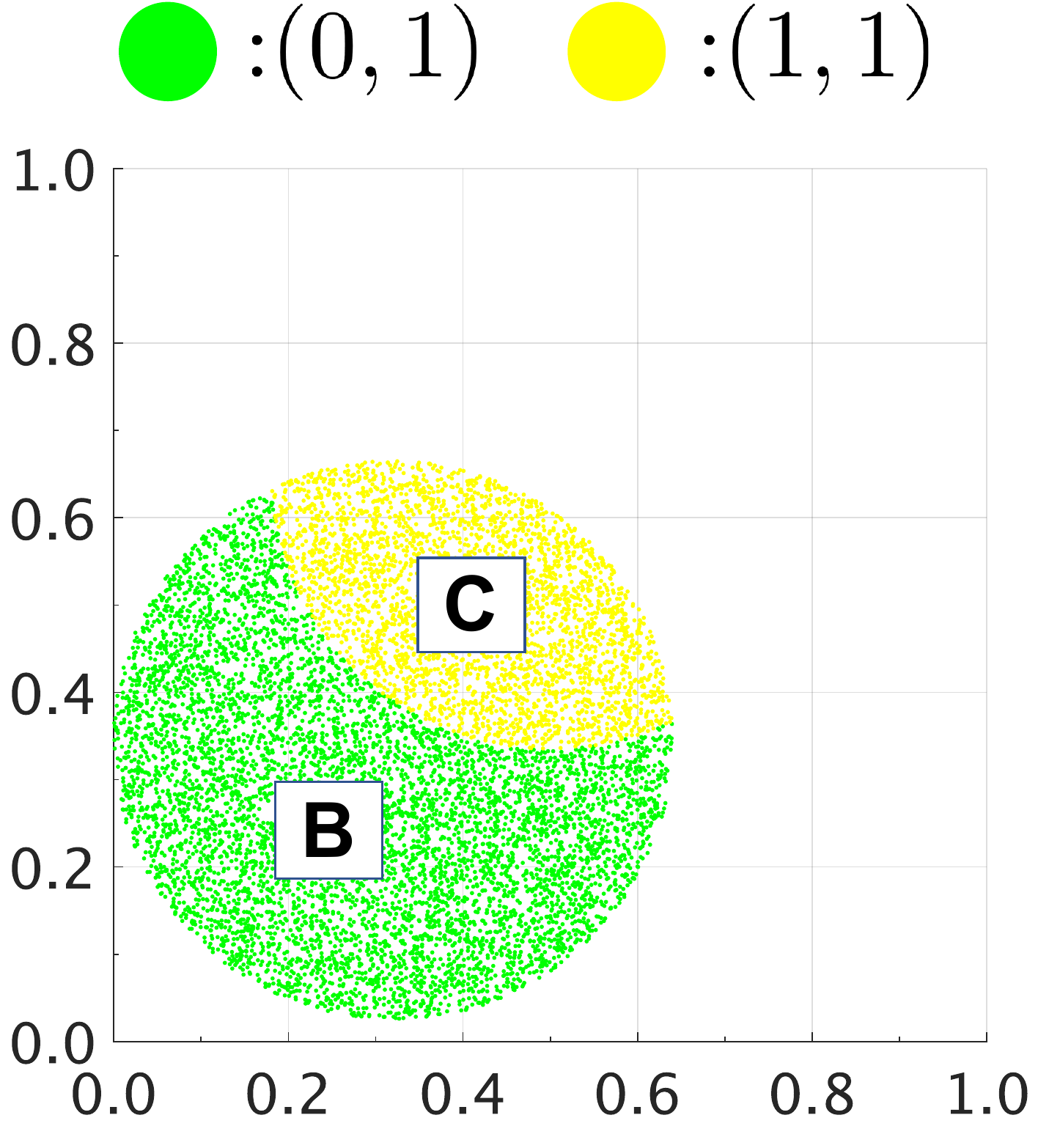}
			\label{fig:twoDimSynthetic_3dist_2}
		}
		\hfil
		\subfloat[Distribution \#3]{
			\includegraphics[height=1.05in]{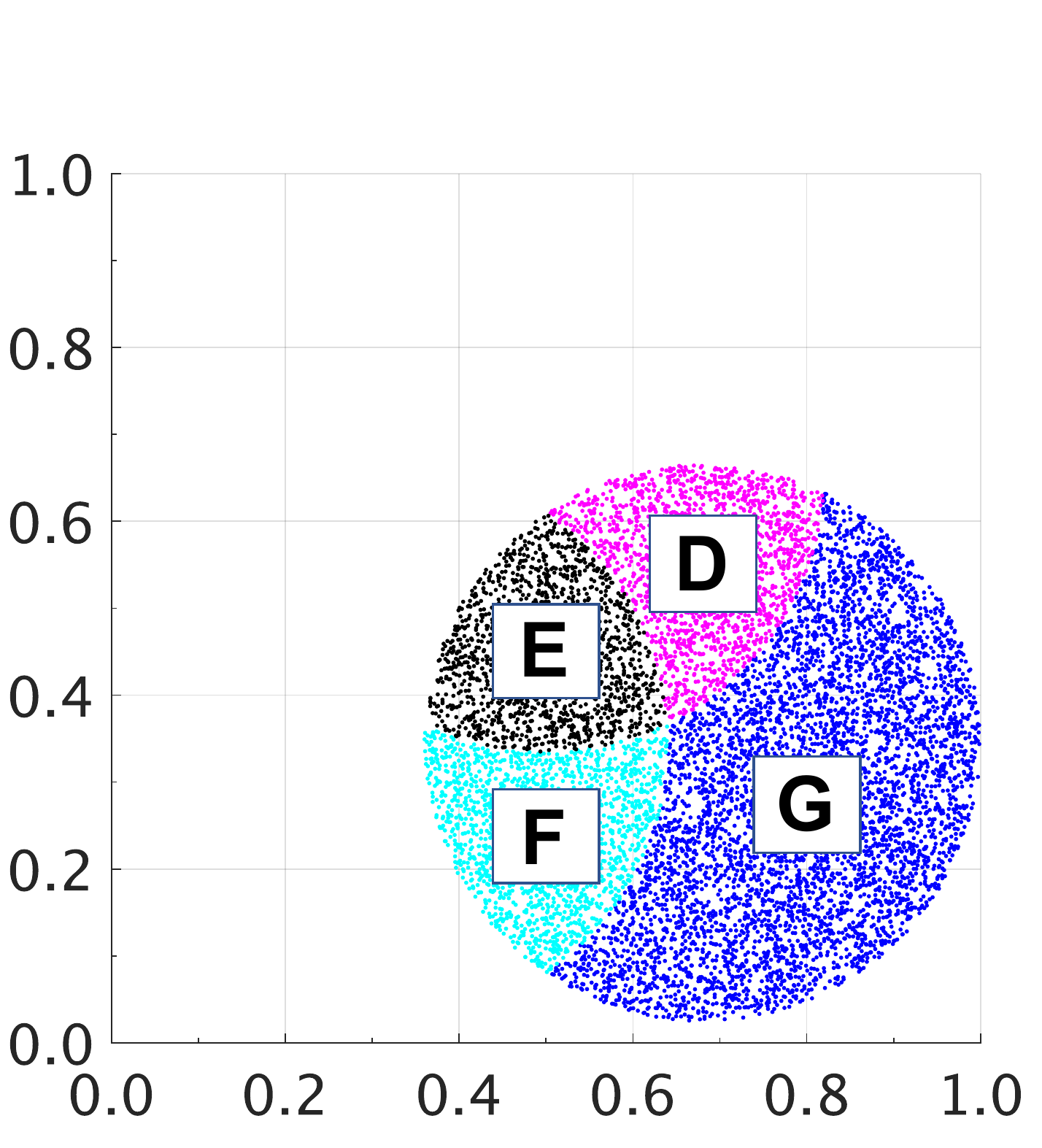}
			\label{fig:twoDimSynthetic_3dist_3}
		}
		\raisebox{6mm}{
			\includegraphics[height=0.5in]{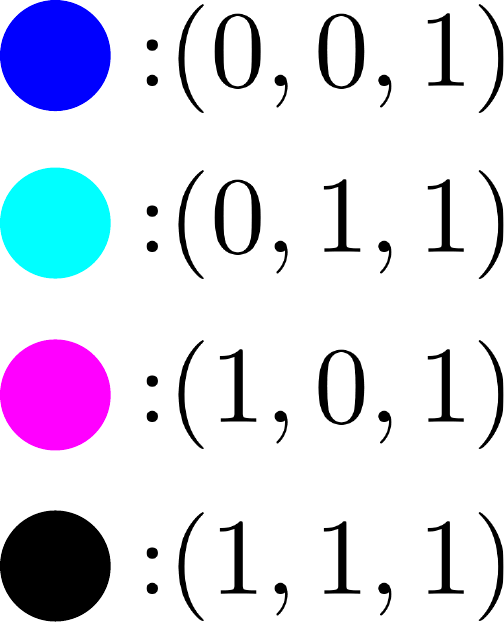}
		}
		\caption{Visualization of giving the three distributions in sequential order.}
		\label{fig:twoDimSynthetic_3dist}
	\end{minipage}
\end{figure*}

\begin{figure*}[htbp]
	\vspace{-1mm}
	\centering
	\subfloat[Distribution \#1]{
		\includegraphics[height=1.0in]{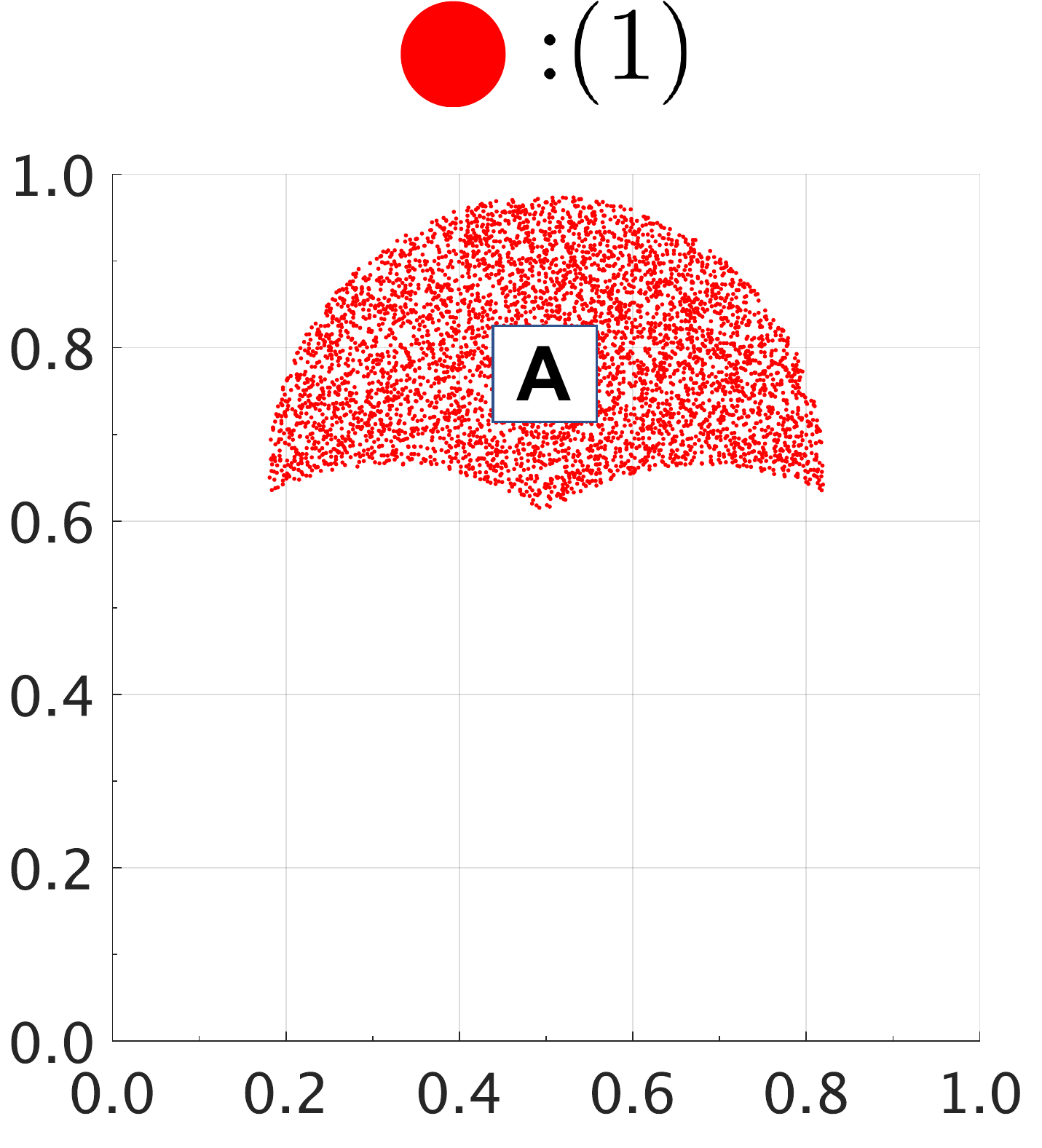}
		\label{fig:twoDimSynthetic_7dist_1}
	}
	\hfil
	\subfloat[Distribution \#2]{
		\includegraphics[height=1.0in]{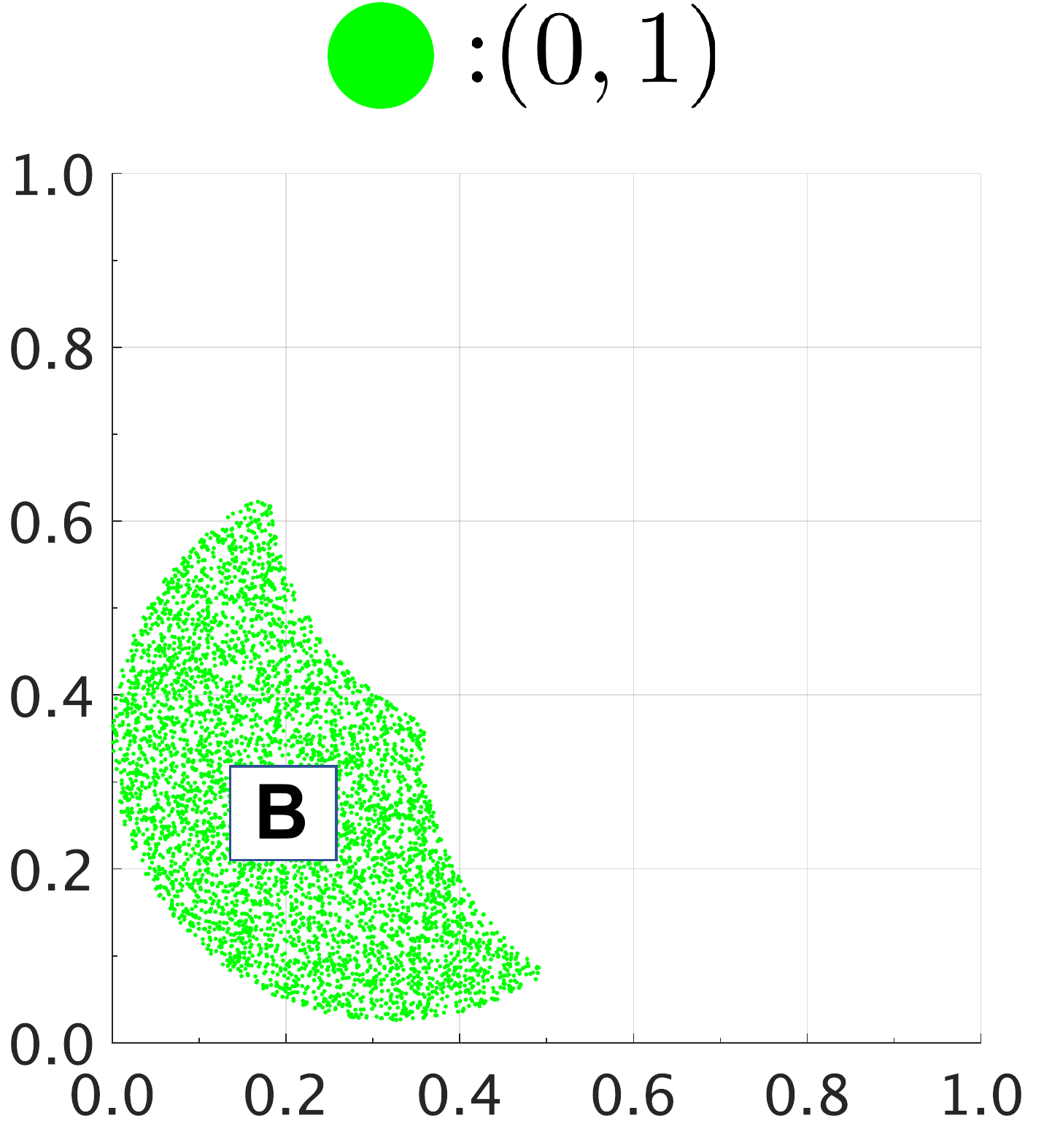}
		\label{fig:twoDimSynthetic_7dist_2}
	}
	\hfil
	\subfloat[Distribution \#3]{
		\includegraphics[height=1.0in]{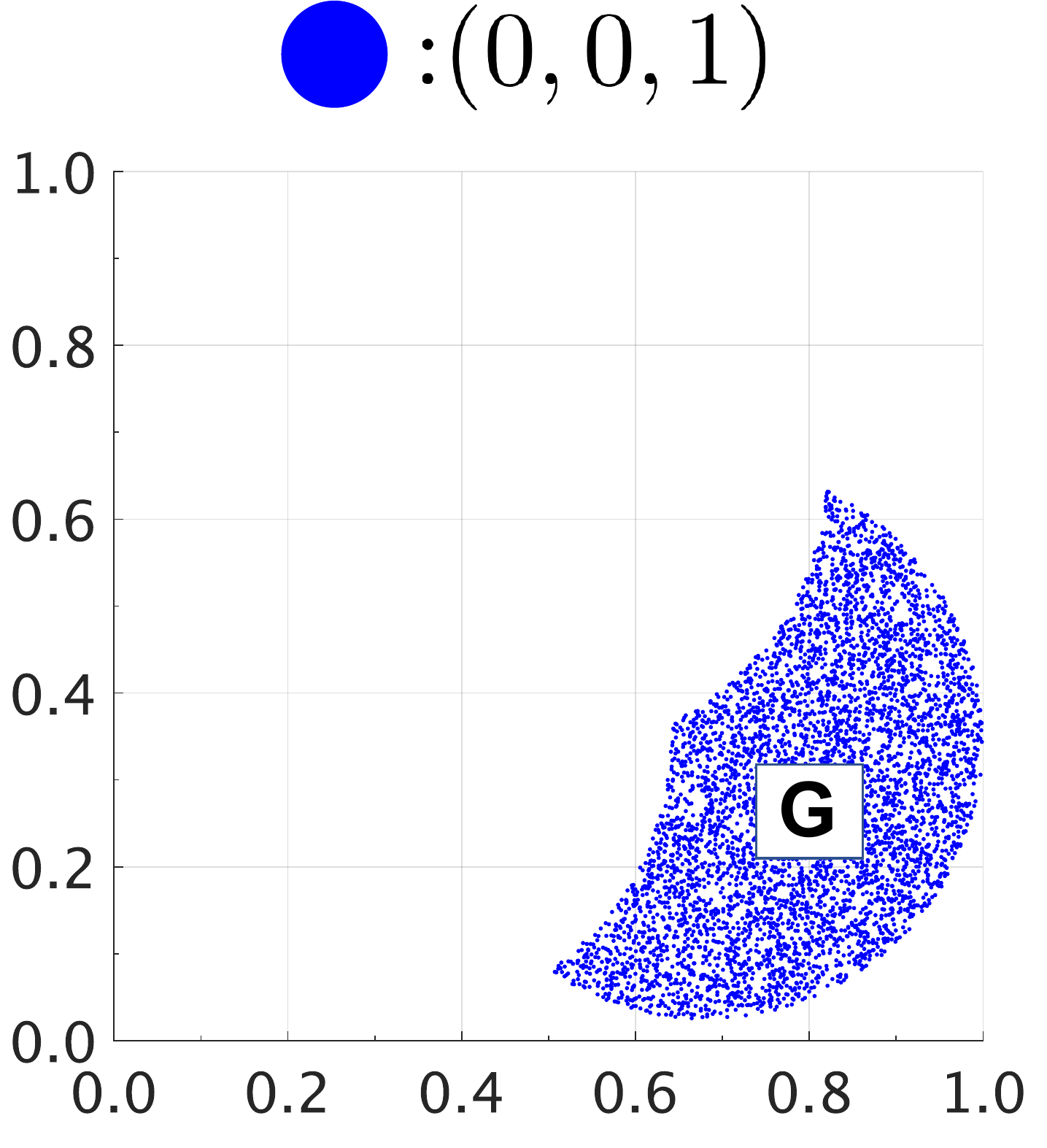}
		\label{fig:twoDimSynthetic_7dist_3}
	}
	\hfil
	\subfloat[Distribution \#4]{
		\includegraphics[height=1.0in]{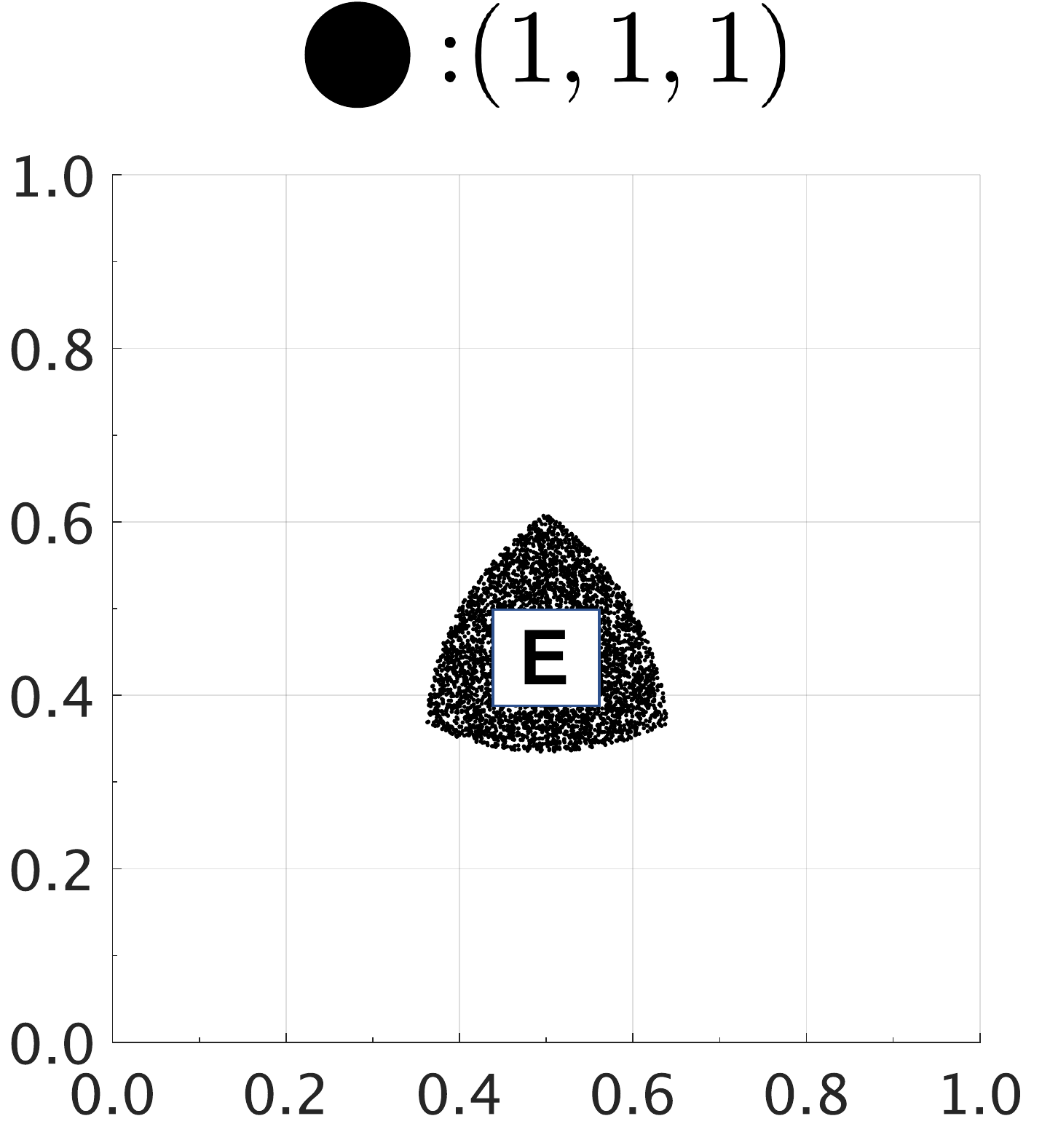}
		\label{fig:twoDimSynthetic_7dist_4}
	}
	\hfil
	\subfloat[Distribution \#5]{
		\includegraphics[height=1.0in]{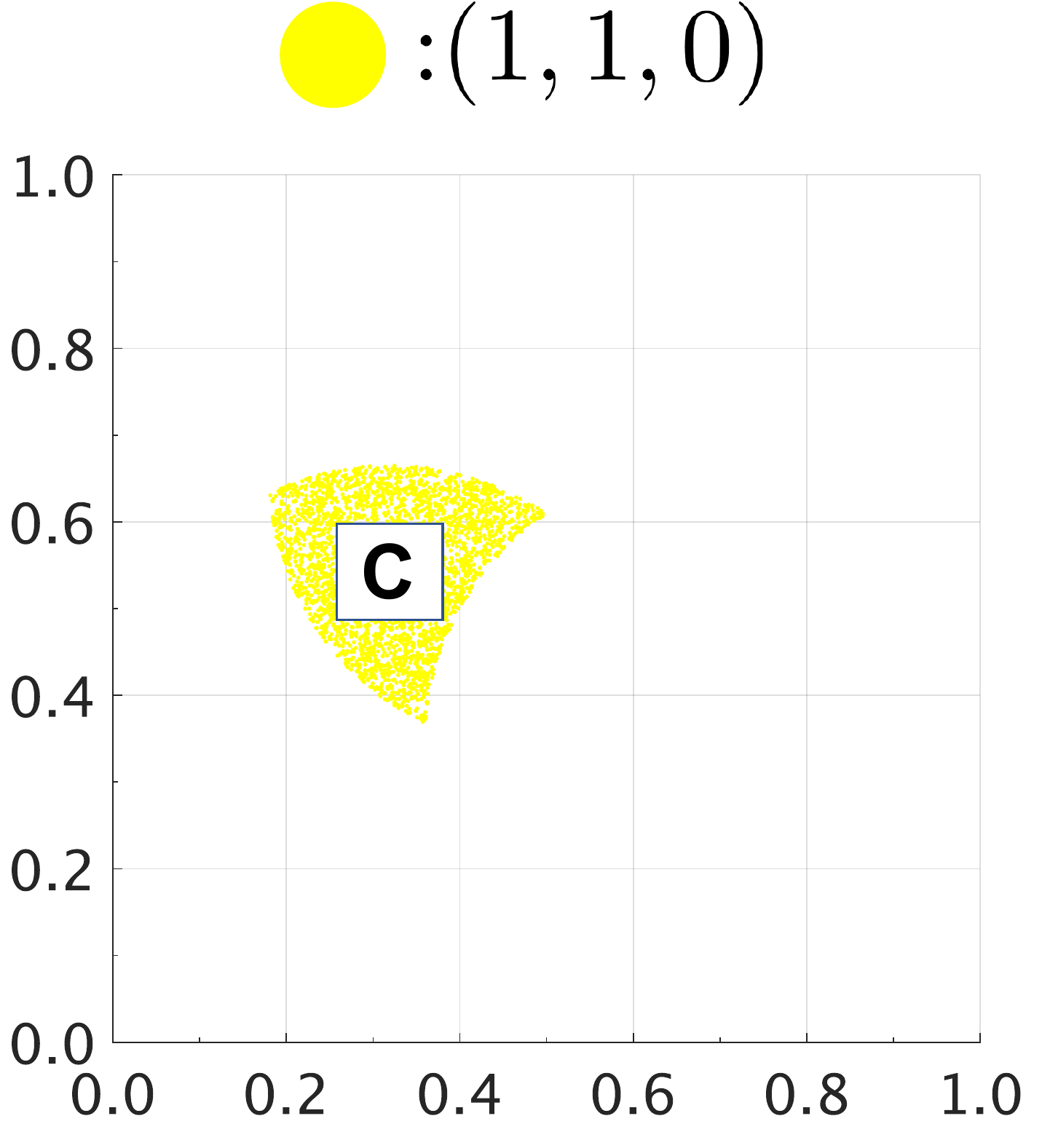}
		\label{fig:twoDimSynthetic_7dist_5}
	}
	\hfil
	\subfloat[Distribution \#6]{
		\includegraphics[height=1.0in]{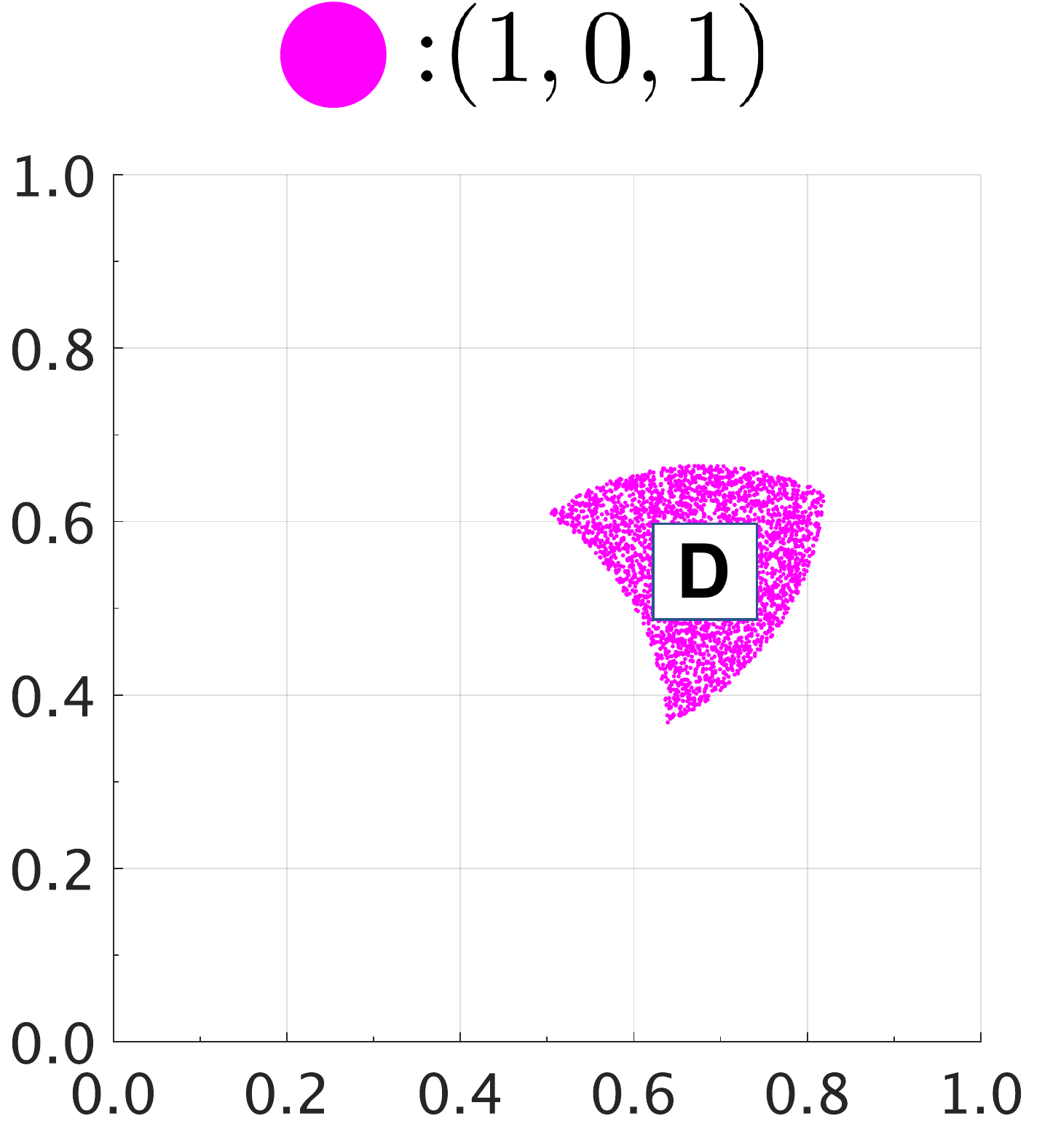}
		\label{fig:twoDimSynthetic_7dist_6}
	}
	\hfil
	\subfloat[Distribution \#7]{
		\includegraphics[height=1.0in]{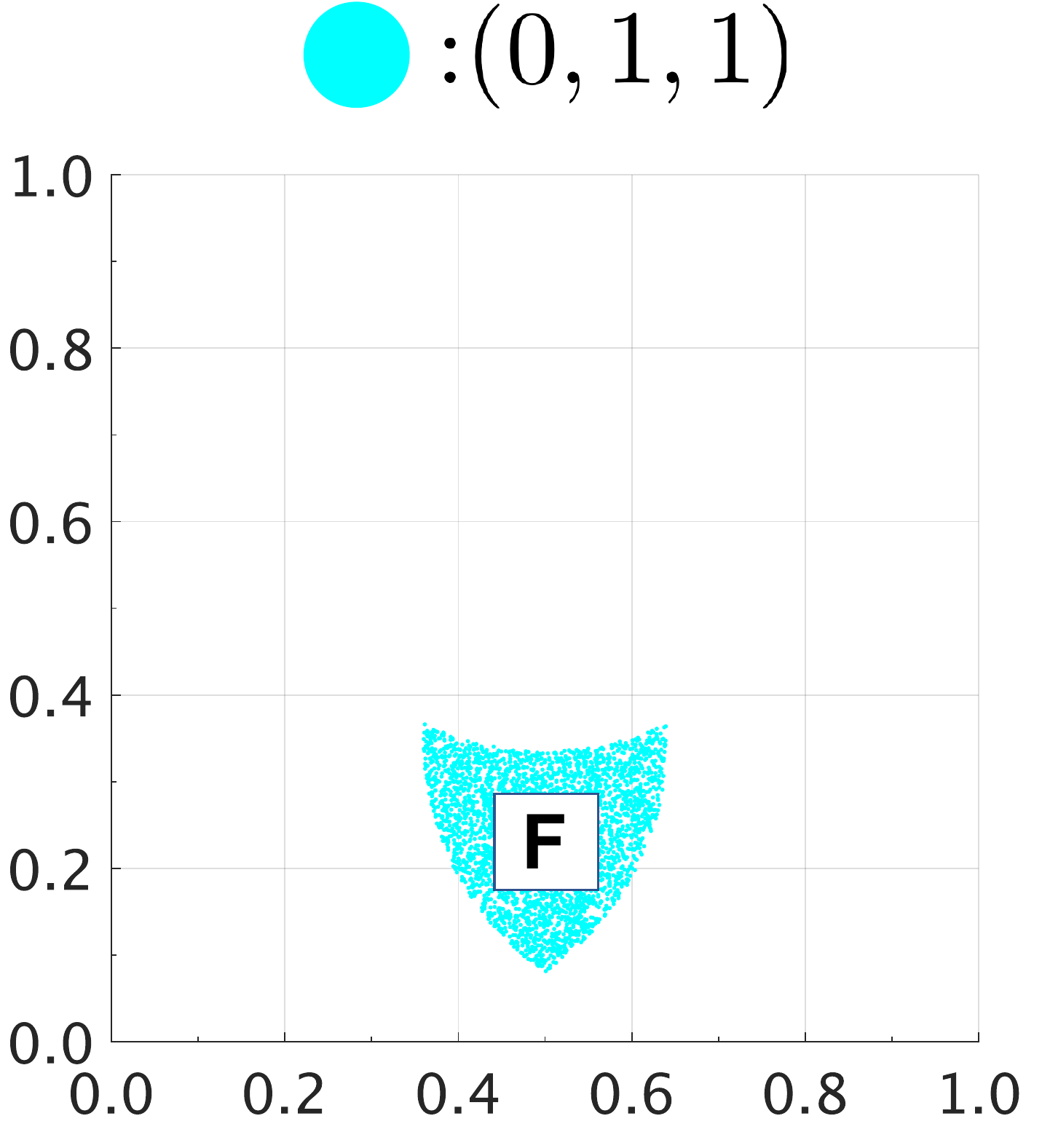}
		\label{fig:twoDimSynthetic_7dist_7}
	}
	\caption{Visualization of giving the seven distributions in sequential order.}
	\label{fig:twoDimSynthetic_7dist}
\end{figure*}

\begin{itemize}
	\setlength{\leftskip}{-3mm}
	\vspace{-2mm}
	\item \textbf{Exact Match}
	
	This is possibly the most strict performance metric. This metric measures whether a predicted set of labels for an instance is exactly equal to the true labels. The higher, the better.
	
	\vspace{1mm}
	\item \textbf{F$ _{1} $-score}
	
	This is a harmonic mean between the Precision and the Recall, namely, a weighted measure of how many true labels are predicted and how many predicted labels are truly relevant. The higher, the better.
	
	\vspace{1mm}
	\item \textbf{Macro-averaged AUC}
	
	This is the area under the receiver operating characteristic curve. The Macro-averaged AUC is the arithmetic mean of the AUC for each label. This metric gives a better sense of the performance across all labels. The higher, the better.
	
	\vspace{1mm}
	\item \textbf{Hamming Loss}
	
	This metric computes the symmetric difference between the predicted and true labels, and divided by the total number of labels in a dataset. The lower, the better.
	
	\vspace{1mm}
	\item \textbf{Ranking Loss}
	
	This metric computes how many times a relevant label (a member of the true labels) appears ranked lower than a non-relevant label, namely, the average proportion of label pairs that are incorrectly ordered for an instance. The lower, the better.
	
	\vspace{1mm}
	\item \textbf{Coverage}
	
	This metric is defined as the distance to cover all possible labels assigned to an instance, namely, how many top-scored predicted labels are included without missing any true labels. The lower, the better. In this paper, we scaled the Coverage by the number of labels $ N_{l} - 1 $ thus the range of the Coverage is [0, 1].
	
	%
	
\end{itemize}

\subsection{Continual Learning Ability}
\label{sec:continualLearning}
In theory, ART-based clustering is capable of learning new knowledge and preserving the learned knowledge without catastrophic forgetting by setting a fixed similarity threshold (i.e., a vigilance parameter in ART). Thus, MLCA can continually learn and preserve knowledge by adaptively generating nodes in response to changes of data distributions. Moreover, since MLCA has a fixed similarity threshold $ V $ and there is no node deletion process, MLCA does not inherently cause catastrophic forgetting.

In this section, we verify the continual learning ability of MLCA by using a two-dimensional synthetic multi-label dataset in the stationary and non-stationary environments. As a comparison algorithm, we apply MCIC \cite{nguyen19} which is capable continual learning through a clustering process. Similar to MLCA, MCIC extracts and accumulates information from data by nodes. Therefore, MCIC is a competitive algorithm for MLCA. Note that, as mentioned in Section \ref{sec:LR_multilabel}, MuENL \cite{zhu18a} can perform sequential learning and class-incremental learning, i.e., continual learning. However, MuENL cannot deal with the situation where new data distributions are provided in the non-stationary environment, and thus MuENL cannot cope with the experimental conditions in this section.

Fig. \ref{fig:twoDimSynthetic_1dist_dataset} shows the two-dimensional synthetic multi-label dataset. The dataset consists of three distributions where each has 10,000 instances that follow a uniform distribution. In addition, as shown in Fig. \ref{fig:twoDimSynthetic_1dist}, seven types of label sets are defined. In this experiment, the instances are given to each algorithm in three different conditions: (1) all the instances are given at the same time (Fig. \ref{fig:twoDimSynthetic_1dist_dataset}), (2) the three distributions are given in sequential order (Fig. \ref{fig:twoDimSynthetic_3dist}), and (3) the seven distributions are given in sequential order (Fig. \ref{fig:twoDimSynthetic_7dist}). The condition (1) is the stationary environment while the conditions (2) and (3) are the non-stationary environment. Here, the label information of each distribution is not changed. It should be noted that the given instances are presented as a sequence of three distributions in Fig. \ref{fig:twoDimSynthetic_3dist} whereas they are divided into seven distributions in Fig. \ref{fig:twoDimSynthetic_7dist}. As a result, Fig. \ref{fig:twoDimSynthetic_7dist} is an easier problem than Fig. \ref{fig:twoDimSynthetic_3dist} because there are no overlap regions.

\begin{table}[htbp]
	\centering
	\caption{Transition of label sets during sequentially giving the three distributions in Fig. \ref{fig:twoDimSynthetic_3dist}}
	\label{tab:syntheticLabel}
	\renewcommand{\arraystretch}{1.2}
	\begin{tabular}{lccccc}
		\hline\hline
		\multirow{2}{*}{Data} & \multirow{2}{*}{Subset} & \multicolumn{3}{c}{Transitions of Label Sets} \\
		&  & Dist. \#1 & Dist. \#2 & Dist. \#3  \\
		\hline
		Distribution \#1 & A & $ \bm{l}_{\text{A}} \!=\! (1) $  & $ \bm{l}_{\text{A}} \!=\! (1,0) $ & $ \bm{l}_{\text{A}} \!=\! (1,0,0) $ \\
		\hline
		Distribution \#2 & B & \multirow{2}{*}{---} & $ \bm{l}_{\text{B}} \!=\! (0,1) $ & $ \bm{l}_{\text{B}} \!=\! (0,1,0) $ \\
		&  C & & $ \bm{l}_{\text{C}} \!=\! (1, 1) $ & $ \bm{l}_{\text{C}} \!=\! (1, 1, 0) $ \\
		\hline
		Distribution \#3 & D &  \multirow{4}{*}{---} &  \multirow{4}{*}{---} & $ \bm{l}_{\text{D}} \!=\! (1,0,1) $ \\
		& E &  &  & $ \bm{l}_{\text{E}}  \!=\! (1,1,1) $ \\
		& F &  &  & $ \bm{l}_{\text{F}} \!=\! (0,1,1) $ \\
		& G &  &  & $ \bm{l}_{\text{G}} \!=\! (0,0,1) $ \\
		\hline\hline
	\end{tabular}
\end{table}

In the case that the three distributions are given in sequential order (Fig. \ref{fig:twoDimSynthetic_3dist}), the instances in the overlapping regions are designed to contain label information of the already given distribution (e.g., $ \bm{l}_{\text{C}} = (1,1) $) for generating a situation where the number of labels increases in a pseudo manner. Thus, as shown in Table \ref{tab:syntheticLabel}, seven types of label sets are defined after the three distributions are given. As the number of labels increases, the length of the label set also changes incrementally, e.g., $ \bm{l}_{\text{A}} \!=\! (1) \rightarrow \bm{l}_{\text{A}} \!=\! (1, 0) \rightarrow \bm{l}_{\text{A}} \!=\! (1, 0, 0) $. The similar label transition is occurred in the case of Fig. \ref{fig:twoDimSynthetic_7dist}.

\begin{table*}[htbp]
	\vspace{2mm}
	\centering
	\caption{Results of classification performance for continual learning under environments in Figs. \ref{fig:twoDimSynthetic_1dist_dataset}, \ref{fig:twoDimSynthetic_3dist}, and \ref{fig:twoDimSynthetic_7dist}}
	\label{tab:resultQualitativeCL}
	\renewcommand{\arraystretch}{1.2}
	\scalebox{0.9}{
		\begin{tabular}{lllccccccc}
			\hline\hline
			Algorithm & Parameter & Metric & Dist. \#1 & Dist. \#2 & Dist. \#3 & Dist. \#4 & Dist. \#5 & Dist. \#6 & Dist. \#7 \\
			\hline
			MLCA & $ V = 0.1 $ & Exact Match & 0.972 (0.001) &  &  &  &  &  &  \\
						&  & Hamming Loss & 0.010 (0.000) & --- & --- & --- & --- & --- & --- \\
						&  & Number of Nodes & 868.7 (16.6) &  &  &  &  &  &  \\
						\cline{2-10}
						& $ V = 0.1 $ & Exact Match & 1.000 (0.000) & 0.958 (0.001) & 0.909 (0.001) &  &  &  &  \\
						&  & Hamming Loss & 0.000 (0.000) & 0.014 (0.000) & 0.031 (0.001) & --- & --- & --- & --- \\
						&  & Number of Nodes & 703.3 (10.6) & 1183.5 (14.0) & 1552.4 (17.0) &  &  &  &  \\
						\cline{2-10}
						& $ V = 0.1 $ & Exact Match & 1.000 (0.000) & 0.999 (0.000) & 0.997 (0.000) & 0.994 (0.001) & 0.957 (0.002) & 0.965 (0.004) & 0.930 (0.008) \\
						&  & Hamming Loss & 0.000 (0.000) & 0.000 (0.000) & 0.001 (0.000) & 0.003 (0.000) & 0.015 (0.001) & 0.012 (0.002) & 0.024 (0.003) \\
						&  & Number of Nodes & 595.4 (13.2) & 1165.5 (17.1) & 1741.0 (19.7) & 1954.1 (21.8) & 2169.6 (20.4) & 2407.7 (21.8) & 2658.5 (24.4) \\
			\hline
			MCIC & $ \epsilon = 0.011 $ & Exact Match & 0.342 (0.039) &  &  &  &  &  &  \\
						&  & Hamming Loss & 0.222 (0.002) & --- & --- & --- & --- & --- & --- \\
						&  & Number of Nodes & 804.6 (9.3) &  &  &  &  &  &  \\
						\cline{2-10}
						& $ \epsilon = 0.007 $ & Exact Match & 1.000 (0.000) & 0.799 (0.001) & 0.392 (0.230) &  &  &  &  \\
						&  & Hamming Loss & 0.000 (0.000) & 0.068 (0.000) & 0.212 (0.065) & --- & --- & --- & --- \\
						&  & Number of Nodes & 770.7 (8.7) & 1320.9 (7.8) & 1749.3 (16.8) &  &  &  &  \\
						\cline{2-10}
						& $ \epsilon = 0.005 $ & Exact Match & 1.000 (0.000) & 1.000 (0.000) & 1.000 (0.000) & 0.234 (0.000) & 0.143 (0.000) & 0.249 (0.000) & 0.334 (0.000) \\
						&  & Hamming Loss & 0.000 (0.000) & 0.000 (0.000) & 0.000 (0.000) & 0.511 (0.000) & 0.286 (0.000) & 0.250 (0.000) & 0.222 (0.000) \\
						&  & Number of Nodes & 620.0 (8.1) & 1236.1 (11.2) & 1865.4 (13.2) & 2133.4 (15.1) & 2377.3 (17.5) & 2625.7 (17.0) & 2869.2 (16.6) \\
			\hline\hline
		\end{tabular}
		}
	\\
	\vspace{1mm}
	\footnotesize \raggedright \hspace{3mm}The standard deviation is indicated in parentheses.
	\vspace{-3mm}
\end{table*}

In order to analyze the continual learning capability, we evaluate the classification performance after each distribution is given. After learning the training instances of each distribution, the instances belonging to the learned distribution is used as test data. Namely, I) after learning the distribution \#1, the classification performance is evaluated by using the test data of distribution \#1. Next, II) after learning the distribution \#2, the classification performance is evaluated by using the test data of the distributions \#1 and \#2. Then, III) after learning the distribution \#3, the classification performance is evaluated by using the test data of the distributions \#1, \#2, and \#3. We continue this procedure until all the distributions are given. We repeat the experiment 20 times with a different random seed to obtain consistent results. The parameters of MLCA are specified as follows: $ N_ {y} $ = 10, $ \lambda = 50 $, and $ V = 0.10 $. The parameters of MCIC are specified as follows: $ \delta = 0.1 $, $ K  = 3 $, $ \lambda = 0.25 $, $ \beta_{\mu} = 2 $, $ \epsilon = \left\{ 0.011, 0.007, 0.005 \right\} $, and a processing speed is 10,000. Under the above parameter settings, MLCA and MCIC generate the similar number of nodes.

Table \ref{tab:resultQualitativeCL} shows classification performance after training the distributions in Figs. \ref{fig:twoDimSynthetic_1dist_dataset}, \ref{fig:twoDimSynthetic_3dist}, and \ref{fig:twoDimSynthetic_7dist}. Focusing on MLCA, since high classification performance is maintained in the stationary and non-stationary environments, it is quantitatively shown that MLCA is capable of continual learning with a multi-label dataset. It is noteworthy that MLCA effectively accumulates the knowledge for the classification by using only about 900 to 2,700 nodes depending on the environment even if the distributions contains 30,000 instances in total. In contrast, the classification performance of MCIC is clearly inferior to MLCA even if the number of nodes is similar. 

\begin{figure}[!t]
	\vspace{-2mm}
	\centering
	\subfloat[MLCA]{
		\includegraphics[height=1.05in]{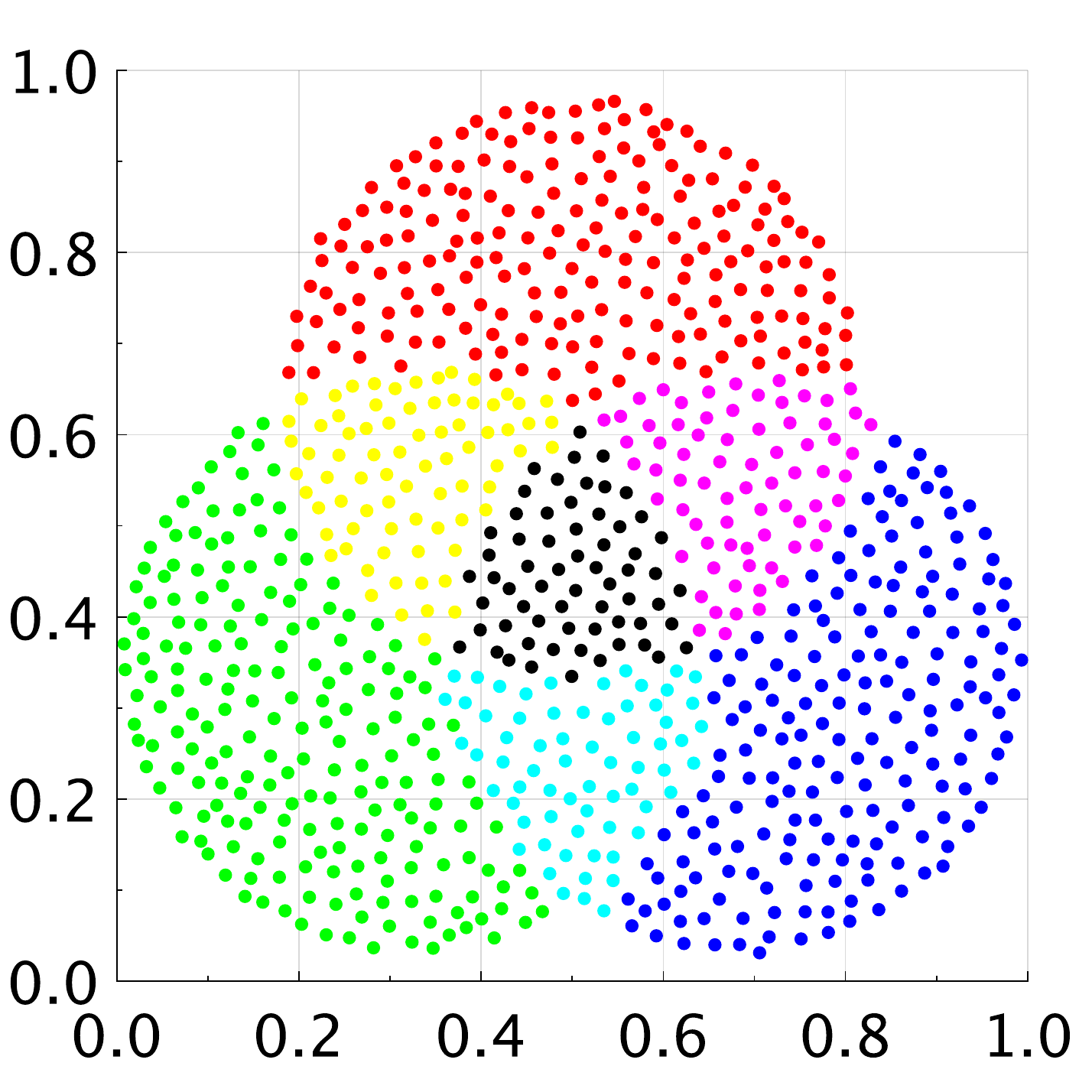}
		\label{fig:MLCA_1dist}
	}
	\hfil
	\subfloat[MCIC]{
		\includegraphics[height=1.05in]{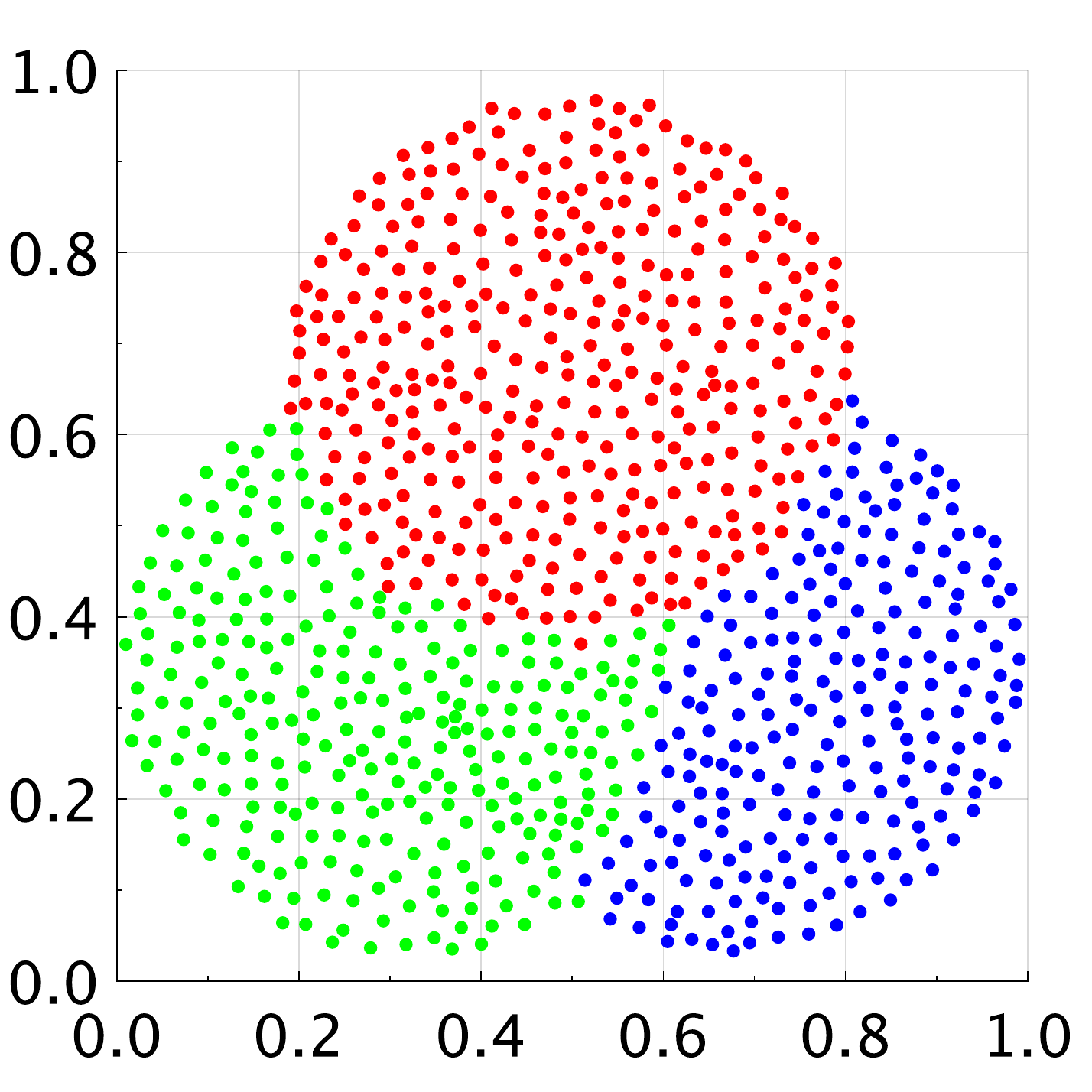}
		\label{fig:MCIC_1dist}
	}
	\raisebox{-0.1mm}{
		\includegraphics[height=1.07in]{figures/twoDimSynthetic_7dist_Legend.pdf}
	}
	\caption{Visualization of self-organizing results in the case the three distributions are given at the same time.}
	\label{fig:MLCA_MCIC_1dist}
\end{figure}

\begin{figure}[!t]
	\vspace{-2mm}
	\centering
	\hspace{-2mm}
	\subfloat[Distribution \#1]{
		\includegraphics[height=1.05in]{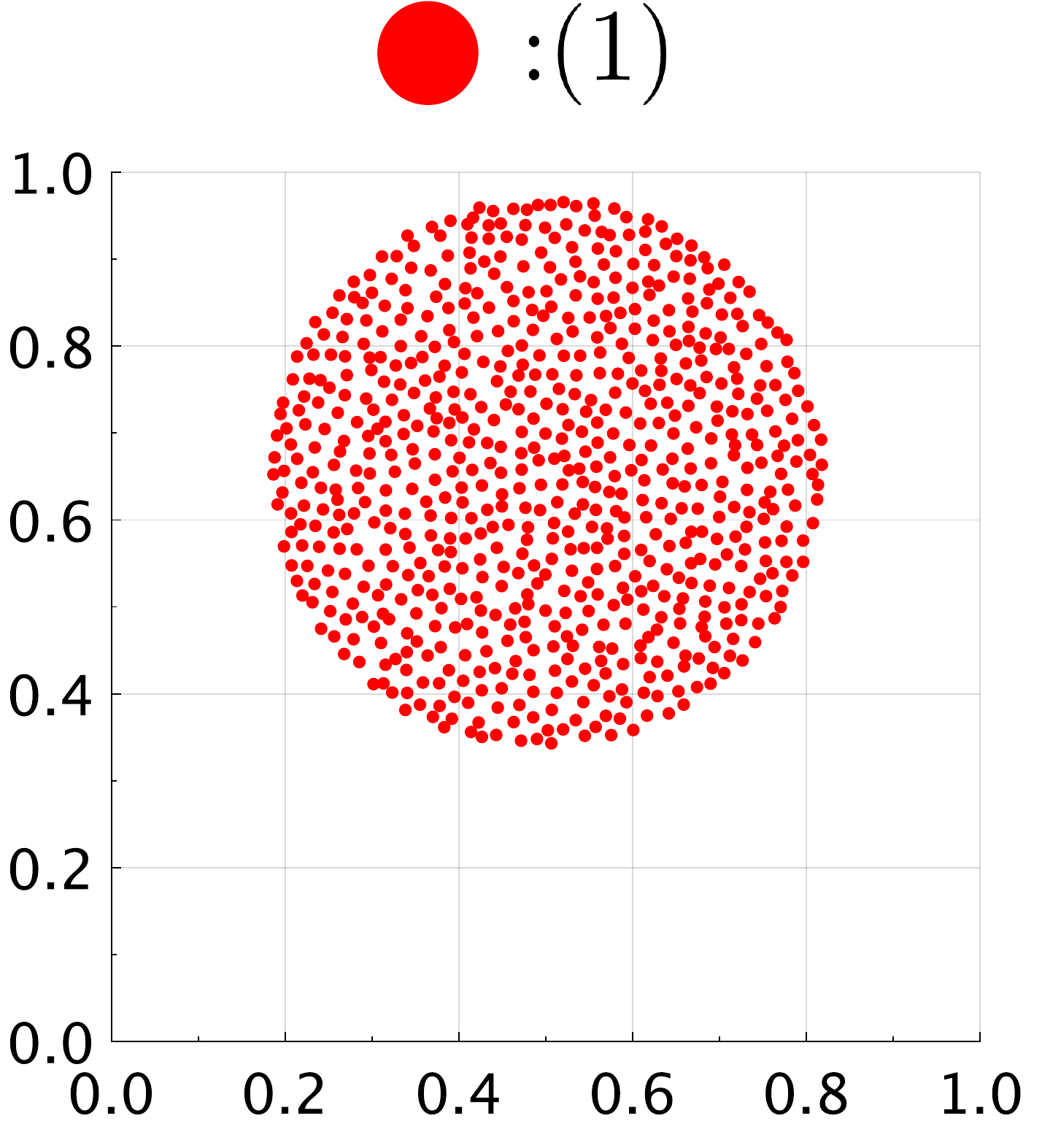}
		\label{fig:MLCA_3dist_1}
	}
	\hspace{-1mm}
	\hfil
	\subfloat[Distribution \#2]{
		\includegraphics[height=1.05in]{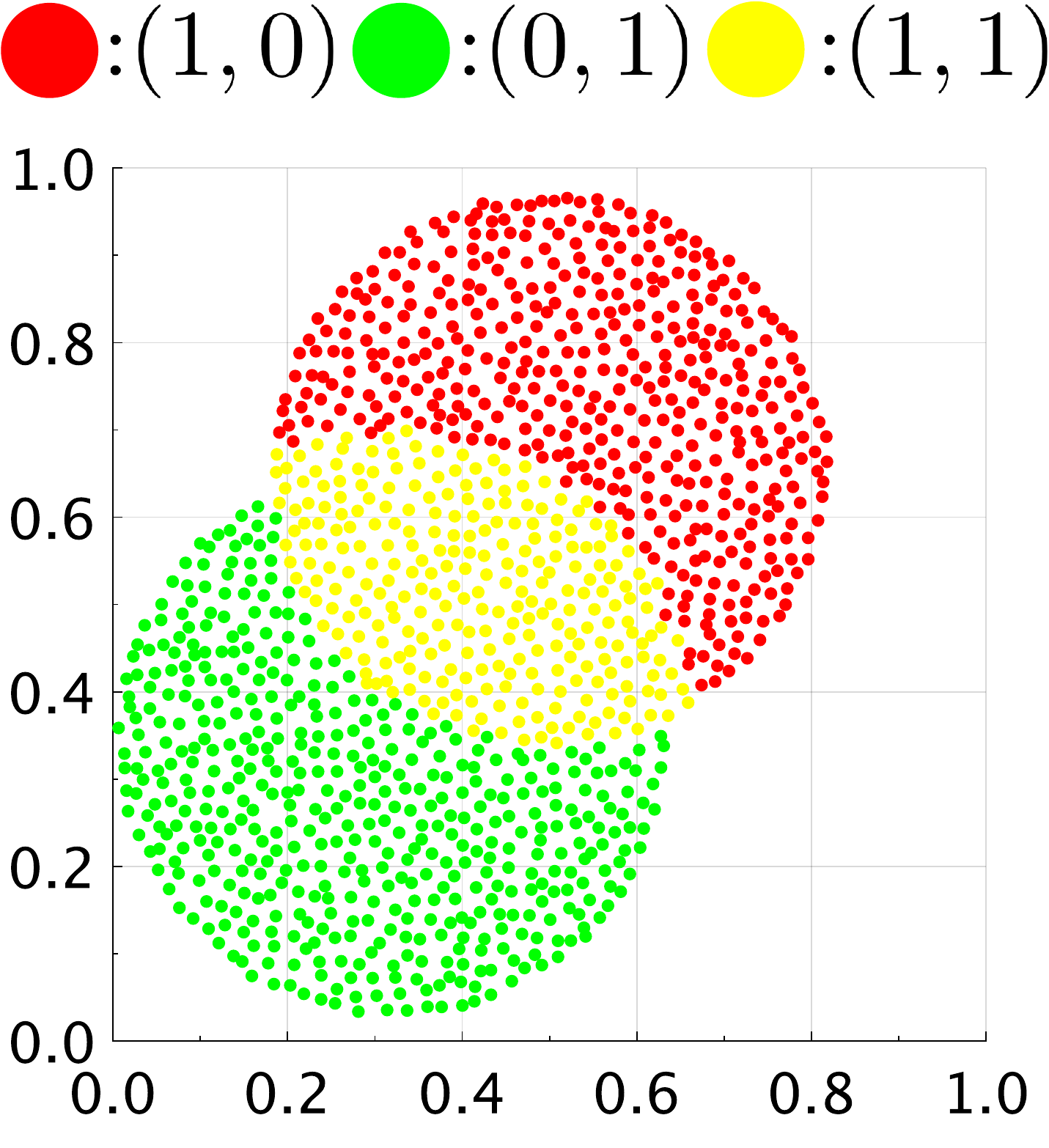}
		\label{fig:MLCA_3dist_2}
	}
	\hspace{-1mm}
	\hfil
	\subfloat[Distribution \#3]{
		\includegraphics[height=1.05in]{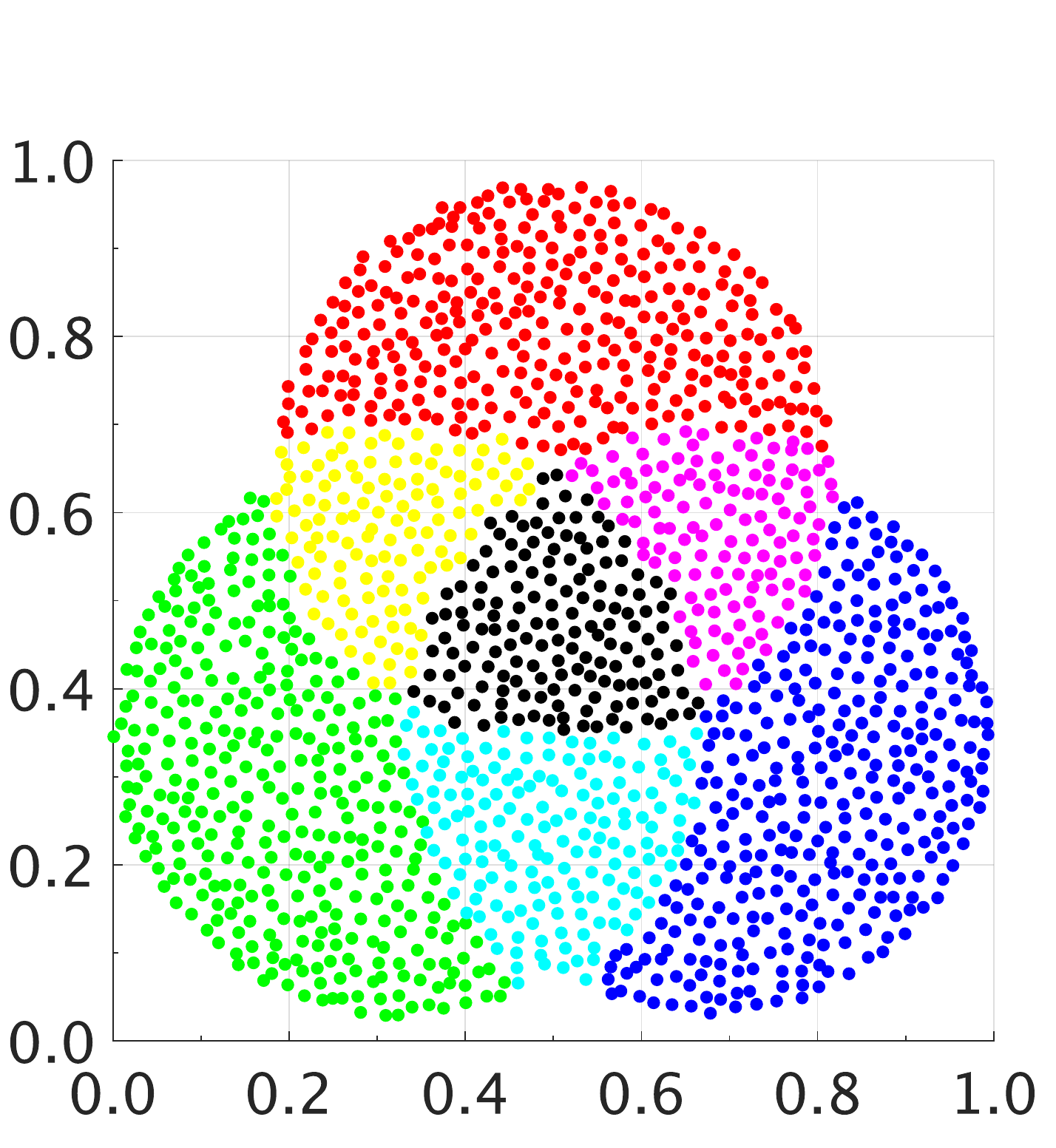}
		\label{fig:MLCA_3dist_3}
	}
	\hspace{-2mm}
	\raisebox{1mm}{
		\includegraphics[height=0.9in]{figures/twoDimSynthetic_7dist_Legend.pdf}
	}
	\caption{Visualization of nodes in MLCA in the case the three distributions are given in sequential order.}
	\label{fig:MLCA_3dist}
\end{figure}

\begin{figure}[!t]
	\vspace{-2mm}
	\centering
	\hspace{-2mm}
	\subfloat[Distribution \#1]{
		\includegraphics[height=1.05in]{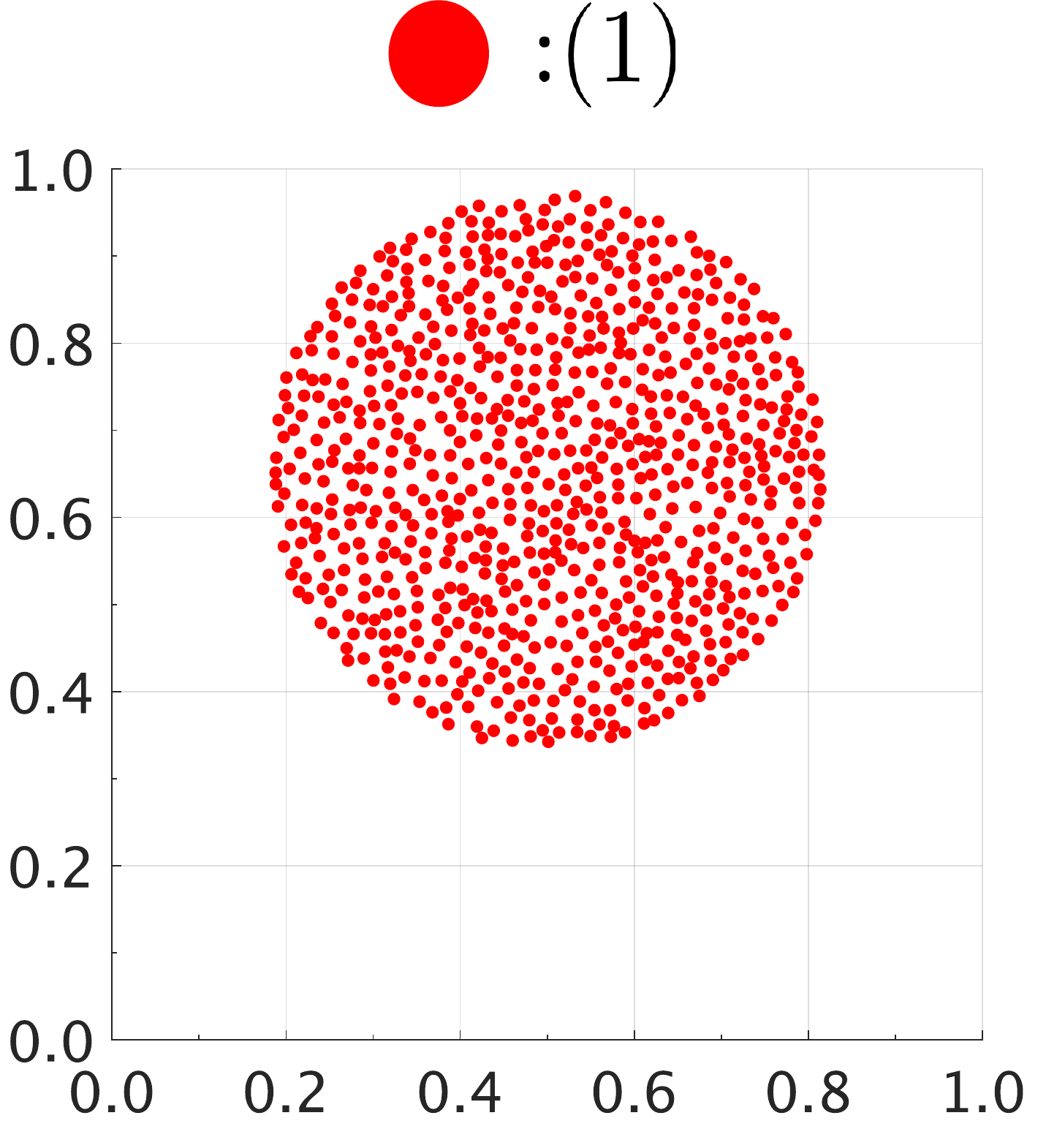}
		\label{fig:MCIC_3dist_1}
	}
	\hspace{-1mm}
	\hfil
	\subfloat[Distribution \#2]{
		\includegraphics[height=1.05in]{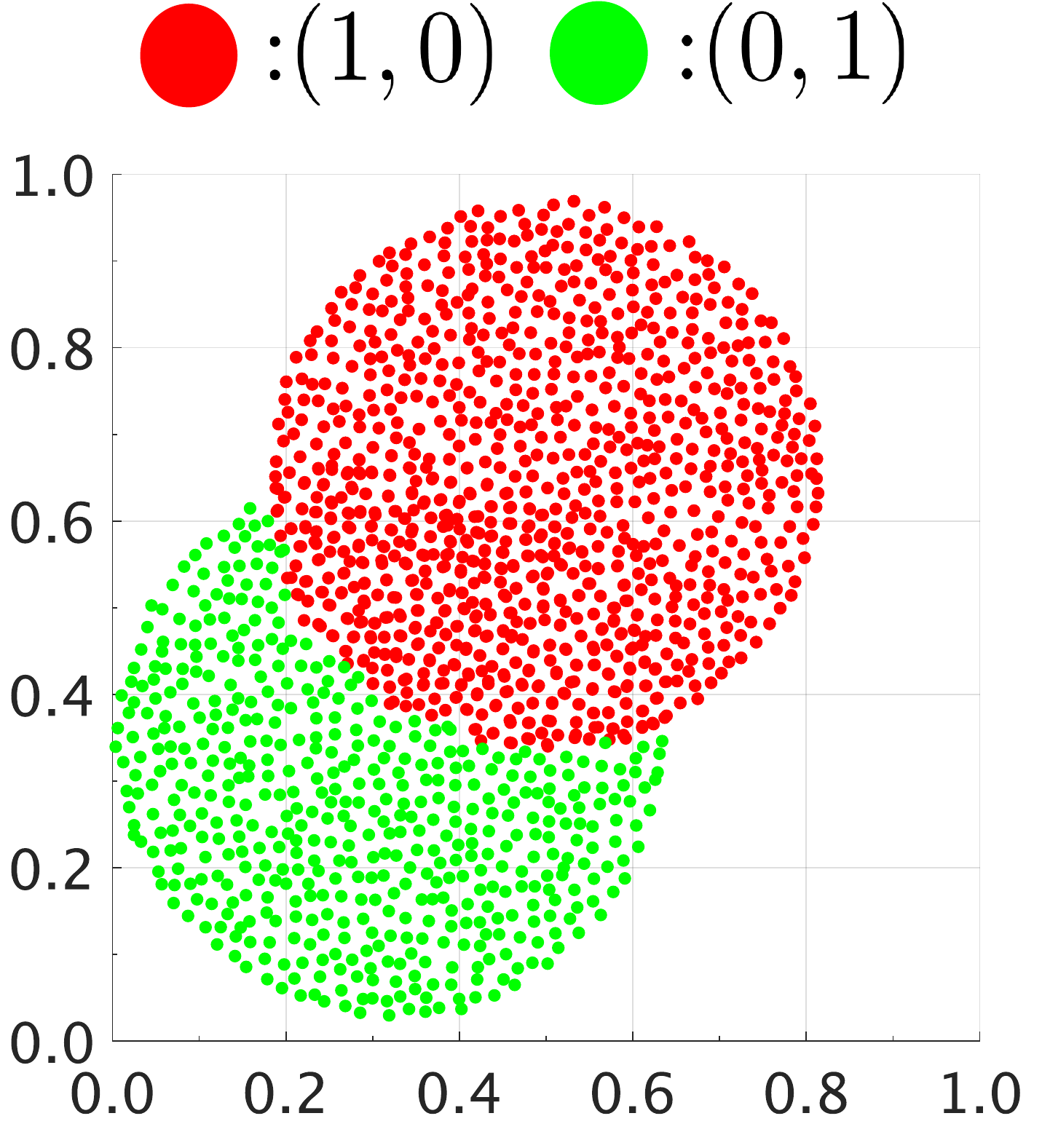}
		\label{fig:MCIC_3dist_2}
	}
	\hspace{-1mm}
	\hfil
	\subfloat[Distribution \#3]{
		\includegraphics[height=1.05in]{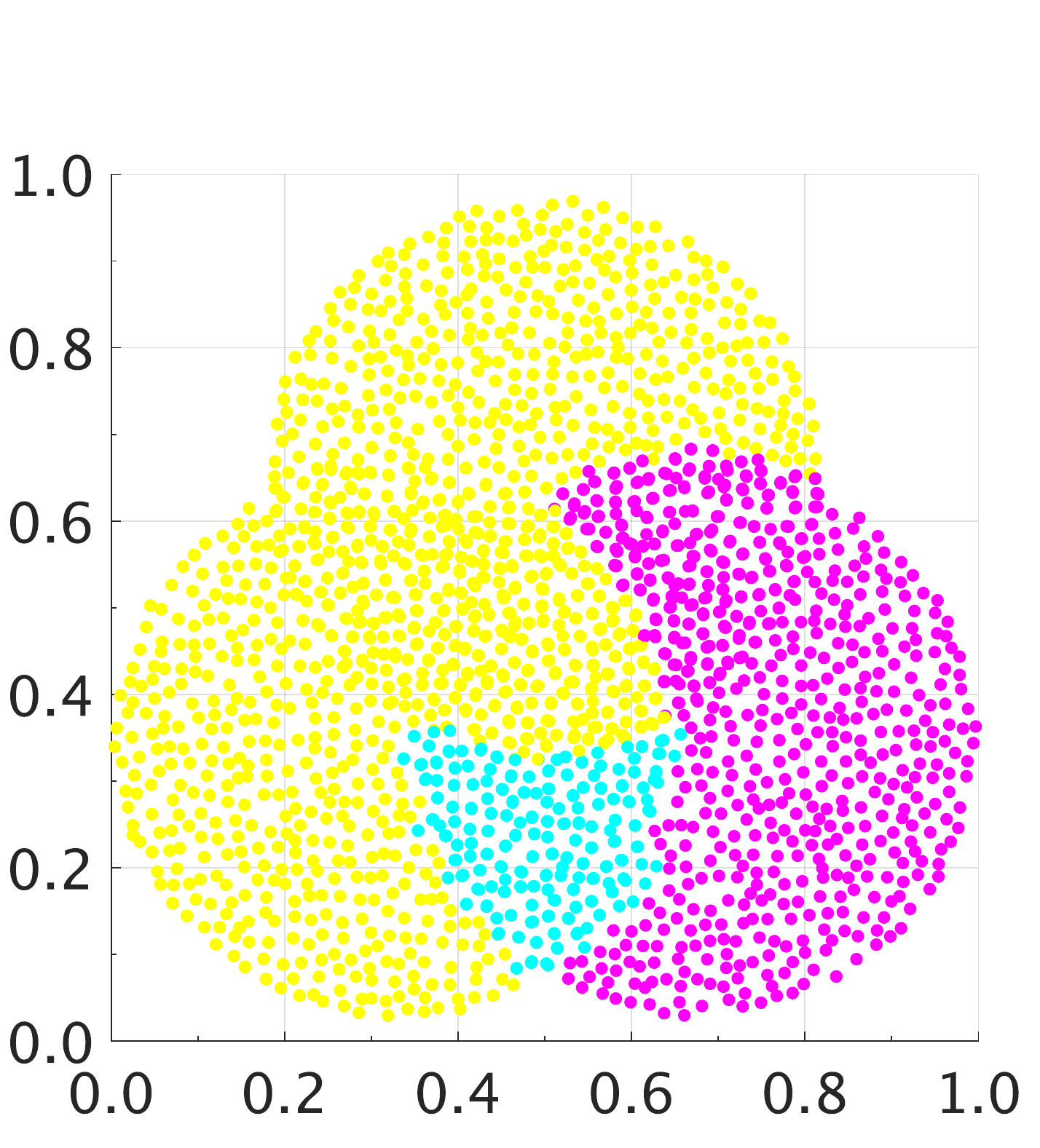}
		\label{fig:MCIC_3dist_3}
	}
	\hspace{-2mm}
	\raisebox{7mm}{
		\includegraphics[height=0.39in]{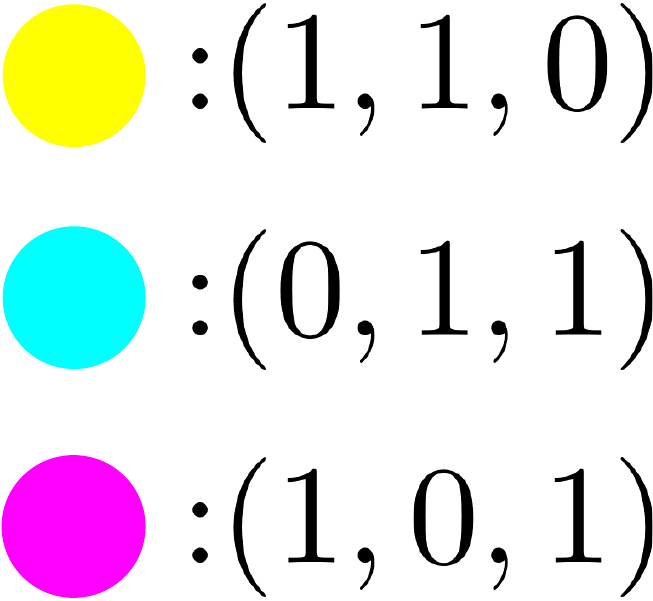}
	}
	\caption{Visualization of nodes in MCIC in the case the three distributions are given in sequential order.}
	\label{fig:MCIC_3dist}
\end{figure}

\begin{figure*}[htbp]
	\vspace{-3mm}
	\centering
	\subfloat[Distribution \#1]{
		\includegraphics[height=1.0in]{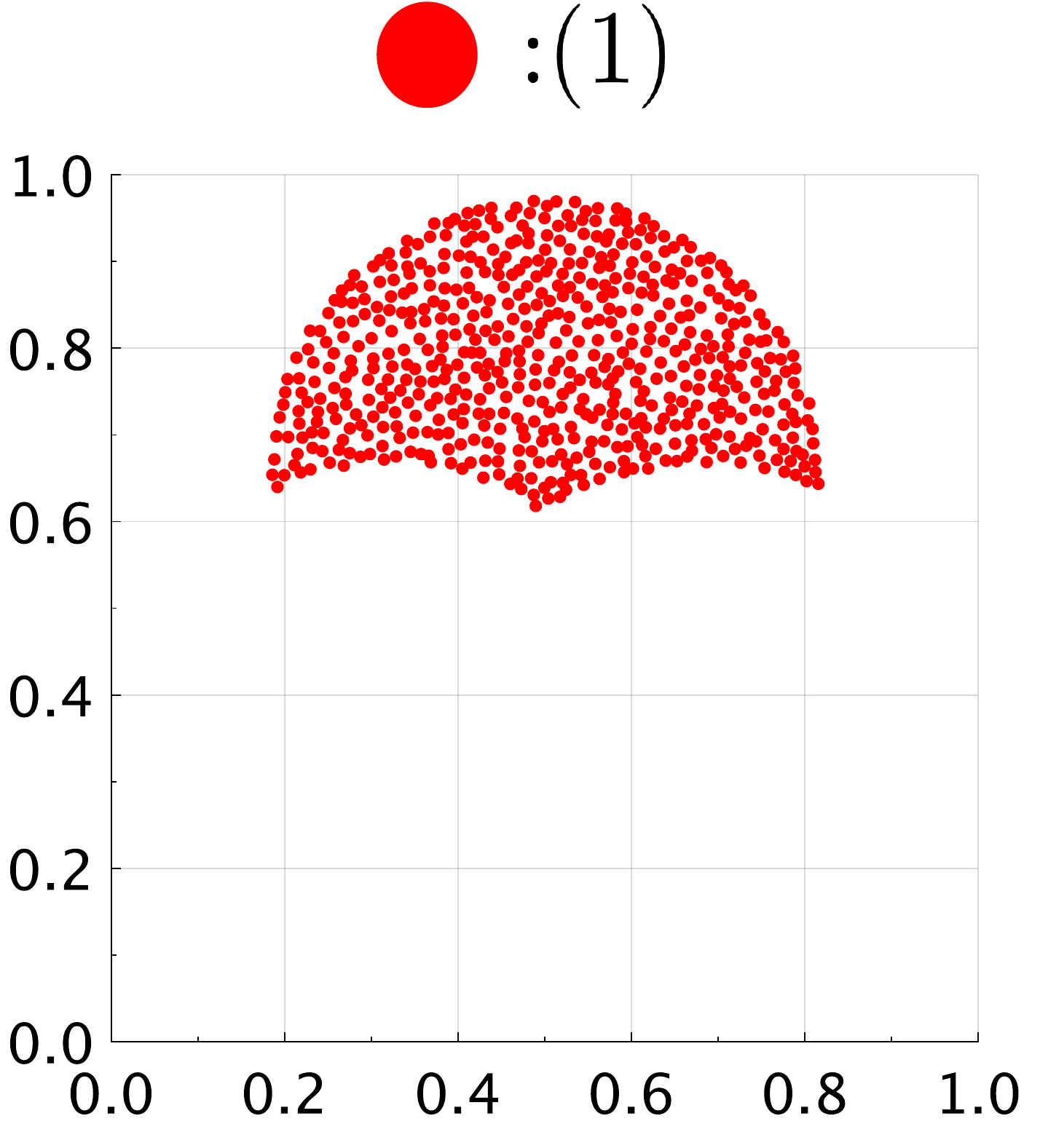}
		\label{fig:MLCA_7dist_1}
	}
	\hfil
	\subfloat[Distribution \#2]{
		\includegraphics[height=1.0in]{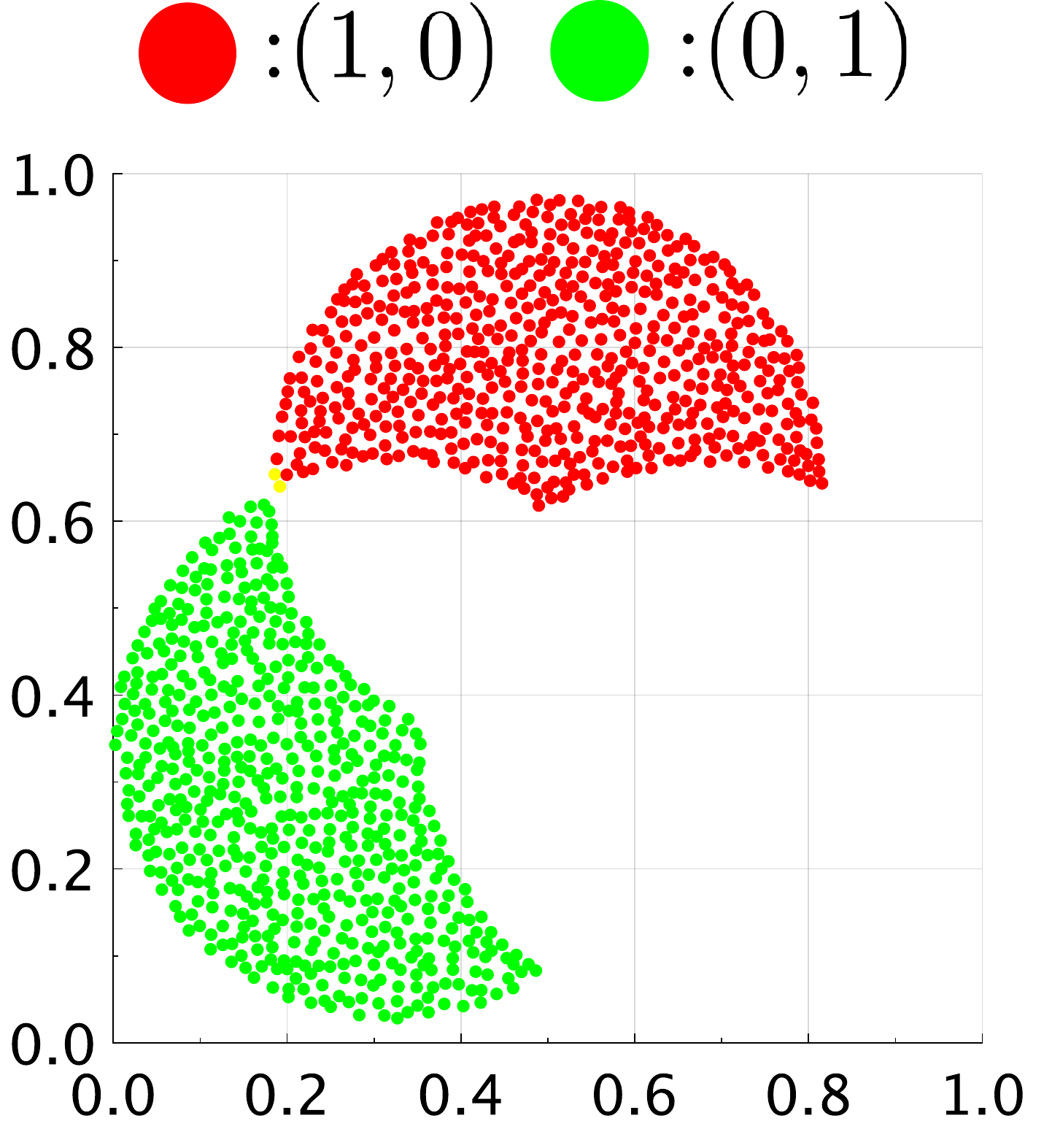}
		\label{fig:MLCA_7dist_2}
	}
	\hfil
	\subfloat[Distribution \#3]{
		\includegraphics[height=1.0in]{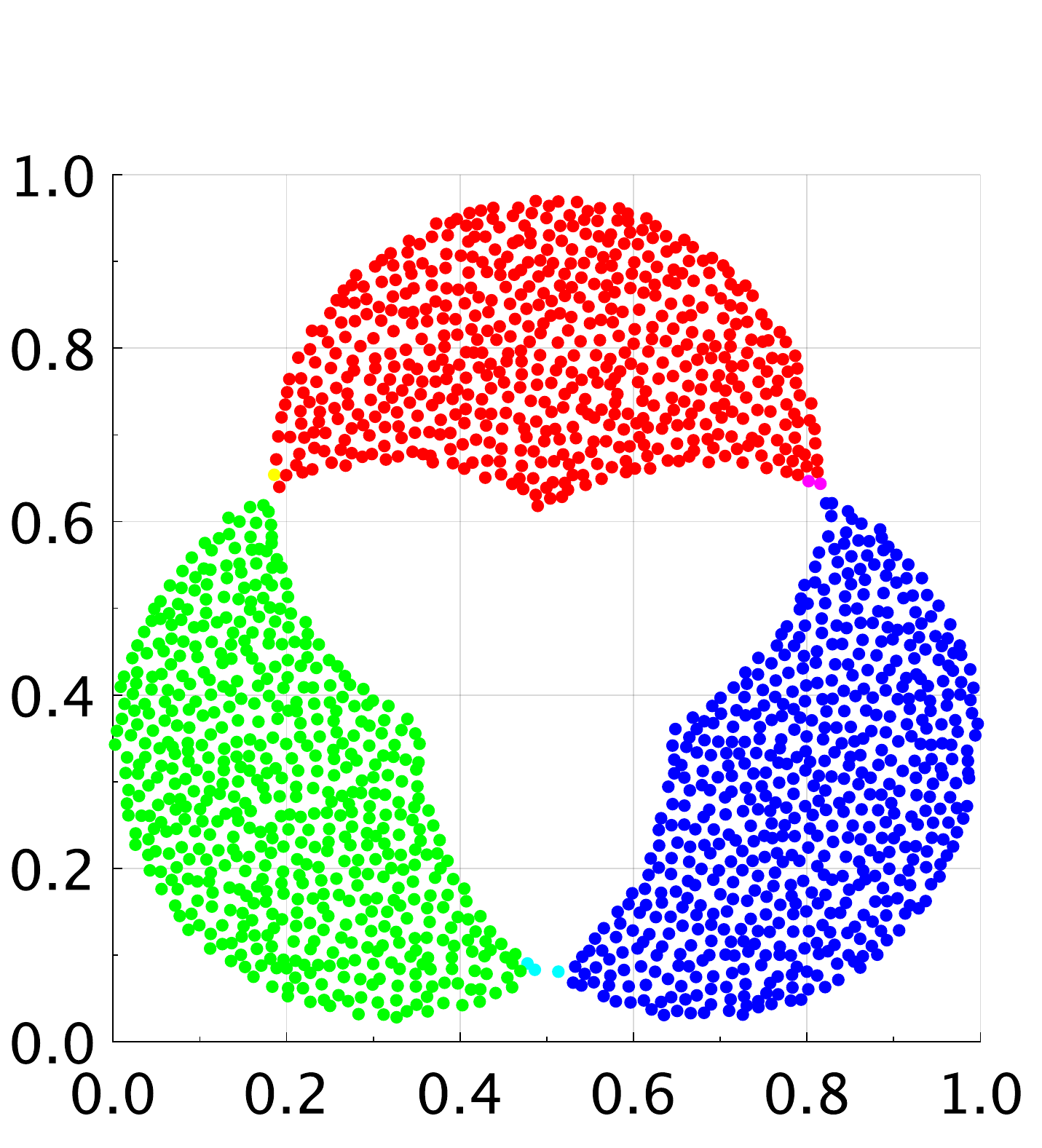}
		\label{fig:MLCA_7dist_3}
	}
	\hfil
	\subfloat[Distribution \#4]{
		\includegraphics[height=1.0in]{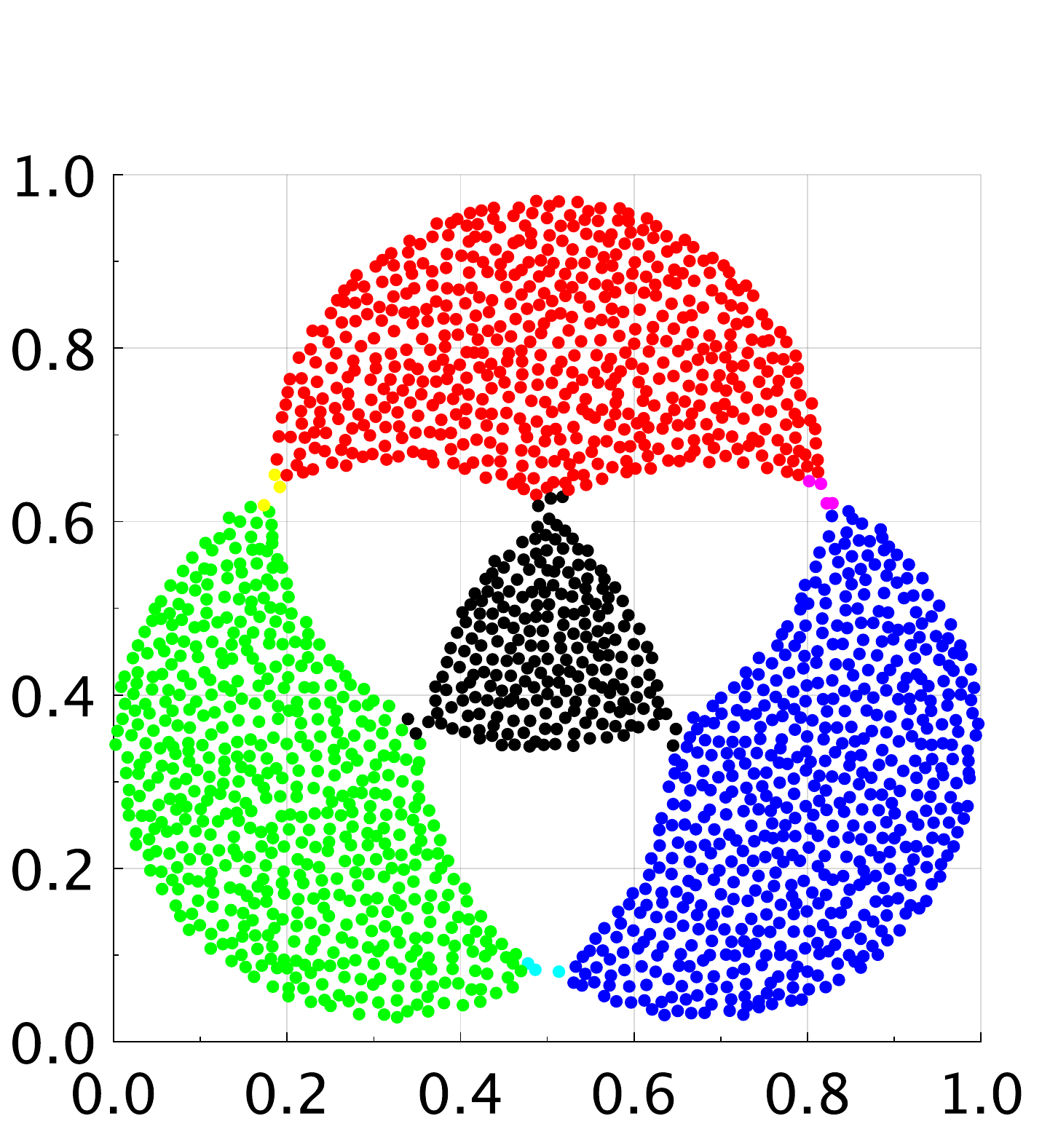}
		\label{fig:MLCA_7dist_4}
	}
	\hfil
	\subfloat[Distribution \#5]{
		\includegraphics[height=1.0in]{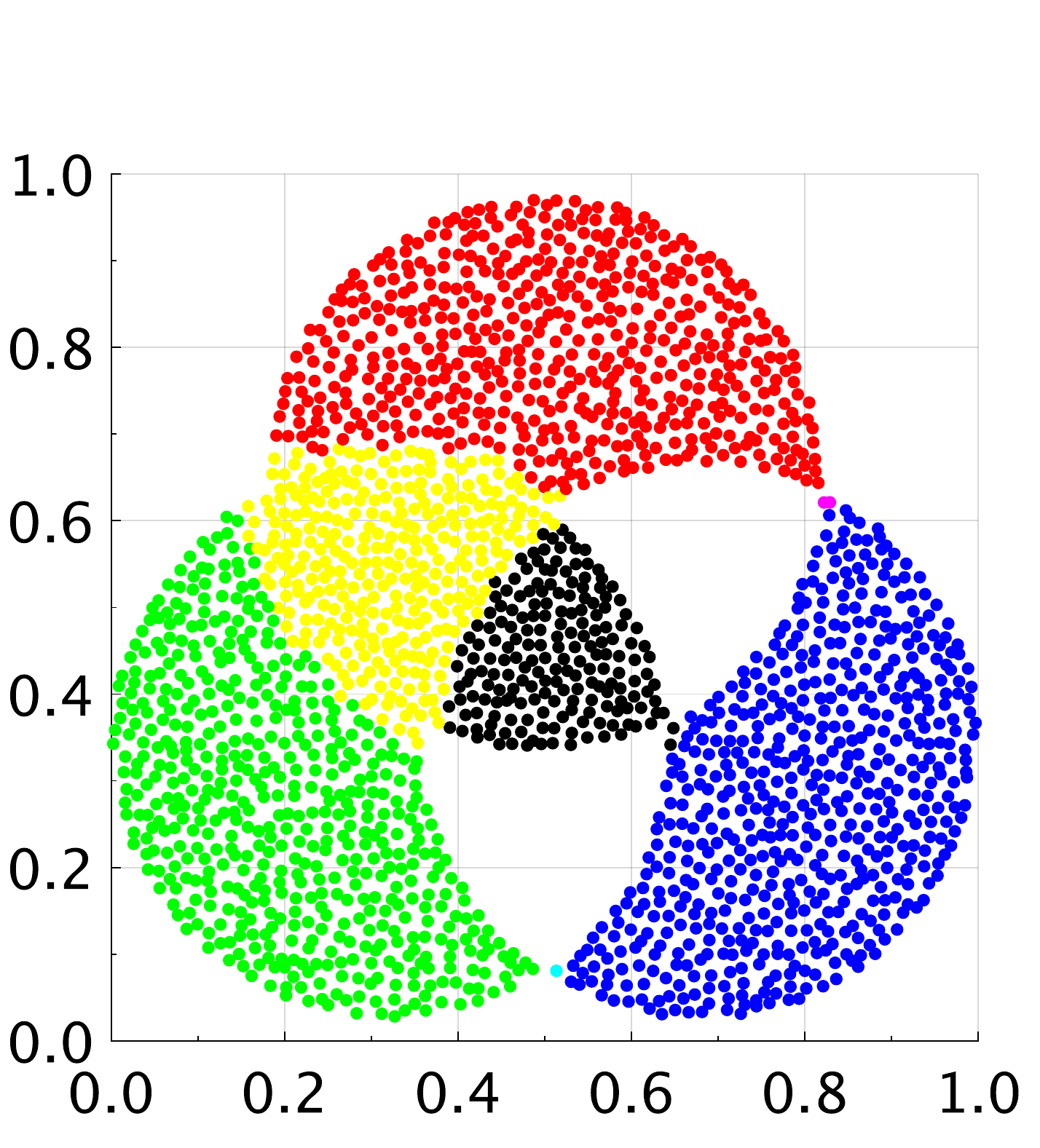}
		\label{fig:MLCA_7dist_5}
	}
	\hfil
	\subfloat[Distribution \#6]{
		\includegraphics[height=1.0in]{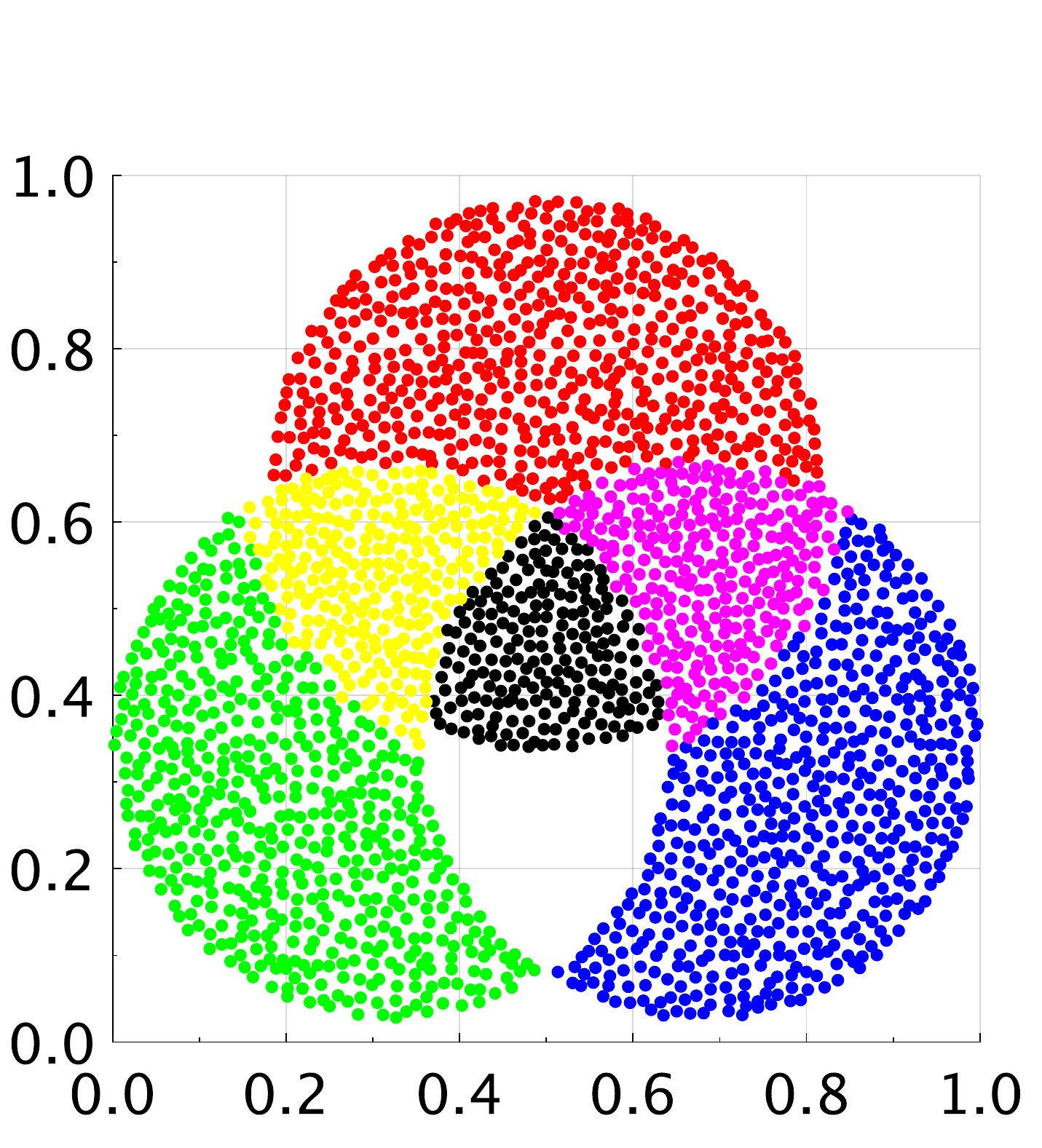}
		\label{fig:MLCA_7dist_6}
	}
	\hfil
	\subfloat[Distribution \#7]{
		\includegraphics[height=1.0in]{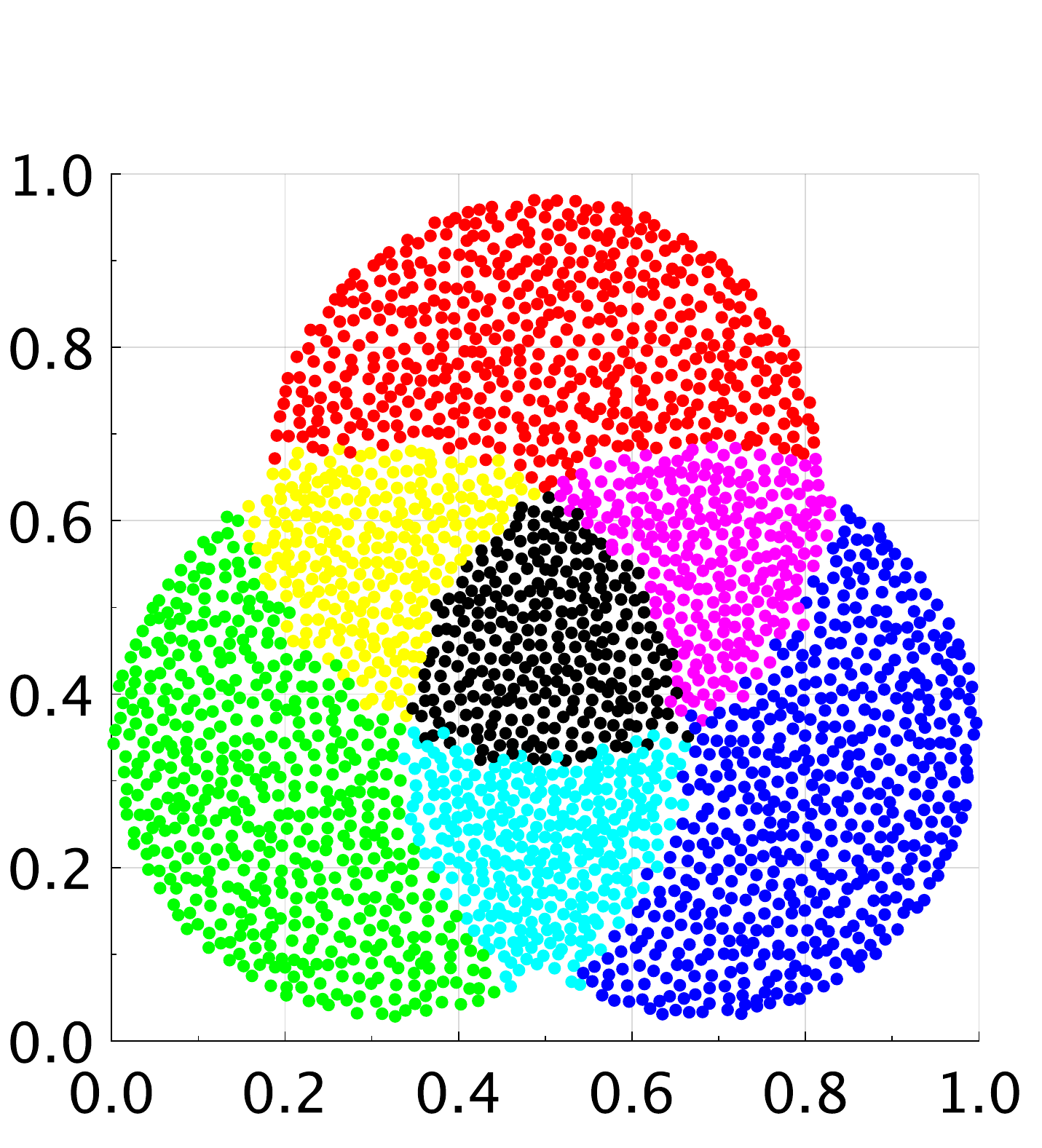}
		\label{fig:MLCA_7dist_7}
	}
	\\
	\vspace{-2mm}
	\subfloat{
		\includegraphics[height=0.13in]{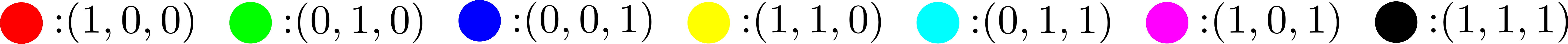}
	}
	\caption{Visualization of nodes in MLCA in the case the seven distributions are given in sequential order.}
	\label{fig:MLCA_7dist}
		\vspace{-3mm}
\end{figure*}

\begin{figure*}[htbp]
	\vspace{1mm}
	\centering
	\subfloat[Distribution \#1]{
		\includegraphics[height=1.0in]{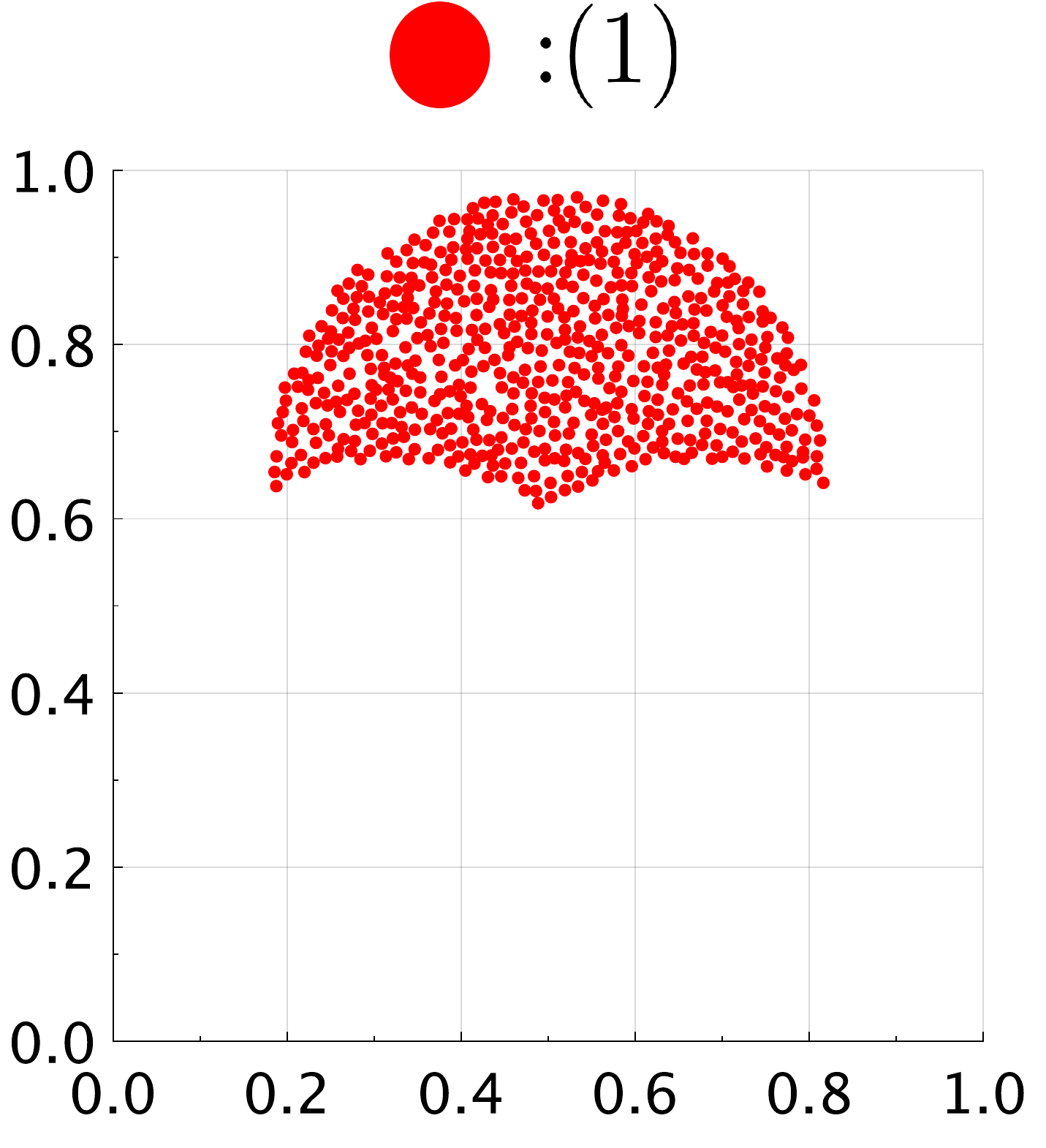}
		\label{fig:MCIC_7dist_1}
	}
	\hfil
	\subfloat[Distribution \#2]{
		\includegraphics[height=1.0in]{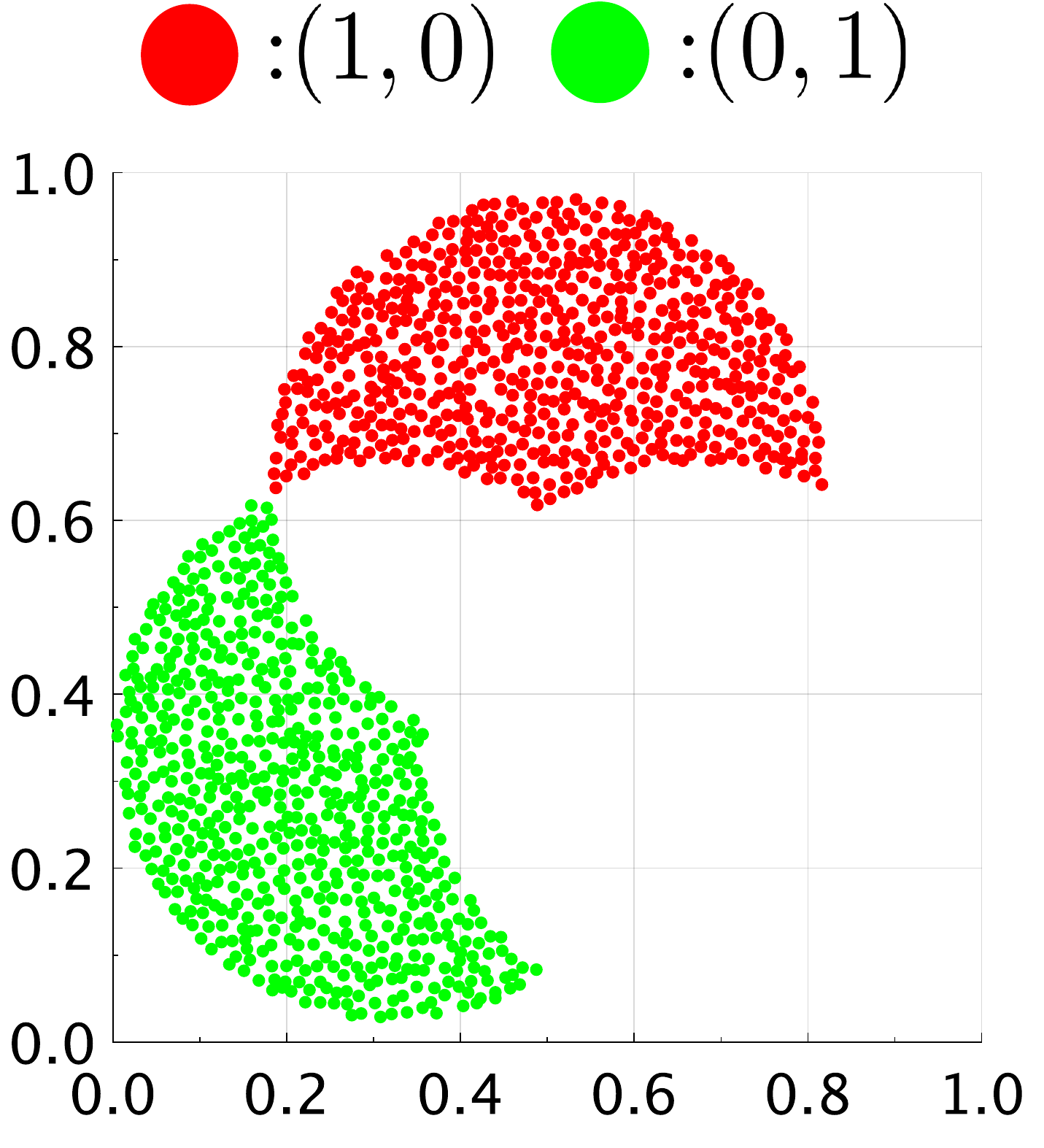}
		\label{fig:MCIC_7dist_2}
	}
	\hfil
	\subfloat[Distribution \#3]{
		\includegraphics[height=1.0in]{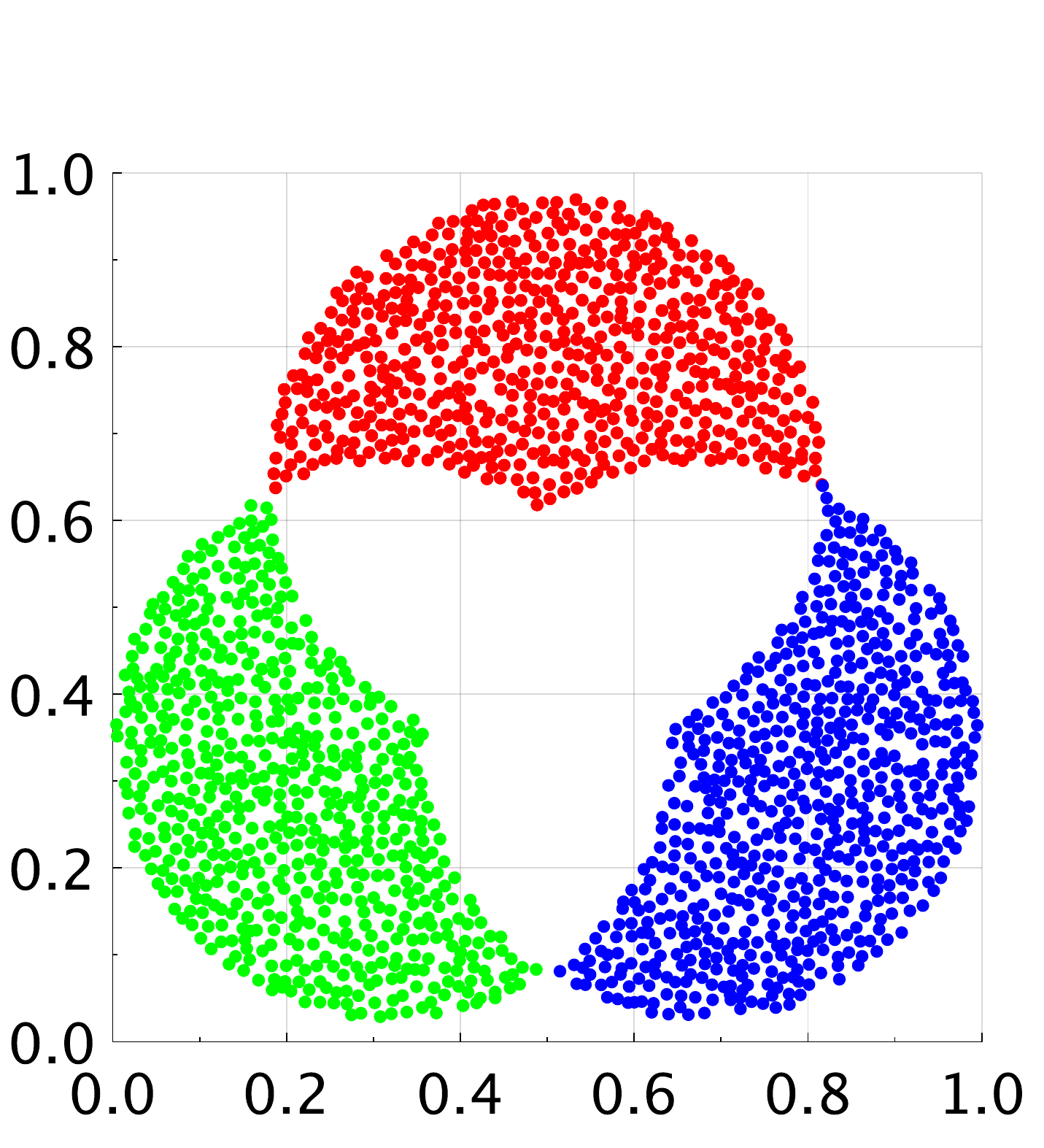}
		\label{fig:MCIC_7dist_3}
	}
	\hfil
	\subfloat[Distribution \#4]{
		\includegraphics[height=1.0in]{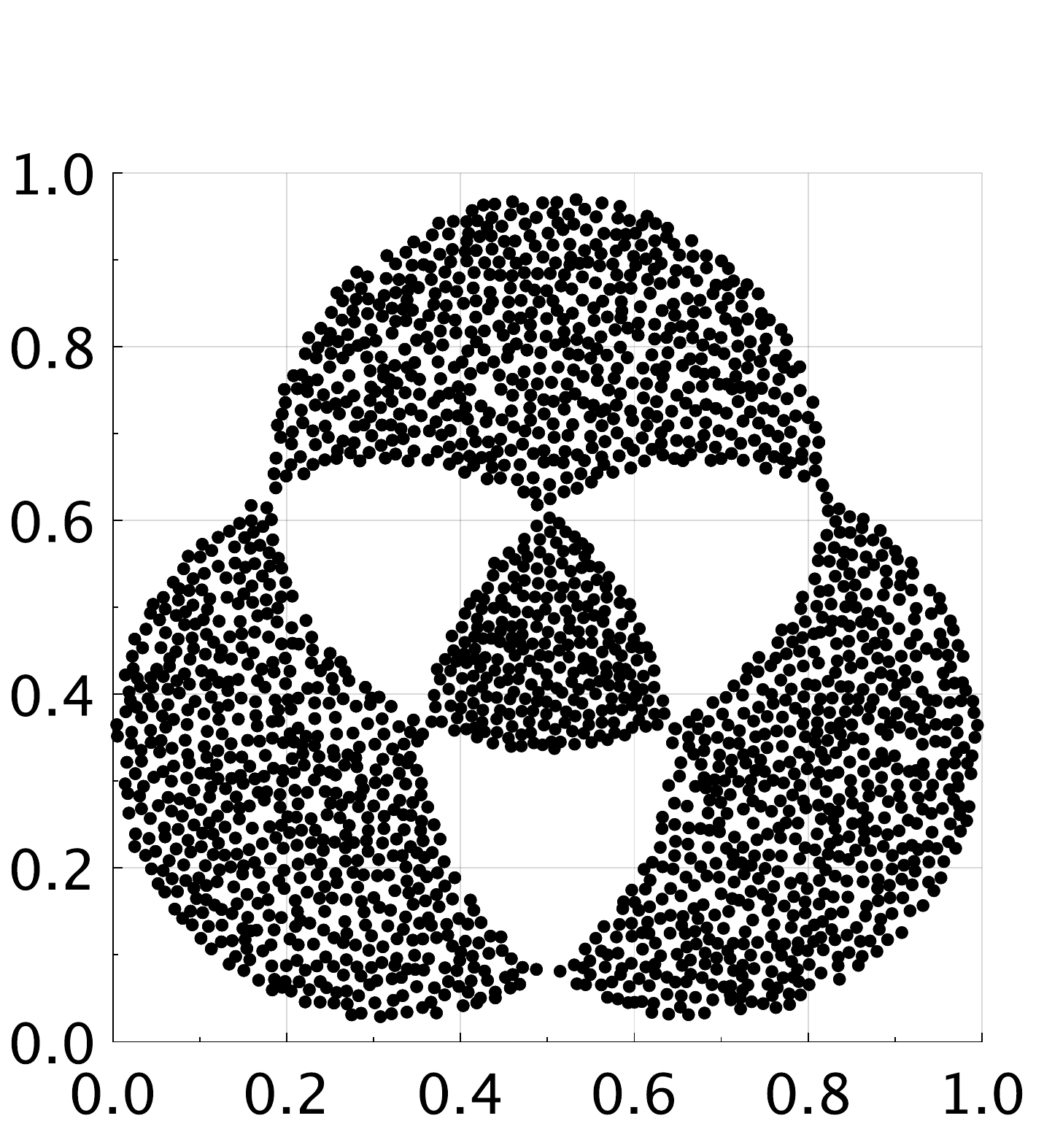}
		\label{fig:MCIC_7dist_4}
	}
	\hfil
	\subfloat[Distribution \#5]{
		\includegraphics[height=1.0in]{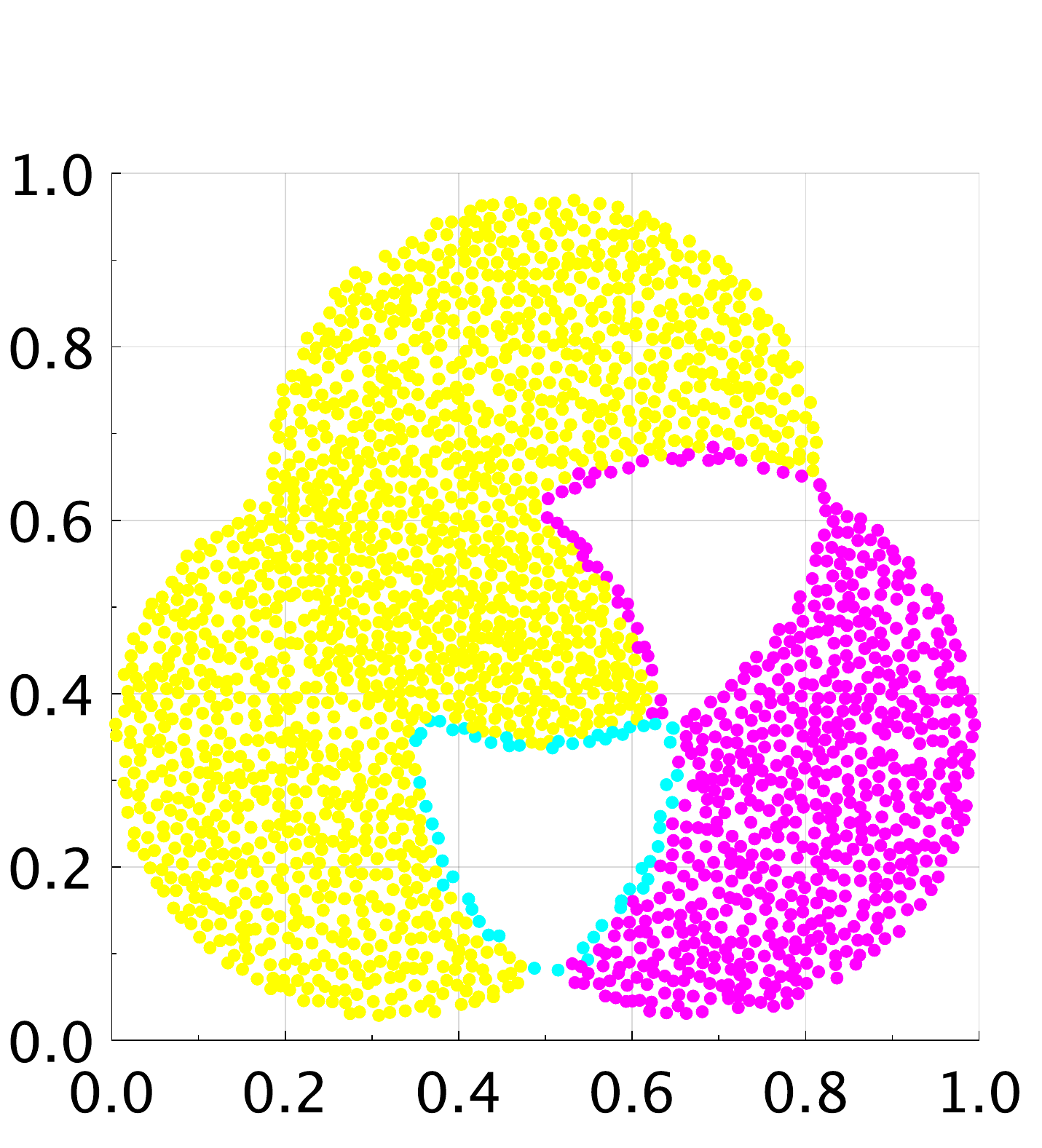}
		\label{fig:MCIC_7dist_5}
	}
	\hfil
	\subfloat[Distribution \#6]{
		\includegraphics[height=1.0in]{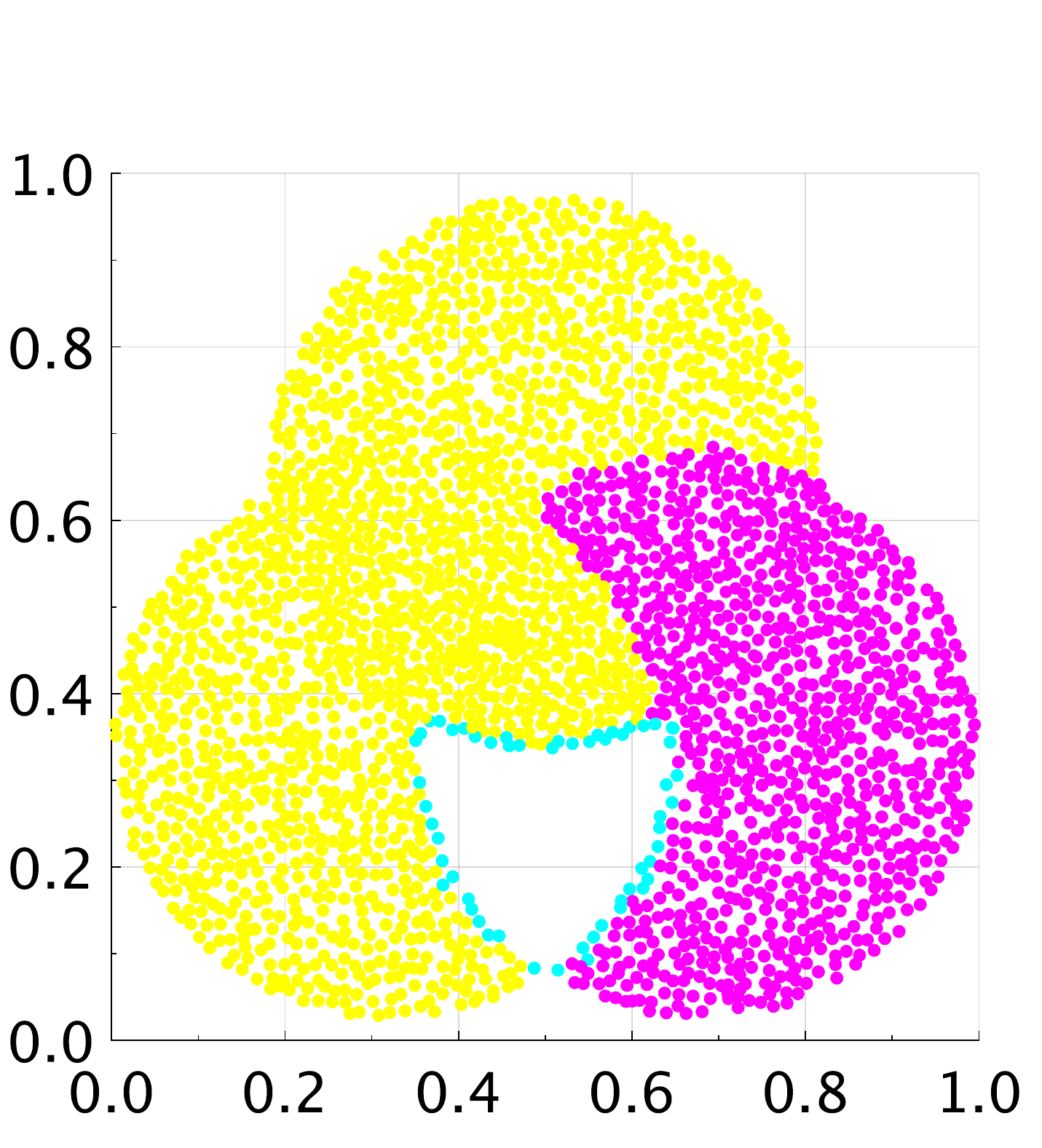}
		\label{fig:MCIC_7dist_6}
	}
	\hfil
	\subfloat[Distribution \#7]{
		\includegraphics[height=1.0in]{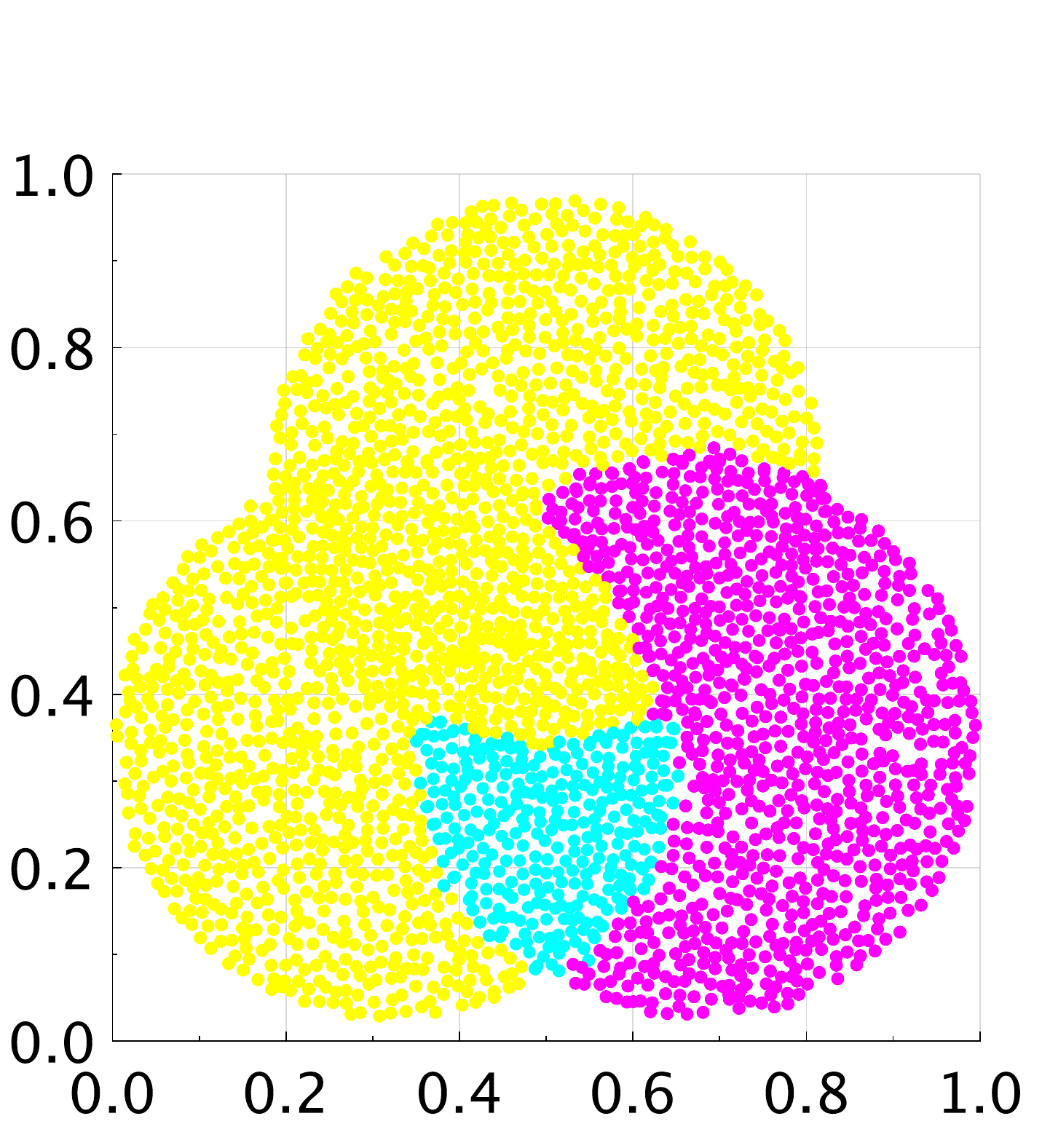}
		\label{fig:MCIC_7dist_7}
	}
	\\
	\vspace{-2mm}
	\subfloat{
		\includegraphics[height=0.13in]{figures/twoDimSynthetic_7dist_Legend_horizonL.pdf}
	}
	\caption{Visualization of nodes in MCIC in the case the seven distributions are given in sequential order.}
	\label{fig:MCIC_7dist}
\end{figure*}

To analyze the above results in detail, Figs. \ref{fig:MLCA_MCIC_1dist}-\ref{fig:MCIC_7dist} show the visualization of generated nodes and their label information in each algorithm. The results in Figs. \ref{fig:MLCA_MCIC_1dist}-\ref{fig:MCIC_7dist} are a trial which showed the highest Exact Match in each algorithm among 20 trials. The color of a node indicates a label set that the node predicts. Once a testing instance is given, the nearest node predicts a label set for the testing instance corresponding to the color shown in the legend of Figs. \ref{fig:MLCA_MCIC_1dist}-\ref{fig:MCIC_7dist}.

Comparing Figs. \ref{fig:MLCA_1dist} and \ref{fig:MCIC_1dist}, MLCA can represent the seven distributions very well while MCIC fails to represent overlapped regions. Focusing on Figs. \ref{fig:MLCA_3dist} and \ref{fig:MCIC_3dist}, MLCA and MCIC can properly preserve the information of the distribution \#1. On the other hand, after learning the distributions \#2 and \#3, MLCA properly represents overlapped regions, but MCIC fails to do so. A similar tendency can be seen in Figs. \ref{fig:MLCA_7dist} and \ref{fig:MCIC_7dist}. These results suggest that MLCA is capable of continuous learning in various environments while MCIC cannot accumulate and preserve information when distributions are adjacent or overlapping.

From the results in this section, it can be seen that MLCA adaptively generates nodes and incrementally learns label information from the given instances while maintaining the extracted knowledge.

In this section, since the label information of each distribution is not changed, the learning ability has been verified only for the knowledge that is represented by node positions/distributions. Regarding the label information (the prior probabilities and likelihood) of nodes in MLCA, it is sequentially updated based on the frequency of a label appearance. Therefore, although the location or distribution of nodes do not change, the meaning of the knowledge may change in the case where the label distribution is changed. This is known as concept drift. 

From another point of view, the label forgetting nature of MLCA may lead to superior classification performance for stream data with concept drift than other algorithms. This is an interesting future research topic.

\subsection{Quantitative Analysis}
\label{sec:quantitative}
This section presents a comparison on the classification performance of MLCA, MLCA-I, and MLCA-C with that of MCIC \cite{nguyen19}, MuENL \cite{zhu18a}, mlODM \cite{tan20}, GLOCAL \cite{zhu18}, MLSA-$ k $NN \cite{roseberry21}, and ML-$ k $NN \cite{zhang07} by utilizing real-world multi-label datasets.

The source code of MCIC \footnote{https://github.com/vu-luong/MCIC}, MuENL \footnote{https://www.lamda.nju.edu.cn/code\_MuENL.ashx}, mlODM \footnote{https://www.lamda.nju.edu.cn/code\_mlODM.ashx}, GLOCAL \footnote{https://www.lamda.nju.edu.cn/code\_Glocal.ashx}, MLSA-$ k $NN \footnote{https://github.com/canoalberto/MLSAkNN}, and ML-$ k $NN \footnote{http://www.lamda.nju.edu.cn/code\_MLkNN.ashx} are provided by authors.

\subsubsection{Datasets}
\label{sec:datasets}
We use 16 real-world multi-label datasets that six numerical and six categorical datasets from the Mulan repository \cite{tsoumakas11} and two numerical and two categorical datasets from the Extreme Classification repository \cite{bhatia16}. Table \ref{tab:datasetML} shows the statistics of the datasets. During our experiments, the training instances in each dataset are presented to each algorithm only once. In regard to pre-processing for datasets, mlODM and GLOCAL need [0, 1] scaling to maintain high classification performance, while other algorithms do not need the scaling. Therefore, we prepare the [0, 1] scaled data for mlODM and GLOCA. For other algorithms, we use the raw data with no pre-processing.

\begin{table}[htbp]
	\caption{Statistics of multi-label datasets}
	\label{tab:datasetML}
	\footnotesize
	\centering
	\renewcommand{\arraystretch}{1.2}
	\scalebox{0.93}{
	\begin{tabular}{lrrrr}
		\hline\hline
		\multirow{2}{*}{Dataset} & Number of & \multicolumn{2}{c}{Number of Attributes}  & Number of  \\
		\cline{3-4}
		&      Instances  & Numerical      & Categorical           &      Labels   \\
		\hline
		\multicolumn{5}{c}{Small-scale} \\
		\hline
		Flags                    & 194                                & 10                 & 9                 & 7   \\
		Emotions                 & 593                                & 72                 & 0                 & 6  \\
		Birds                    & 645                                & 258                & 2                 & 19  \\
		Image                    & 2,000                              & 294                & 0                 & 5   \\
		Scene                    & 2,407                              & 294                & 0                 & 6  \\
		Yeast                    & 2,417                              & 103                & 0                 & 14  \\
		\cline{1-5}
		VirusGO                 & 207                                &   0              & 749                 & 6  \\
		GpositiveGO              & 519                                &  0               & 912                 & 4   \\
		Genbase                  & 662                                &  0            & 1,186                  & 27  \\
		Medical                  & 978                                &   0           & 1,449                 & 45  \\
		PlantGo                  & 978                                &    0          &  3,091                 & 12  \\
		Langlog                  & 1,460                              &   0            & 1,004                 & 75   \\
		\hline
		\multicolumn{5}{c}{Large-scale} \\
		\hline
		EURLex-4K         & 19,348                            & 5,000          & 0                 & 3,993   \\
		Mediamill            & 43,907                            & 120          & 0                 & 101   \\
		\hline
		Bibtex                    & 7,395                              & 0                & 1,836             & 159   \\
		Delicious                & 16,105                            & 0                 & 500               & 983   \\
		\hline\hline
	\end{tabular}
	}
\end{table}

\subsubsection{Parameter Specifications}
\label{sec:parameterSettings}
MLCA, MLCA-I, MLCA-C, and all the comparison algorithms have parameters which have an impact on the classification performance. This section presents parameter specifications of each algorithm in detail.

We use grid search to specify parameter values of each algorithm. Before grid search, we separate a dataset into training instances and test instances. The training instances are 90\% of the dataset, and the testing instances are the remaining 10\%. The testing instances are used only for the performance evaluation of the designed classifier (i.e., they are not used for parameter specifications). During grid search, we train an algorithm with the training instances, and test the algorithm by using the training instances again. Thus, we obtain a parameter setting which shows the highest classification performance to the training instances. Once the best parameter setting, which is an optimal to the training instances, is specified, the classification performance of the algorithm with the best parameter setting is evaluated by using the testing instances. Since grid search for specifying the parameters does not use the testing instances, the generalization ability of each algorithm can be properly evaluated in Section \ref{sec:resultsPerformance}. 

For specifying the best parameter setting, we calculate the Exact Match in each parameter specification for the training instances in each datasets listed in Table \ref{tab:datasetML} except for large-scale datasets due to a time-consuming training process. Regarding the parameters for the large-scale datasets, we apply the same specification with Langlog dataset because it has the largest number of labels among small-scale datasets.

We repeat the evaluation 20 times (i.e., 2$ \times $10-fold Cross Validation (CV)) with training instances selected by different random seeds. Although six evaluation metrics were introduced in Section \ref{sec:metrics}, the Exact Match is used for specifying the parameters since it is the most strict evaluation metric. Table \ref{tab:paramAlgorithms} summarizes the parameters of all the algorithms. In the following, the settings and results of grid search are explained in detail by separating MLCA and its variants and comparison algorithms.

\begin{table*}[htbp]
	\vspace{1mm}
	\caption{Parameter specifications of each algorithm}
	\label{tab:paramAlgorithms}
	\footnotesize
	\centering
	\renewcommand{\arraystretch}{1.2}
	\scalebox{0.95}{
		\begin{tabular}{lllll}
			\hline\hline
			Algorithm & Paramter & Value & Grid Range & Description \\
			\hline
			MLCA & $ V $ & Grid Search & \{0.01, 0.05, 0.10, $ \ldots $, 0.95\} & Similarlity threshold \\
			& $ \lambda $ & 50 & --- & Interval for adapting $ \sigma $ in the CIM  \\
			& $ N_{y} $ & 10 & --- & The number of neighbor nodes \\
			\hline
			MLCA-I & V & Grid Search & \{0.01, 0.05, 0.10, $ \ldots $, 0.95\} & Similarlity threshold \\
			& $ \lambda $ & 50 & --- & Interval for adapting $ \sigma $ in the CIM \\
			& $ N_{y} $ & 10 & --- & The number of neighbor nodes \\
			\hline
			MLCA-C & V & Grid Search & \{0.01, 0.05, 0.10, $ \ldots $, 0.95\} & Similarlity threshold \\
			& $ \lambda $ & 50 & --- & Interval for adapting $ \sigma $ in the CIM  \\
			& $ N_{y} $ & 10 & --- & The number of neighbor nodes \\
			\hline
			MCIC & $ \theta $ & Grid Search & \{0.100, 0.200, 0.300, 0.400, 0.495\} & Boundary of a cluster \\
			& $ \lambda $ & Grid Search & \{0.00, 0.05, 0.10, $ \ldots $, 0.50\} & Decay control parameter \\
			& $ \delta $ & 0.1 & --- & Parameter for label set learning \\
			& $ K $ & 3 & --- & The number of nearest neighbors \\
			& $ \beta_{mu}$ & 3 & --- & Threshold of a mature weight \\
			\hline
			MuENL & $ \lambda_{1} $ & Grid Search & \{1.0E-6, 1.0E-5, $ \ldots $, 1.0E-0\} & Trade-off parameter of ranking loss for a multi-label classifier \\
			& $ \lambda_{2} $ & Grid Search & \{1.0E-6, 1.0E-5, $ \ldots $, 1.0E-0\} & Trade-off parameter of $ l_{2} $ regularization for a multi-label   classifier \\
			& $ I $ & 20 & --- & The maximum iteration for updating a weight \\
			\hline
			mlODM & $ \theta $ & Grid Search & \{0.10, 0.20, $ \ldots $, 0.90\} & Approximation parameter for block coordinate descent \\
			& $ \mu $ & Grid Search & \{0.10, 0.20, $ \ldots $, 0.90\} & Trade-off parameter for block coordinate descent \\
			& $ \gamma $ & 0.5 & --- & Bandwidth of a kernel function in rank-SVM \\
			& $ C $ & 1 & --- & Cost for rank-SVM \\
			& $ I $ & 4 & --- & The maximum number of iterations \\
			\hline
			GLOCAL & $ \lambda_{3} $ & Grid Search & \{0.0, 1.0E-6, 1.0E-5, $ \ldots $, 1.0E-0\} & Regularization parameter for a global and local manifold \\
			& $ \lambda_{4} $ & Grid Search & \{0.0, 1.0E-6, 1.0E-5, $ \ldots $, 1.0E-0\} & Regularization parameter for a global and local manifold \\
			& $ \lambda_{1} $ & 1 & --- & Regularization parameter for a global and local manifold \\
			& $ \lambda_{2} $ & 0.125 & --- & Regularization parameter for a global and local manifold \\
			& $ k $ & 20 & --- & Dimension of a latent space \\
			& $ g $ & 3 & --- & The number of clusters of $ k $-means \\
			\hline
			MLSA-$ k $NN & $ m_{\text{max}} $ & Grid Search & \{200, 400, $ \ldots $, 2,000\} & The maximum size of the window \\
			& $ kAdj_{\text{max}} $ & Grid Search & \{20, 40, $ \ldots $, 200\} & History length to specify $ k $ value of $ k $NN \\
			& $ m_{\text{min}} $ & 50 & --- & The minimum size of the window \\
			& $ p $ & 1 & --- & Penalty ratio \\
			& $ r  $& 0.5 & --- & Reduction ratio \\
			\hline
			ML-$ k $NN
			& $ k $ &  3 & \{3, 6, $ \ldots $, 30\}  & The number of nearest neighbors \\
			\hline\hline
		\end{tabular}
	}
	\vspace{-2mm}
\end{table*}

\vspace{1mm}\hspace{-3mm}$ \bullet $\hspace{1mm} MCLA and its Variants \vspace{0.5pt}

MLCA, MLCA-I, and MLCA-C have three parameters, i.e., the number of neighbor nodes $ N_{y} $, an interval $ \lambda $ for adapting $ \sigma $ in the CIM, and a similarity threshold $ V $. Among those parameters, the similarity threshold $ V $ has a large impact on the classification performance. Therefore, we perform grid search for $ V $ in increments of 0.05 over the range of $ V = [0.05, 0.95] $ while fixing $ N_{y} = 10 $ and $ \lambda = 50 $. The parameters $ N_{y} $ and $ \lambda $ are the same specification as in \cite{masuyama20a}. In additition, the case of $ V = 0.01 $ is also considered. This is because that MLCA with a smaller $ V $ will tend to generate more nodes which yields higher classification performance than that with a larger $ V $.

The detailed results of grid search for the similarity threshold $ V $ in MLCA and its variants are shown in Fig.1 of the supplementary file.

In this paper, the parameters of MLCA and its variants are not specified for each dataset, but the same settings are applied to all datasets. Note that the parameters of the comparison algorithms are specified for each dataset in order to achieve the maximum classification performance. To specify the appropriate parameters for all datasets, we adopt the Friedman test and Nemenyi post-hoc analysis \cite{demvsar06} to conduct statistical comparisons among the different parameter specifications ($ V = [0.01, 0.45] $ in Fig. 1 of the supplementary file) by using results on the Exact Match of all datasets. The Friedman test is used to test the null hypothesis that all algorithms perform equally. If the null hypothesis is rejected, the Nemenyi post-hoc analysis is then conducted. The Nemenyi post-hoc analysis is utilized for all pairwise comparisons based on the ranks of results on the Exact Match of all datasets. The difference in the performance of two algorithms is treated as statistically significant if the $ p $-value defined by the Nemenyi post-hoc analysis is smaller than the significance level. Here, the null hypothesis is rejected at the significance level of $ 0.05 $ both in the Friedman test and the Nemenyi post-hoc analysis.

\begin{figure*}[htbp]
	\centering
	\subfloat[MLCA]{
		\includegraphics[width=2.2in]{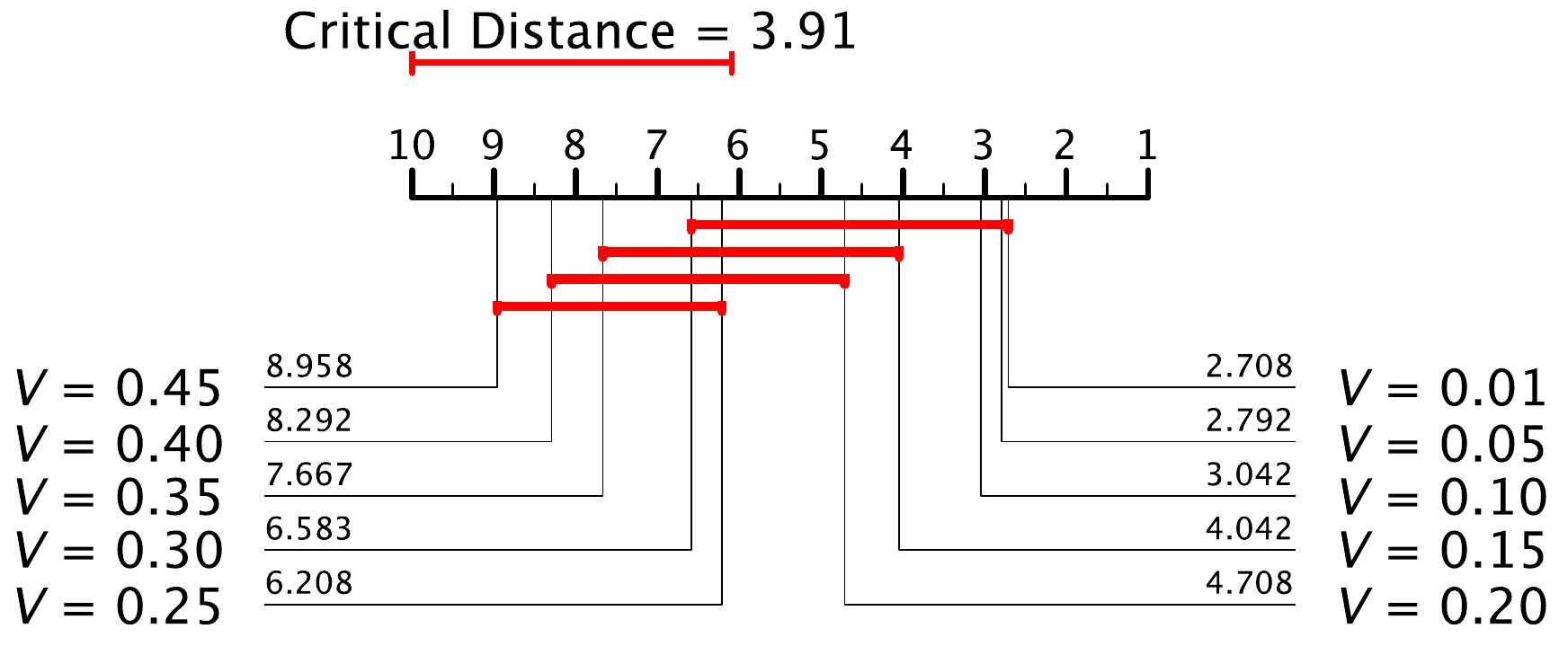}
		\label{fig:paramCD_O}
	}
	\hfil
	\subfloat[MLCA-I]{
		\includegraphics[width=2.2in]{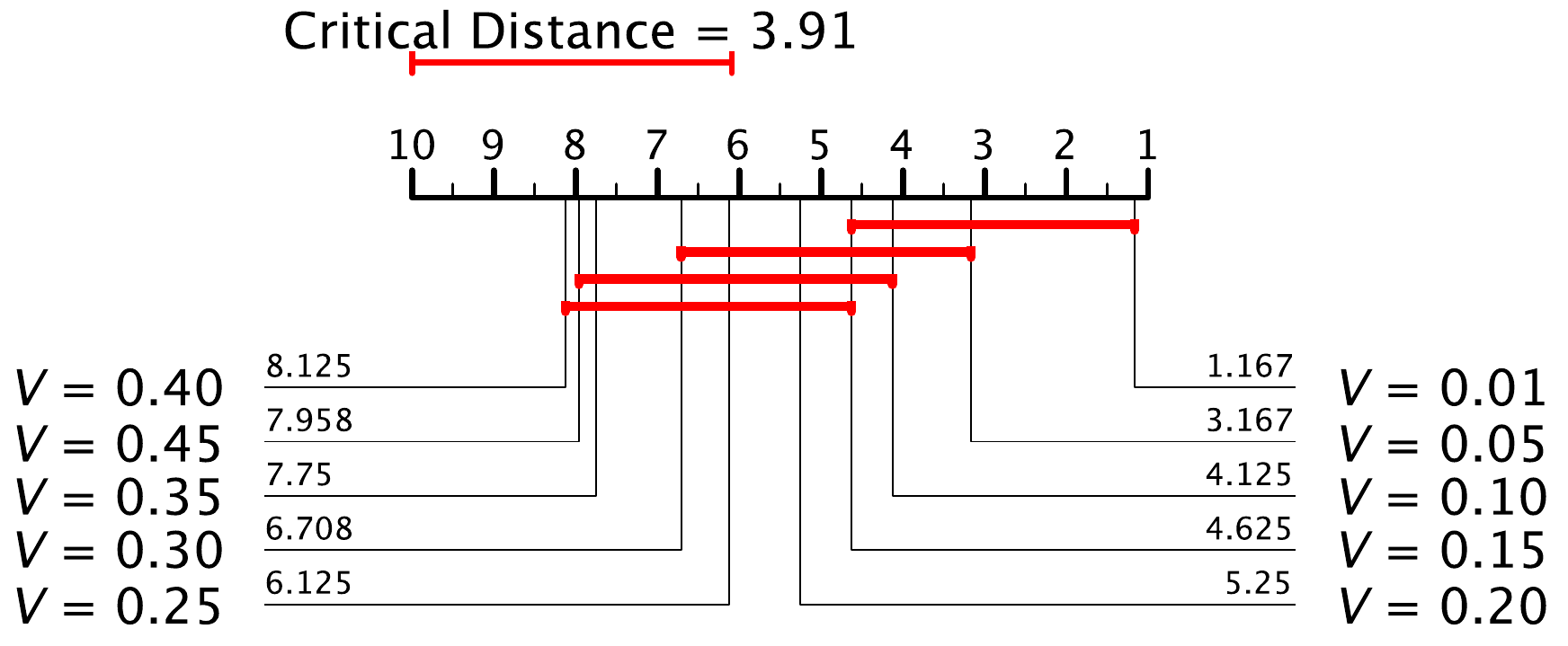}
		\label{fig:paramCD_A}
	}
	\hfil
	\subfloat[MLCA-C]{
		\includegraphics[width=2.2in]{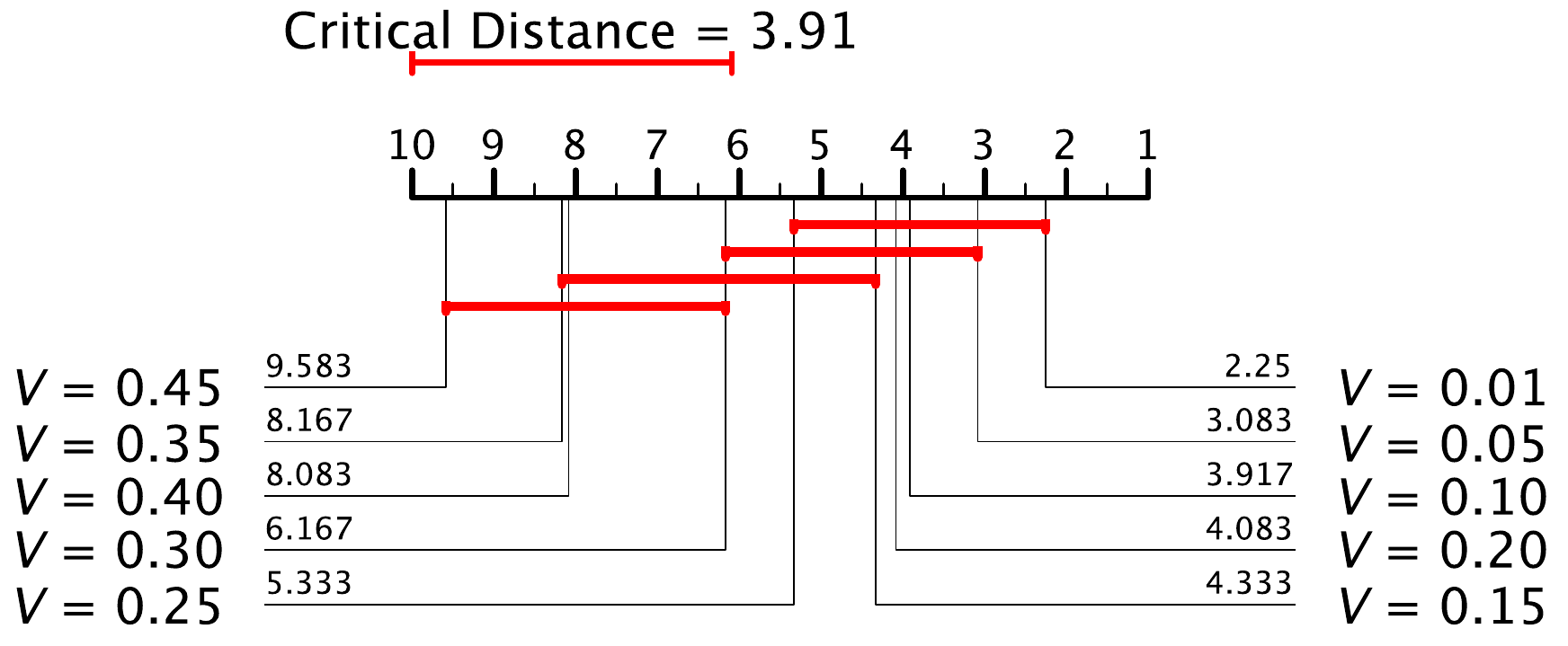}
		\label{fig:paramCD_C}
	}
	\caption{Critical difference diagram for the Exact Match by different parameter specifications.}
	\label{fig:paramCD}
\end{figure*}

\begin{table*}[htbp]
	\vspace{2mm}
	\caption{Parameter specifications for each dataset by grid search}
	\label{tab:paramGrid}
	\footnotesize
	\centering
	\renewcommand{\arraystretch}{1.2}
	\scalebox{0.93}{
		\begin{tabular}{lccccccccccccccc}
			\hline\hline
			\multicolumn{1}{c}{\multirow{3}{*}{Dataset}} & \multicolumn{2}{c}{\multirow{2}{*}{MCIC}} & \multicolumn{2}{c}{\multirow{2}{*}{MuENL}} & \multicolumn{2}{c}{mlODM} & \multicolumn{2}{c}{mlODM} & \multicolumn{2}{c}{GLOCAL} & \multicolumn{2}{c}{GLOCAL} & \multicolumn{2}{c}{\multirow{2}{*}{MLSA-$k$NN}} & \multicolumn{1}{c}{\multirow{2}{*}{ML-$k$NN}} \\
			\multicolumn{1}{c}{} & \multicolumn{2}{c}{} & \multicolumn{2}{c}{} & \multicolumn{2}{c}{(Scaling)} & \multicolumn{2}{c}{(Raw)} & \multicolumn{2}{c}{(Scaling)} & \multicolumn{2}{c}{(Raw)} & \multicolumn{2}{c}{} & \multicolumn{1}{c}{} \\
			\cline{2-16}
			\multicolumn{1}{c}{} & $ \theta $ & $ \lambda $ & $ \lambda_{3} $ & $ \lambda_{4} $ & $ \theta $ & $ \mu $ & $ \theta $ & $ \mu $ & $ \lambda_{3} $ & $ \lambda_{4} $ & $ \lambda_{3} $ & $ \lambda_{4} $ & $ m_{\text{max}} $ & $ kAdj_{\text{max}} $ & $ k $ \\
			\hline
			Flags & 0.100 & 0.00 & 1.0 & 1.0 & 0.90 & 0.30 & 0.10 & 0.30 & 0.0 & 1.0 & 0.0 & 0.0 & 200 & 20 & 3 \\
			Emotions & 0.100 & 0.00 & 1.0E-6 & 1.0E-6 & 0.70 & 0.10 & 0.70 & 0.30 & 1.0E-4 & 1.0 & 1.0E-6 & 1.0E-6 & 200 & 20 & 3 \\
			Birds & 0.100 & 0.00 & 1.0 & 1.0 & 0.70 & 0.40 & 0.10 & 0.10 & 1.0E-1 & 1.0 & 0.0 & 0.0 & 200 & 20 & 3 \\
			Image & 0.100 & 0.00 & 1.0E-6 & 1.0E-6 & 0.70 & 0.40 & 0.70 & 0.40 & 1.0E-5 & 1.0E-4 & 1.0E-5 & 1.0E-4 & 200 & 20 & 6 \\
			Scene & 0.100 & 0.00 & 1.0 & 1.0& 0.90 & 0.10 & 0.90 & 0.10 & 0.0 & 0.0 & 1.0E-2 & 1.0 & 200 & 20 & 3 \\
			Yeast & 0.100 & 0.00 & 1.0 & 1.0& 0.90 & 0.30 & 0.50 & 0.10 & 1.0 & 1.0E-1 & 0.0 & 0.0 & 600 & 20 & 3 \\
			VirusGO & 0.100 & 0.00 & 1.0E-1 & 1.0E-6 & 0.70 & 0.10 & 0.70 & 0.10 & 0.0 & 0.0 & 0.0 & 0.0 & 200 & 20 & 3 \\
			GpositiveGO & 0.495 & 0.45 & 1.0 & 1.0 & 0.90 & 0.10 & 0.90 & 0.10 & 0.0 & 0.0 & 0.0 & 0.0 & 200 & 20 & 3 \\
			Genbase & 0.200 & 0.00 & 1.0 & 1.0E-6 & 0.10 & 0.10 & 0.10 & 0.10 & 0.0 & 0.0 & 0.0 & 1.0E-6 & 600 & 20 & 3 \\
			Medical & 0.100 & 0.20 & 1.0 & 1.0 & 0.10 & 0.10 & 0.10 & 0.10 & 0.0 & 1.0E-4 & 1.0E-6 & 1.0E-4 & 200 & 20 & 3 \\
			PlantGO & 0.100 & 0.00 & 1.0 & 1.0E-6 & 0.20 & 0.10 & 0.20 & 0.10 & 1.0E-3 & 1.0E-4 & 1.0E-3 & 1.0E-4& 400 & 20 & 6 \\
			Langlog & 0.100 & 0.10 & 1.0 & 1.0 &  \multicolumn{2}{c}{N/A} &  \multicolumn{2}{c}{N/A} &  1.0E-1 & 1.0E-5  &  1.0E-2 & 1.0E-3   & 800 & 20 & 3 \\
			\hline
			EURLex-4K & 0.100 & 0.10 & 1.0 & 1.0 & \multicolumn{2}{c}{N/A} &  \multicolumn{2}{c}{N/A} &  1.0E-1 & 1.0E-5  &  1.0E-2 & 1.0E-3  & 800 & 20 & 3 \\
			Mediamill & 0.100 & 0.10 & 1.0 & 1.0 &  \multicolumn{2}{c}{N/A} &  \multicolumn{2}{c}{N/A} &  1.0E-1 & 1.0E-5  &  1.0E-2 & 1.0E-3  & 800 & 20 & 3 \\
			Bitbtex & 0.100 & 0.10 & 1.0 & 1.0 & \multicolumn{2}{c}{N/A} &  \multicolumn{2}{c}{N/A} &  1.0E-1 & 1.0E-5  &  1.0E-2 & 1.0E-3  & 800 & 20 & 3 \\
			Delicious & 0.100 & 0.10 & 1.0 & 1.0 & \multicolumn{2}{c}{N/A} &  \multicolumn{2}{c}{N/A} &  1.0E-1 & 1.0E-5  &  1.0E-2 & 1.0E-3  & 800 & 20 & 3 \\
			\hline\hline
		\end{tabular}
	}
	\\
	\vspace{1mm}
	\footnotesize \raggedright	\hspace{4mm}N/A indicates that an algorithm could not build a predictive model within 5 days under the available computational resources.
	\vspace{-2mm}
\end{table*}

Fig. \ref{fig:paramCD} shows critical difference diagrams for different parameter specifications of each algorithm. A better specification has lower average ranks, i.e., on the right side of a critical distance diagram. In theory, the parameter specifications within a critical distance (i.e., a red line) do not have a statistically significance difference \cite{demvsar06}. From the results in Fig. \ref{fig:paramCD}, parameter values to be used in comparisons of classification performance by testing instances are specified as follows: $ V = 0.01 $ and $ 0.30 $ for MLCA, $ V = 0.01 $ and $ 0.15 $ for MLCA-I, and $ V = 0.01 $ and $ 0.25 $ for MLCA-C. As mentioned earlier, MLCA and its variants utilize the above $ V $ values for all the datasets in the classification experiments.

\vspace{1mm}\hspace{-3mm}$ \bullet $\hspace{1mm} Comparison Algorithms \vspace{1mm}

When grid search was used in the original paper where each algorithm was proposed, we use the same grid search with the same range of parameter values for each algorithm. Otherwise, we choose one or two parameters which have the largest effects on the classification performance. The details are as follows:

{\bf MCIC}: The range of the decay controlled parameter $ \lambda $ is the same range in the original paper. Although the boundary of a cluster $ \theta $ is fixed in the original paper, it has a large effect to the clustering performance. The range of $ \theta $ is theoretically defined as $ 0 < \theta < 0.5 $.

{\bf MuENL}: In the original paper, grid search is performed to $ \lambda_{1}, \lambda_{2} \in \{ 0.001, 0.01, 0.1, 1 \} $. In this paper, we set a wider range that includes the above one.

{\bf mlODM}: The parameters and their grid ranges are the same as in the original paper.

{\bf GLOCAL}: The parameters and their grid ranges are the same as in the original paper.

{\bf MLSA-$ k $NN}: In the original paper, the authors mentioned that a window size $ kAdj_{\text{max}} $ and the maximum window size $ m_{\text{max}} $ need to be set appropriately values in advance depending on datasets. In the original paper, these parameters are fixed as $ kAdj_{\text{max}} = 100 $ and $ m_{\text{max}} = 1,000 $.
In this paper, grid search was performed in the ranges around these values.

{\bf ML-$ k $NN}: In the original paper, grid search was performed in the range of $ k = \{8, 9, 10, 11, 12 \} $. However, there was no significant difference in the classification performance. In this paper, we set wider range for finding an appropriate value.

Table \ref{tab:paramAlgorithms} summarizes the parameters for grid search and their ranges. Table \ref{tab:paramGrid} shows the parameters that indicate the highest Exact Match for the training instances in each dataset as determined by grid search. N/A indicates that an algorithm can not build a predictive model within 5 days under the available computational resources. In this paper, we assign the worst evaluation value to each metric if an algorithm can not build a predictive model.

Since MuENL has 10 parameters, we only consider three parameters because the rest of the parameters are related to class-incremental learning that is not performed in this section. Regarding mlODM and GLOCAL, there is a large difference in the classification performance depending on whether a dataset is pre-processed or not. In this paper, therefore, we use the [0, 1] scaled data and raw data without pre-processing in the learning process of mlODM and GLOCAL, and denote them as mlODM (Scaling), mlODM (Raw), GLOCAL (Scaling), and GLOCAL (Raw), respectively. Note that MCIC and MuENL are capable of continual learning while mlODM, GLOCAL, MLSA-$ k $NN and ML-$ k $NN cannot deal with it.

\subsubsection{Experimental Conditions}
\label{sec:conditionsQualitative}

We evaluate the classification performance of each algorithm by using datasets in Table \ref{tab:datasetML} and parameter specifications in Table \ref{tab:paramAlgorithms}. For small-scale datasets, we repeat the evaluation 20 times (i.e., 2$ \times $10 CV) with a different random seed for obtaining consistent averaging results. In this section, each algorithm is trained by using the same training instances with the one in Section \ref{sec:parameterSettings}, and 10\% of instances, which are not used in Section \ref{sec:parameterSettings}, are used as testing instances. For the large-scale datasets, indexes for training and testing data are provided in \cite{bhatia16}. Thus, we use those indexes for training and testing in each algorithm. Similar to Section \ref{sec:parameterSettings}, the Friedman test and Nemenyi post-hoc analysis \cite{demvsar06} are utilized. The Friedman test is used to test the null hypothesis that all algorithms perform equally. If the null hypothesis is rejected, the Nemenyi post-hoc analysis is then conducted. The Nemenyi post-hoc analysis is used for all pairwise comparisons based on the ranks of results over all the evaluation metrics of all datasets. Here, the null hypothesis is rejected at the significance level of $ 0.05 $ both in the Friedman test and the Nemenyi post-hoc analysis. All computations are carried out on Matlab 2020a with 2.2GHz Xeon Gold 6238R processor and 768GB RAM.

\subsubsection{Experimental Results}
\label{sec:resultsPerformance}
We compare the classification performance of each algorithm by using a critical difference diagram defined by the Nemenyi post-hoc analysis. The detailed results of six evaluation metrics for each dataset are summarized in Tables 1-4 on the supplementary file.

\begin{figure}[htbp]
	\begin{minipage}[b]{1.0\linewidth}
		\vspace{3mm}
		\centering
		\includegraphics[width=3.5in]{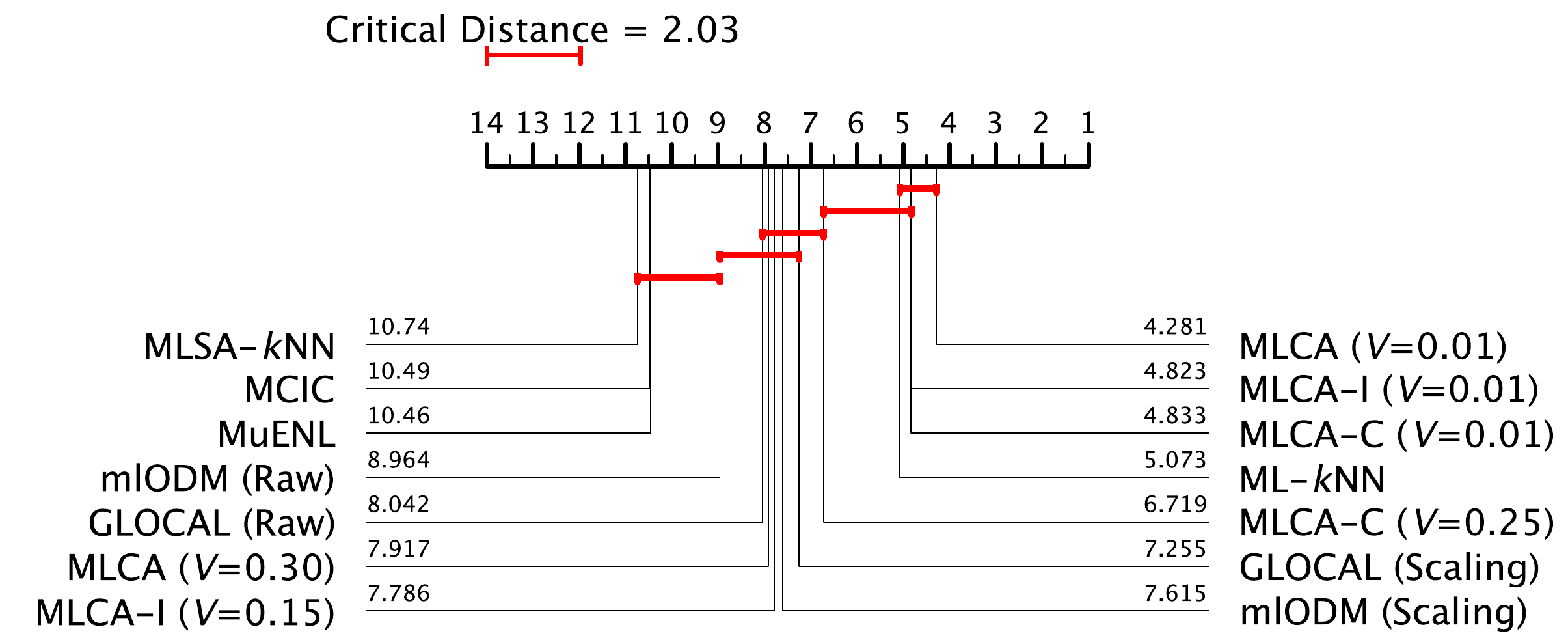}
		\caption{Critical difference diagram based on the classification performance for all the datasets.}
		\label{fig:CD_all}
		\vspace{6mm}
	\end{minipage} \\
	\begin{minipage}[b]{1.0\linewidth}
		\centering
		\includegraphics[width=3.5in]{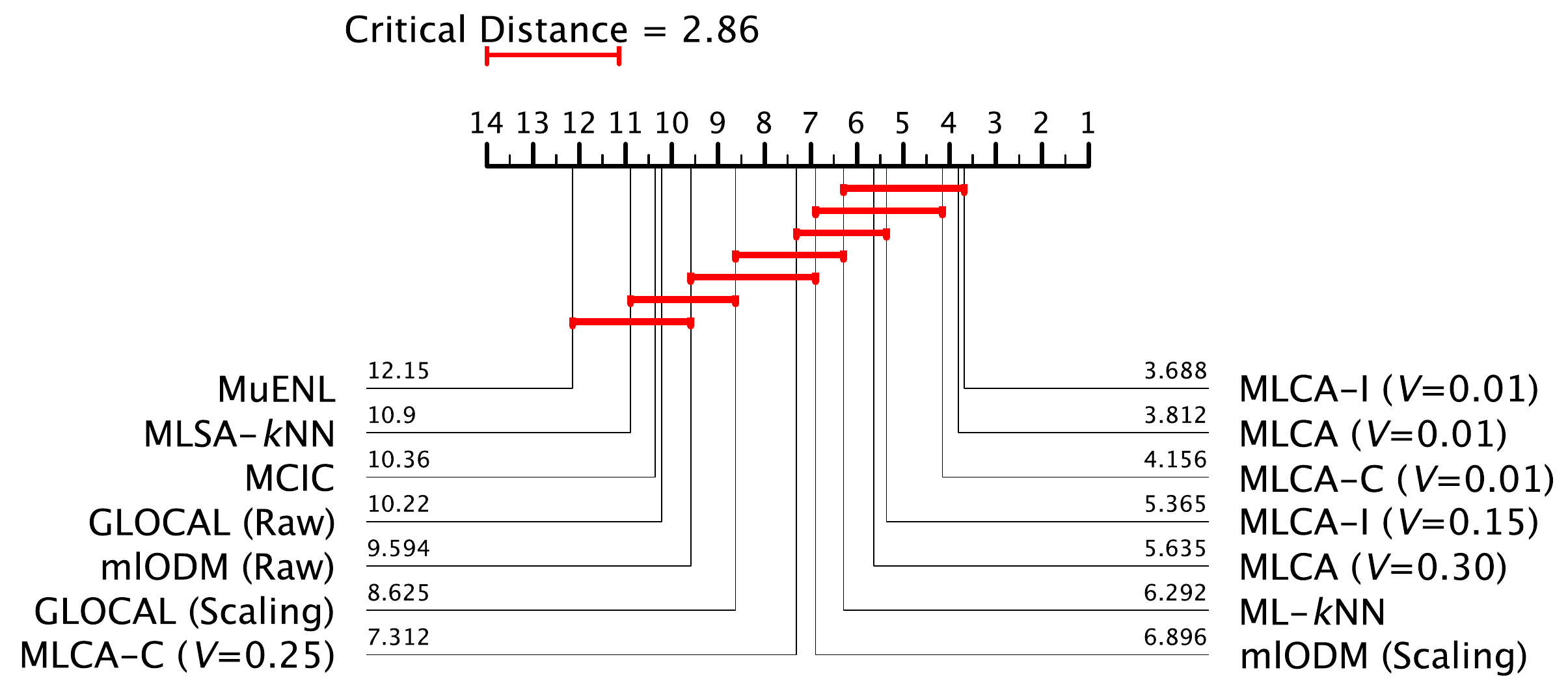}
		\caption{Critical difference diagram based on the classification performance for numerical datasets.}
		\label{fig:CD_numerical}
		\vspace{6mm}
	\end{minipage} \\
	\begin{minipage}[b]{1.0\linewidth}
		\centering
		\includegraphics[width=3.5in]{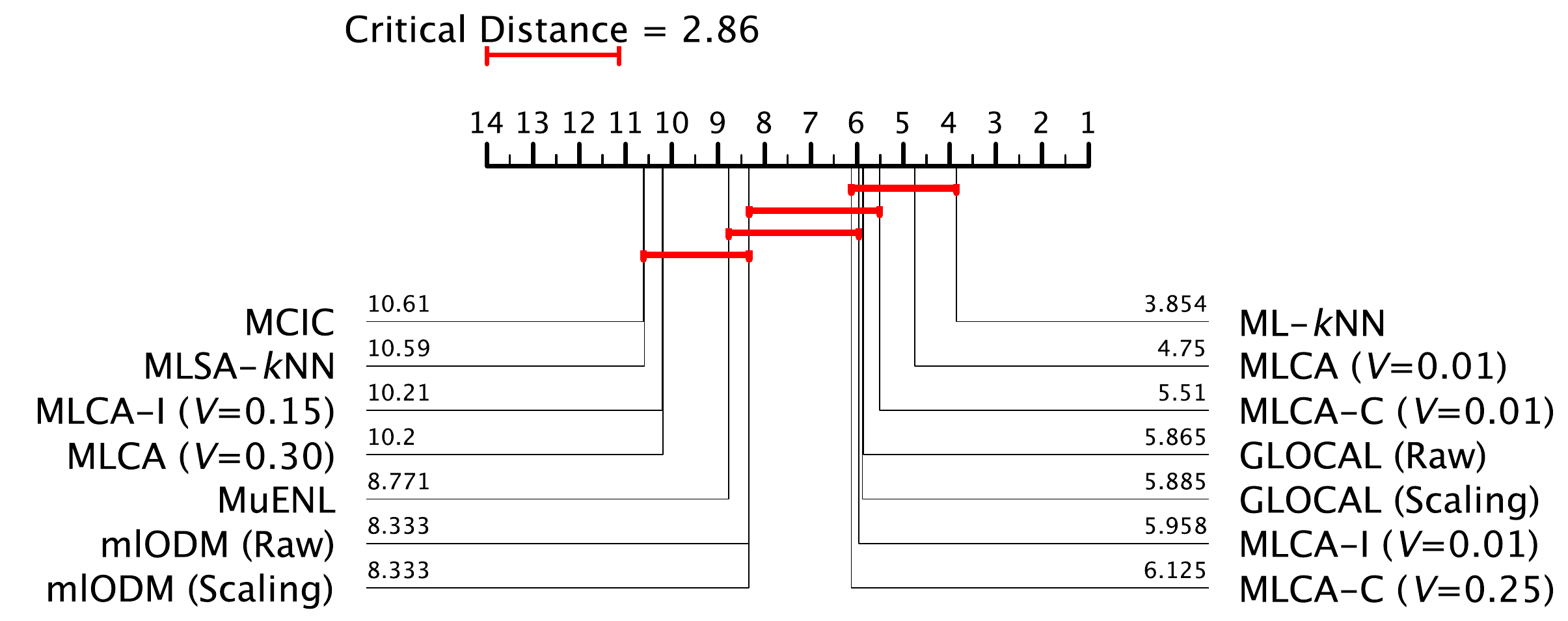}
		\caption{Critical difference diagram based on the classification performance for numerical datasets.}
		\label{fig:CD_nominal}
	\end{minipage}
\end{figure}

Fig. \ref{fig:CD_all} shows a critical difference diagram based on the classification performance for all the datasets. A better specification has lower average ranks, i.e., on the right side of a critical distance diagram. In theory, the parameter specifications within a critical distance (i.e., a red line) do not have a statistically significance difference \cite{demvsar06}.

The results are derived from all the evaluation metrics for all the datasets. Therefore, we regard that MLCA ($ V $=\hspace{1mm}0.01) shows excellent classification performance on various datasets. MLCA-I ($ V $=\hspace{1mm}0.01) and MLCA-C ($ V $=\hspace{1mm}0.01) are also superior algorithms than other comparison algorithms. Focusing on MLCA and its variants with a larger $ V $ specification, MLCA-C ($ V $=\hspace{1mm}0.25) shows a lower rank than MLCA-I ($ V $=\hspace{1mm}0.15) and MLCA ($ V $=\hspace{1mm}0.30). It means that MLCA-C is capable of maintaining high classification performance while compressing information.

Regarding comparison algorithms, ML-$ k $NN shows a superior classification performance than other algorithms except for MLCA and its variants. In Fig. \ref{fig:CD_all}, mlODM (Scaling) and GLOCAL (Scaling) are better than mlODM (Raw) and GLOCAL (Raw), respectively. Since MCIC, MuENL, and MLSA-$ k $NN are capable of handling streaming data, their classification performance is inferior to algorithms that perform batch learning such as ML-$ k $NN, mlODM, and GLOCAL.

In order to discuss the features of each algorithm in detail, Figs. \ref{fig:CD_numerical} and \ref{fig:CD_nominal} show critical difference diagrams defined by numerical datasets and categorical datasets, respectively. In the case of numerical datasets (Fig. \ref{fig:CD_numerical}), the classification performance of MLCA and its variants are still superior to comparison algorithms. MLCA-I ($ V $=\hspace{1mm}0.01) shows the lowest rank. In addition, MLCA-I ($ V $=\hspace{1mm}0.15) also shows a good performance. Thus, we regard that MLCA-I is suitable for numerical data sets. On categorical datasets (Fig. \ref{fig:CD_nominal}), ML-$ k $NN and GLOCAL perform very well in contrast to the case of numerical datasets. However, these algorithms have an obvious drawback compared to MLCA and its variants because they require all the training instances in advance. In terms of MLCA and its variants, MLCA ($ V $=\hspace{1mm}0.01) shows the lowest rank. The performance of MLCA and MLCA-I is greatly affected by a specification of a similarity threshold $ V $. In contrast, the difference in performance between MLCA-C ($ V $=\hspace{1mm}0.01) and MLCA-C ($ V $=\hspace{1mm}0.25) is small. Therefore, we regard that MLCA-C has stable classification performance on nominal datasets.

Table \ref{tab:nodesMLCA} shows the average number of generated nodes after learning the training instances. Focusing on MLCA and MLCA-I, these algorithms generate only a very small number of nodes in the case of nominal datasets, especially when a similarity threshold $ V $ is large. On the other hand, MLCA-C can maintain the sufficient number of nodes for classification even in the case of nominal datasets. Thus, MLCA-C is considered to be a strong algorithm for nominal datasets, and this property can be seen in Figs. \ref{fig:grid_virusgo}-\ref{fig:grid_langlog}.

From the results in Figs. \ref{fig:CD_all}-\ref{fig:CD_nominal}, and Table \ref{tab:nodesMLCA}, the characteristics of MLCA and its variants can be analyzed as follows:

\begin{table*}[htbp]
	\vspace{2mm}
	\centering
	\caption{Average number of nodes generated by MLCA, MLCA-I, and MLCA-C}
	\renewcommand{\arraystretch}{1.2}
	\label{tab:nodesMLCA}
	\begin{tabular}{l|r|r|r|r|r|r}
		\hline\hline
		\multicolumn{1}{l|}{\multirow{2}{*}{Dataset}} & \multicolumn{2}{c|}{MLCA} & \multicolumn{2}{c|}{MLCA-I} & \multicolumn{2}{c}{MLCA-C} \\
		\multicolumn{1}{c|}{} & \multicolumn{1}{c|}{$ V = 0.01 $} & \multicolumn{1}{c|}{$ V = 0.30 $} & \multicolumn{1}{c|}{$ V = 0.01 $} & \multicolumn{1}{c|}{$ V = 0.15 $} & \multicolumn{1}{c|}{$ V = 0.01 $} & \multicolumn{1}{c}{$ V = 0.25 $} \\
		\hline
		Flags & 174.6 (1.00) & 163.9 (0.94) & 170.5 (0.98) & 88.2 (0.51) & 167.7 (0.96) & 19.1 (0.11) \\
		Emotions & 533.7 (1.00) & 532.9 (1.00) & 533.7 (1.00) & 532.1 (1.00) & 533.7 (1.00) & 244.6 (0.46) \\
		Birds & 580.5 (1.00) & 580.5 (1.00) & 580.5 (1.00) & 146.0 (0.25) & 580.0 (1.00) & 170.6 (0.29) \\
		Image & 1800.0 (1.00) & 1597.1 (0.89) & 1800.0 (1.00) & 1736.8 (0.96) & 1800.0 (1.00) & 1729.0 (0.96) \\
		Scene & 2159.0 (1.00) & 2076.3 (0.96) & 2159.0 (1.00) & 2151.0 (0.99) & 2159.0 (1.00) & 2148.0 (0.99) \\
		Yeast & 2175.3 (1.00) & 2123.1 (0.98) & 2175.3 (1.00) & 2168.8 (1.00) & 2175.3 (1.00) & 2015.6 (0.93) \\
		\hline
		VirusGO & 159.3 (0.86) & 11.2 (0.06) & 86.3 (0.46) & 5.2 (0.03) & 159.3 (0.86) & 122.1 (0.66) \\
		GpositiveGO & 412.3 (0.88) & 28.3 (0.06) & 183.0 (0.39) & 8.8 (0.02) & 412.3 (0.88) & 308.8 (0.66) \\
		Genbase & 192.0 (0.32) & 1.7 (0.00) & 31.9 (0.05) & 1.0 (0.00) & 192.0 (0.32) & 130.5 (0.22) \\
		Medical & 871.0 (0.99) & 55.0 (0.06) & 450.9 (0.51) & 14.0 (0.02) & 871.0 (0.99) & 736.5 (0.84) \\
		PlantGO & 775.6 (0.88) & 39.5 (0.04) & 301.5 (0.34) & 12.5 (0.01) & 766.4 (0.87) & 644.0 (0.73) \\
		Langlog & 1130.6 (0.86) & 761.4 (0.58) & 1116.7 (0.85) & 513.1 (0.39) & 1130.6 (0.86) & 1118.4 (0.85) \\
		\hline\hline
	\end{tabular}
	\\
	\vspace{1mm}
	\centering{
		\footnotesize \raggedright The value in the parentheses indicates the ratio to the number of generated nodes against the number of training instances.}
\end{table*}

\begin{itemize}
	\setlength{\leftskip}{-3mm}
	\vspace{-1mm}
	\item \textbf{MLCA}
	
	This algorithm can be the first-choice algorithm because it shows stable and high classification performance for both numerical and nominal datasets. In other words, it has an advantage if the attribute type of the dataset is unknown. Furthermore, in the case of a numerical dataset, MLCA shows high classification performance regardless of a specification of the similarity threshold $ V $. This means that MLCA can maintain high classification performance and high information compression performance, simultaneously.

	\vspace{1mm}
	\item \textbf{MLCA-I}
	
	This algorithm shows the outstanding classification performance on the numerical datasets. On the other hand, the classification performance on the nominal datasets is low in comparison with the results on to numerical datasets. Therefore, MLCA-I has an advantage if the dataset contains only numerical attributes.
	
	\vspace{1mm}
	\item \textbf{MLCA-C}
	
	This algorithm shows stable and high classification performance for both numerical and nominal datasets although it is not as good as MLCA. It is notable that MLCA-C stably has small rank values for both nominal and numerical datasets even when a specification of a similarity threshold $ V $ is large. Therefore, MLCA-C can achieve high classification performance and high information compression for both numerical and nominal datasets.
	
\end{itemize}

Table \ref{tab:characteristicsMLCA} summarizes the characteristics of MLCA and its variants based on the above analysis.

\subsection{Effects of a Multi-Epoch Learning Process}
\label{sec:multiEpochs}

MLCA, MLCA-I, and MLCA-C utilize generated nodes as classifiers. This means that the clustering performance on the training instances has a huge impact on the classification performance. The nodes of MLCA, MLCA-I, and MLCA-C are adaptively and continually generated/updated by the given instances. Therefore, it is possible to improve the clustering performance by learning the training instances in multiple epochs, and consequently to improve the classification performance of MLCA, MLCA-I, and MLCA-C. This feature is one of the advantages of MLCA, MLCA-I, and MLCA-C against the other compared algorithms.

\begin{figure*}[htbp]
	\centering
	\subfloat[Flags]{
		\includegraphics[width=1.2in]{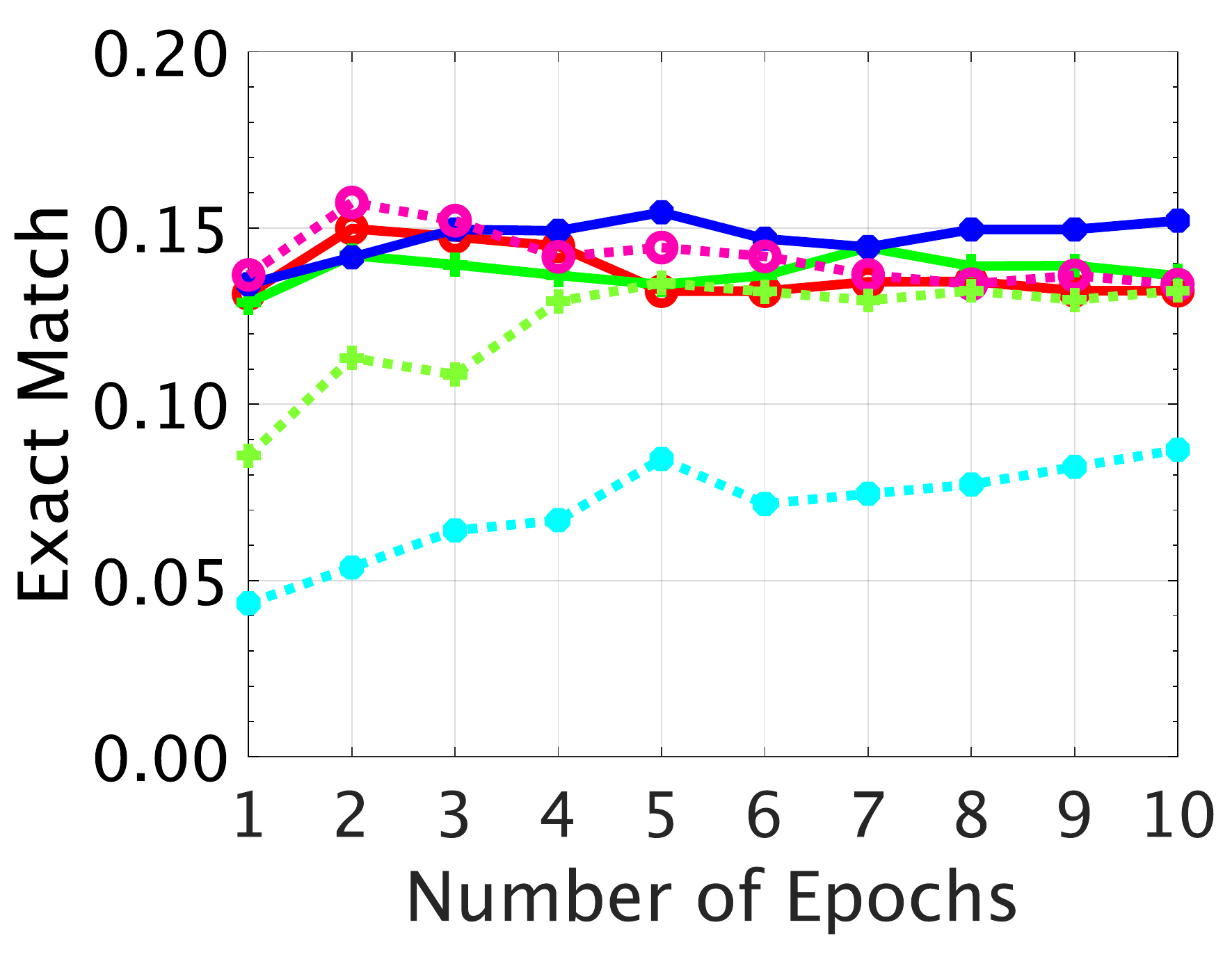}
		\label{fig:multi_flags}
	}
	\hfil
	\hspace{-4mm}
	\subfloat[Emotions]{
		\includegraphics[width=1.18in]{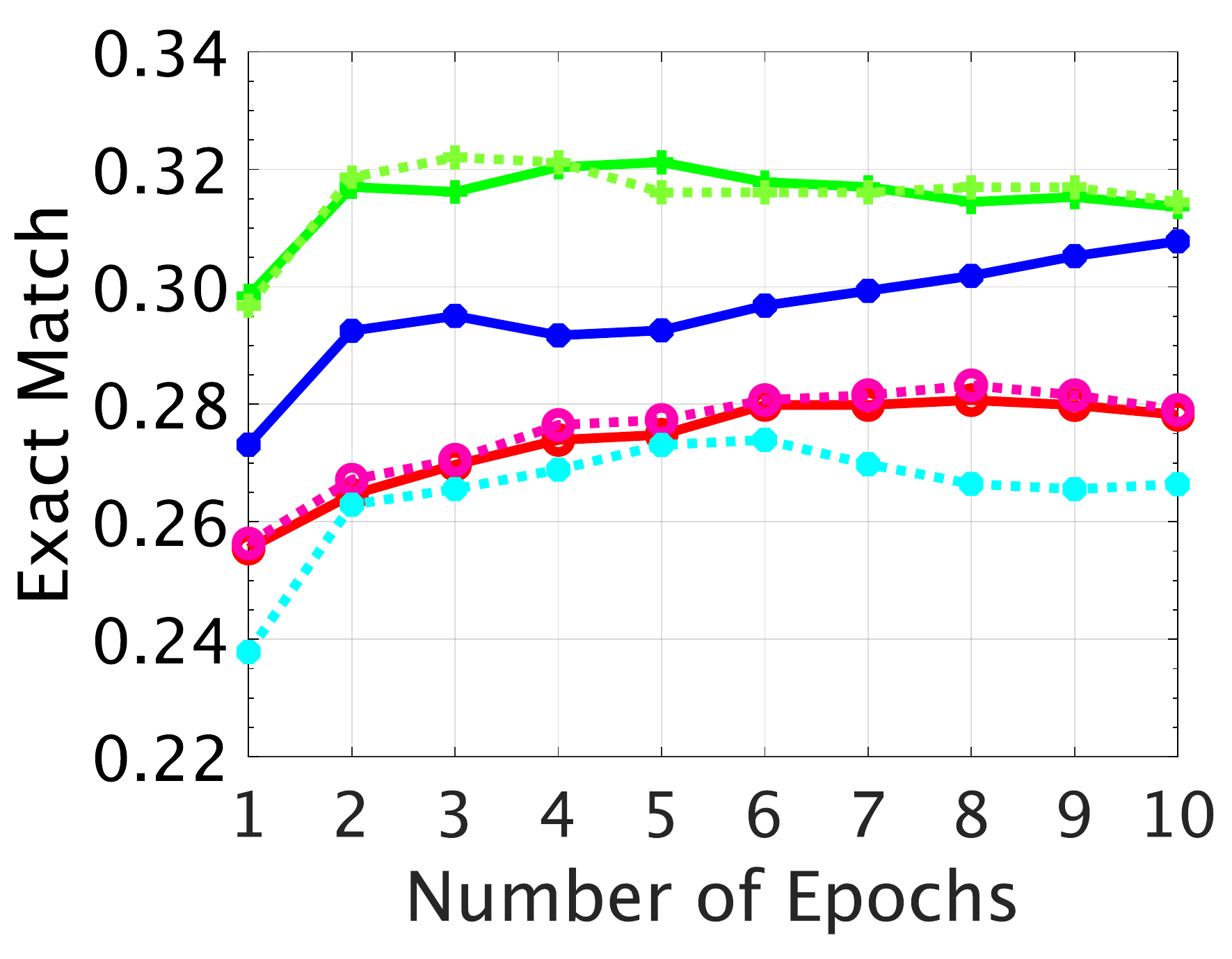}
		\label{fig:multi_emotions}
	}
	\hfil
	\hspace{-4mm}
	\subfloat[Birds]{
		\includegraphics[width=1.18in]{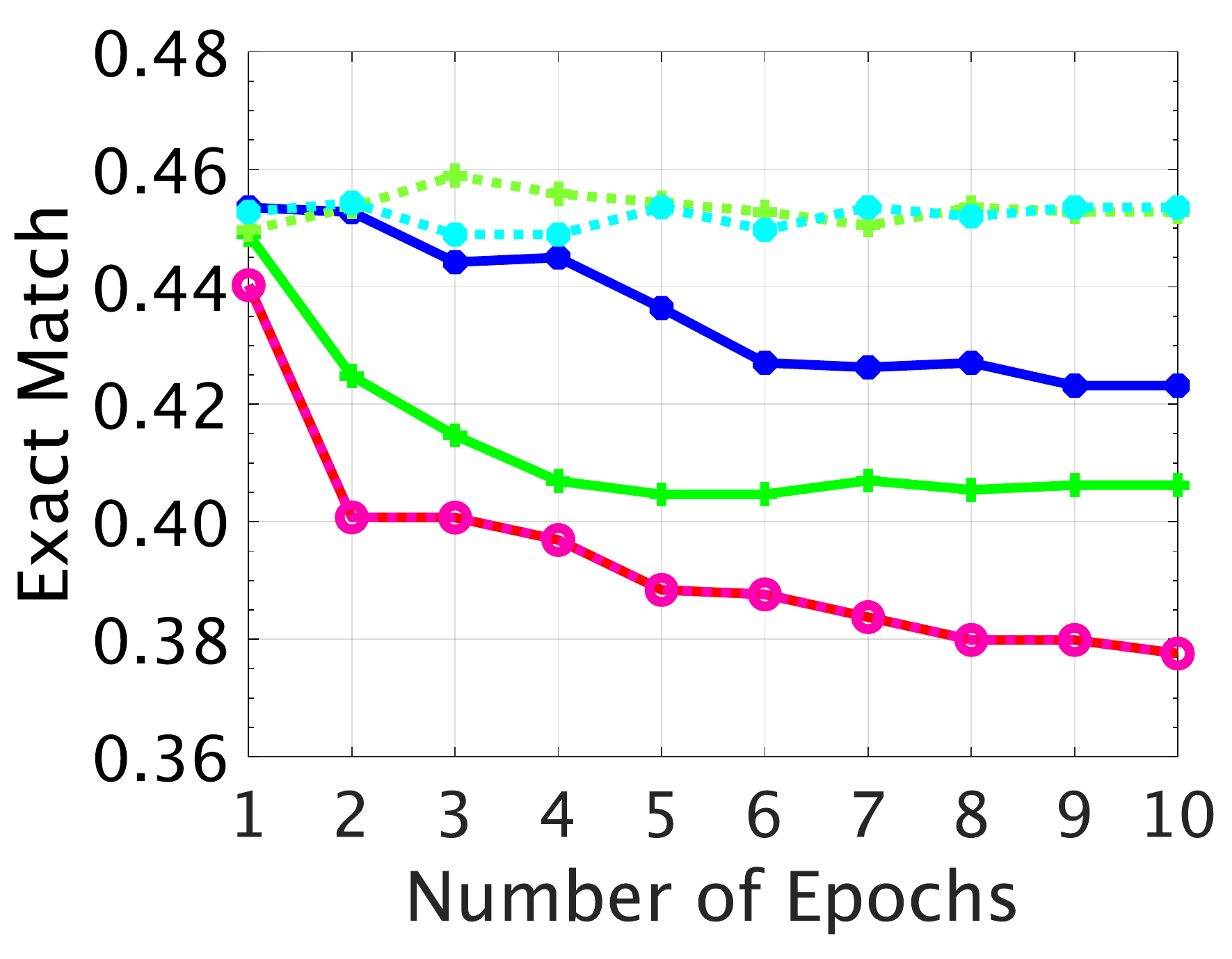}
		\label{fig:multi_birds}
	}
	\hfil
	\hspace{-4mm}
	\subfloat[Image]{
		\includegraphics[width=1.18in]{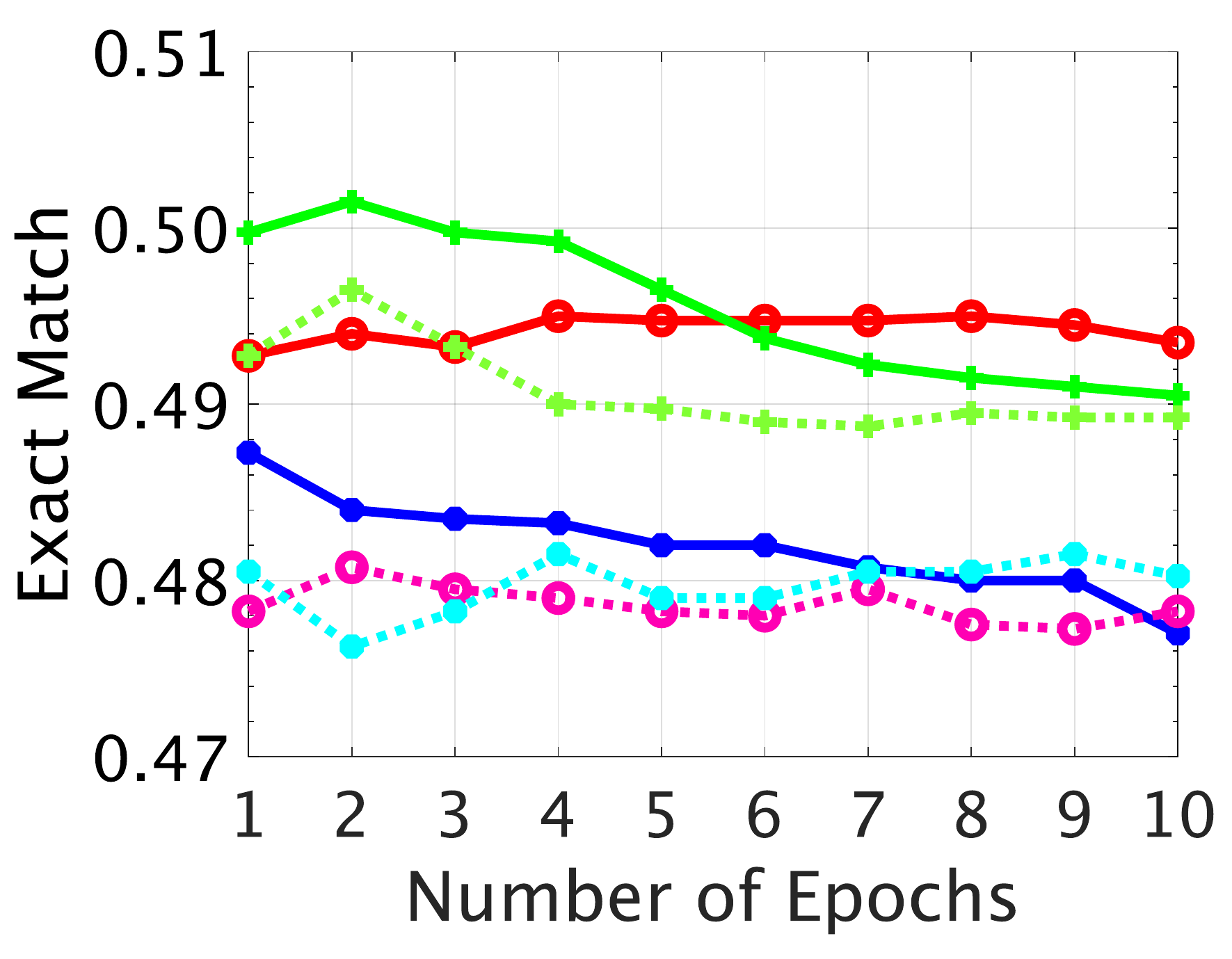}
		\label{fig:multi_Image}
	}
	\hfil
	\hspace{-4mm}
	\subfloat[Scene]{
		\includegraphics[width=1.18in]{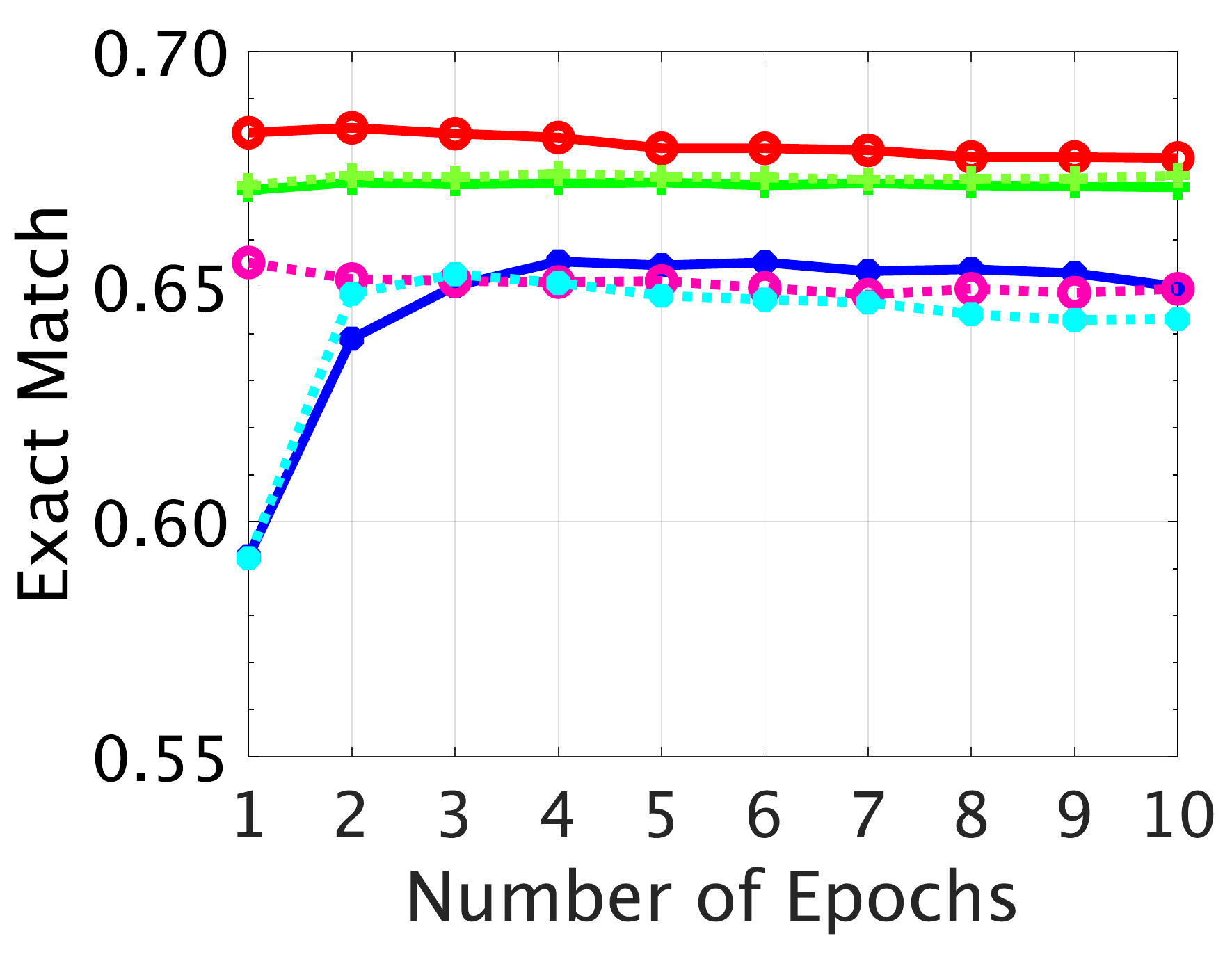}
		\label{fig:multi_scene}
	}
	\hfil
	\hspace{-4mm}
	\subfloat[Yeast]{
		\includegraphics[width=1.18in]{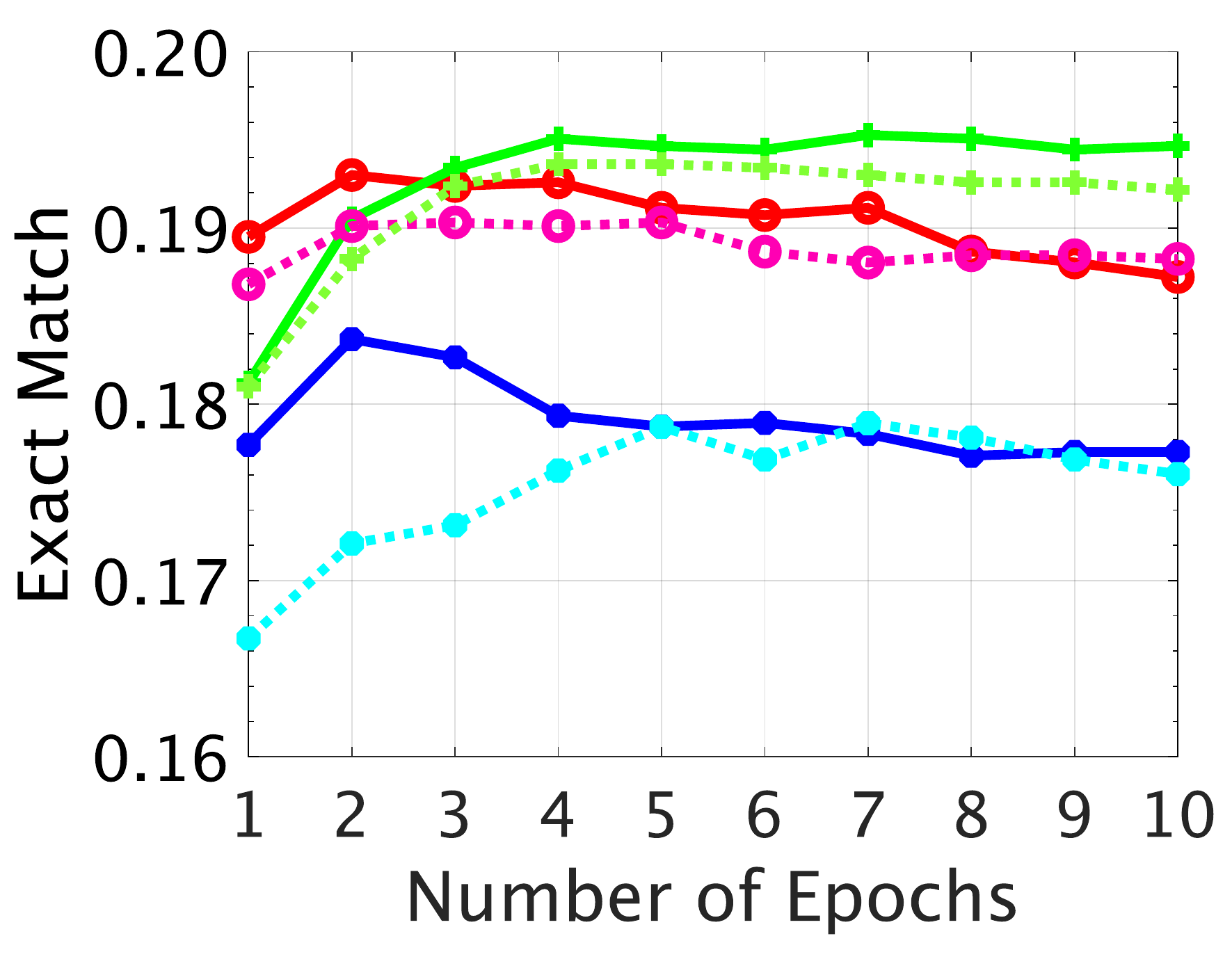}
		\label{fig:multi_yeast}
	}
	\\
	\vspace{-1mm}
	\subfloat[VirusGO]{
		\includegraphics[width=1.18in]{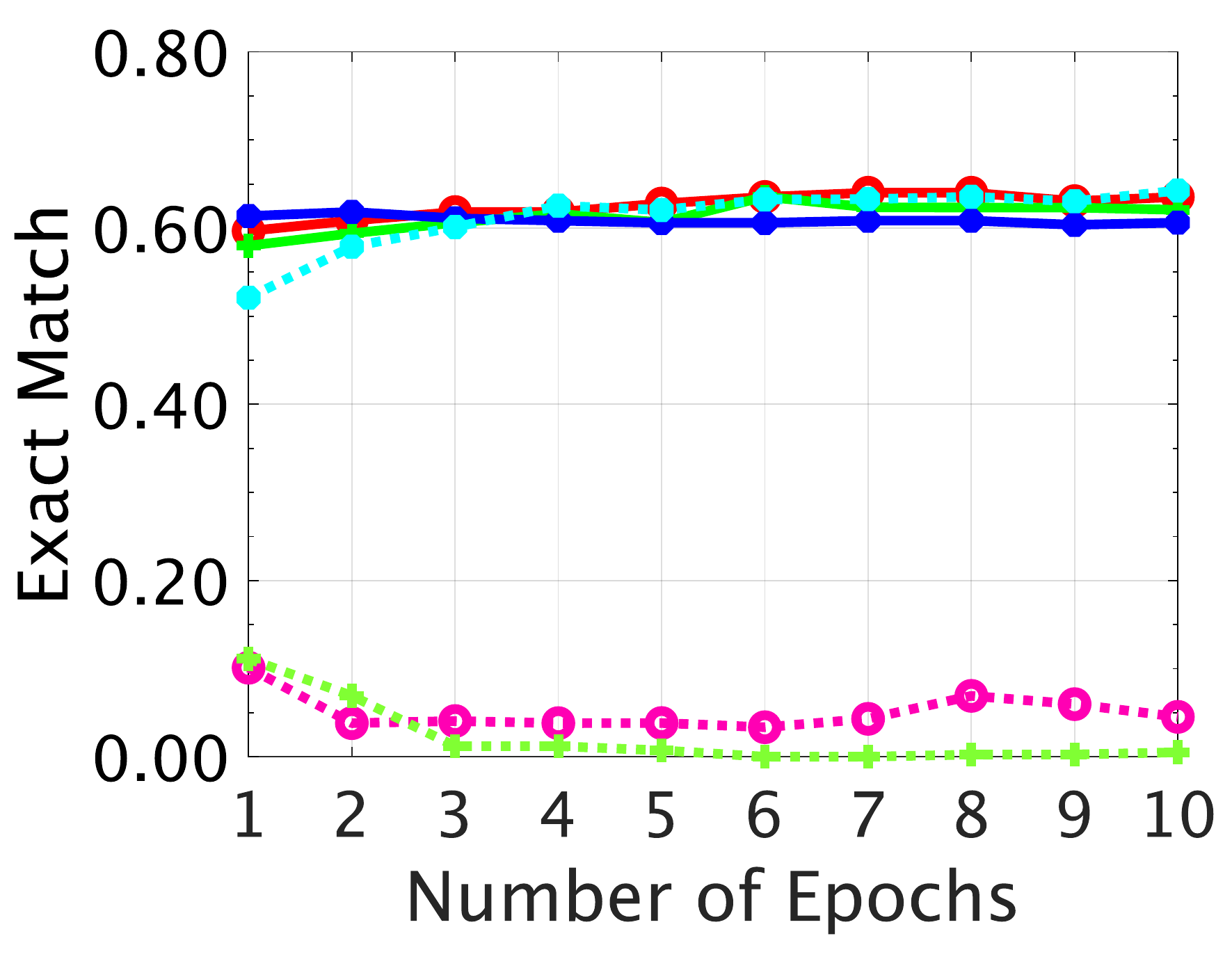}
		\label{fig:multi_virusgo}
	}
	\hfil
	\hspace{-4mm}
	\subfloat[GpositiveGO]{
		\includegraphics[width=1.18in]{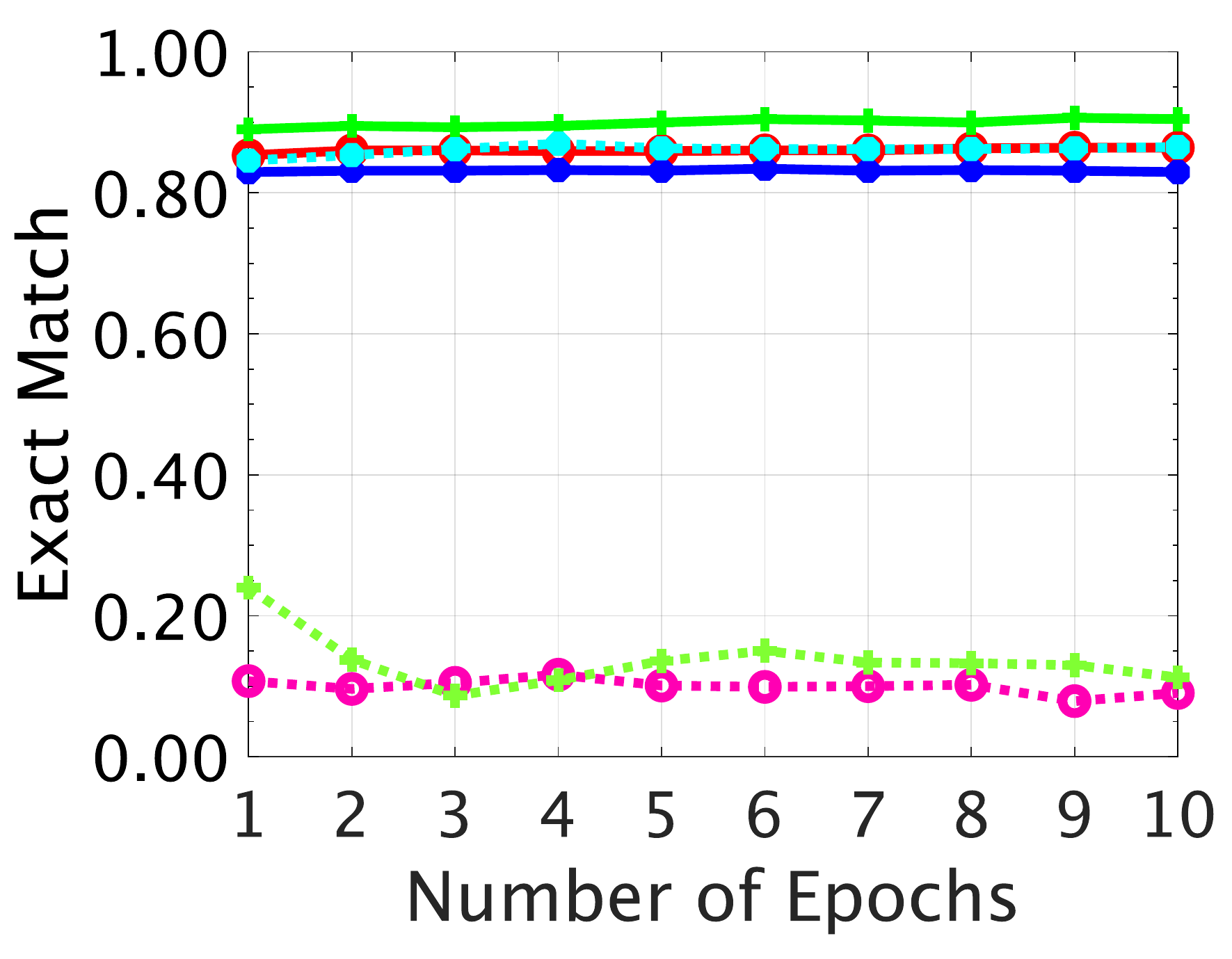}
		\label{fig:multi_gpositivego}
	}
	\hfil
	\hspace{-4mm}
	\subfloat[Genbase]{
		\includegraphics[width=1.18in]{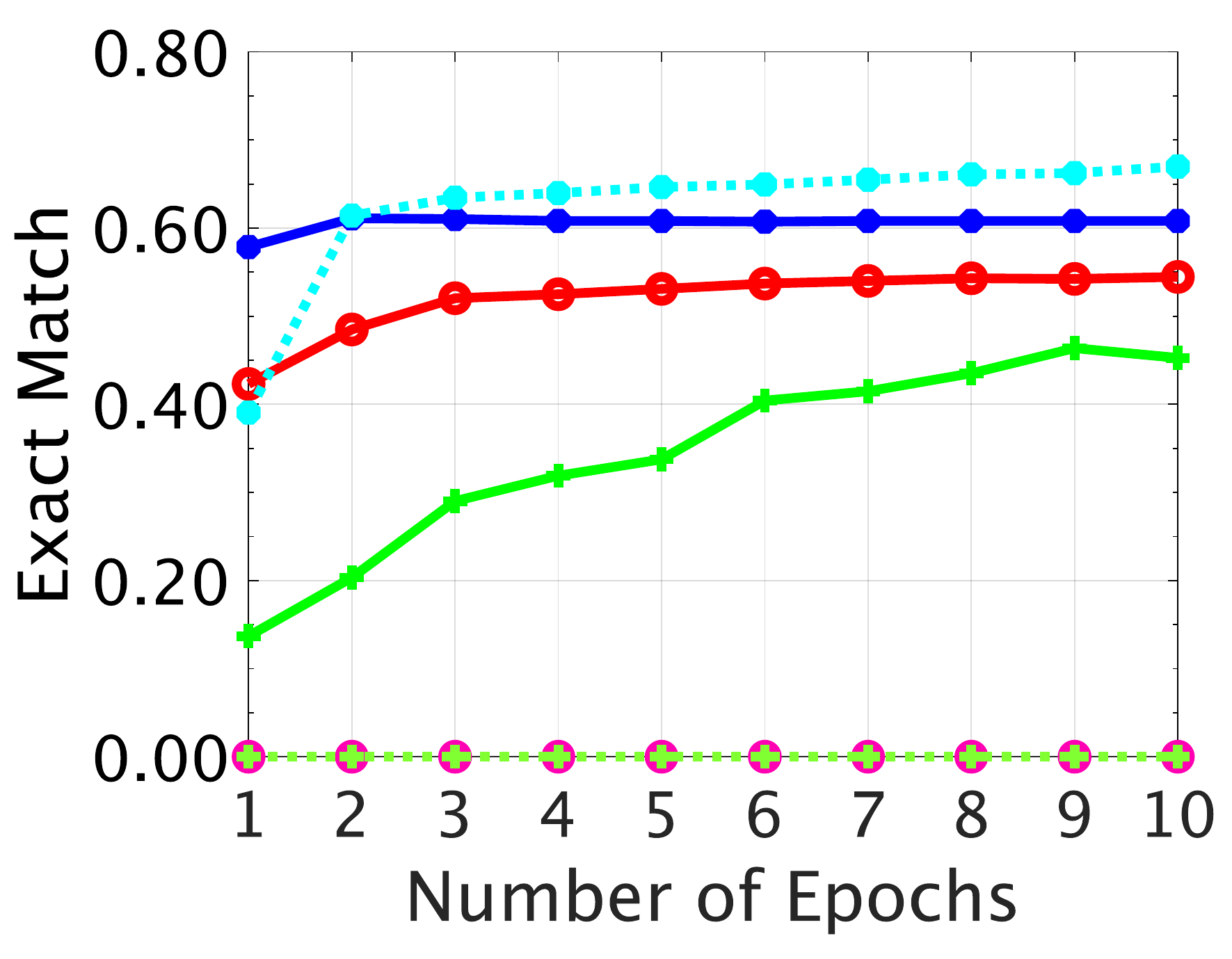}
		\label{fig:multi_genbase}
	}
	\hfil
	\hspace{-4mm}
	\subfloat[Medical]{
		\includegraphics[width=1.18in]{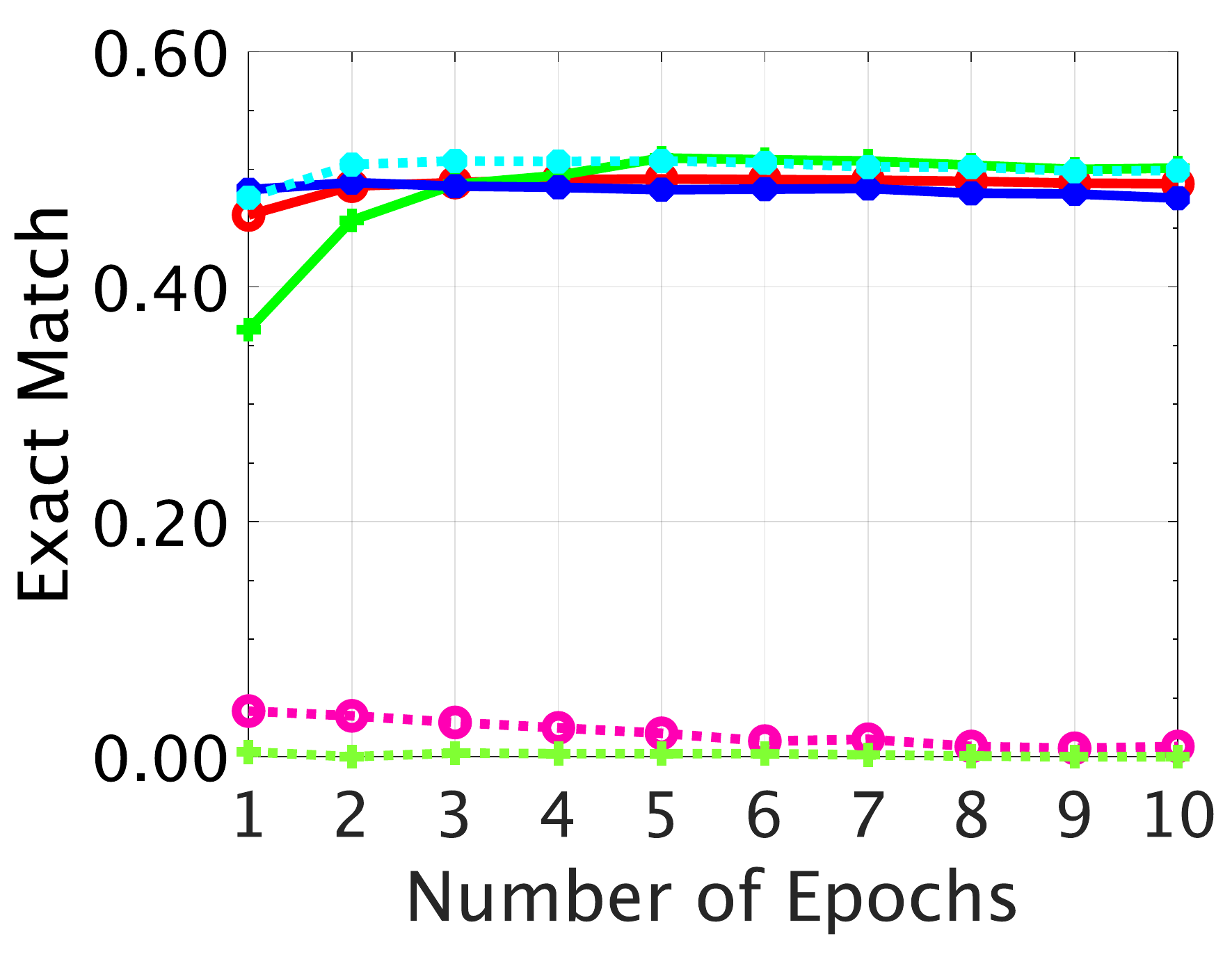}
		\label{fig:multi_medical}
	}
	\hfil
	\hspace{-4mm}
	\subfloat[PlantGO]{
		\includegraphics[width=1.18in]{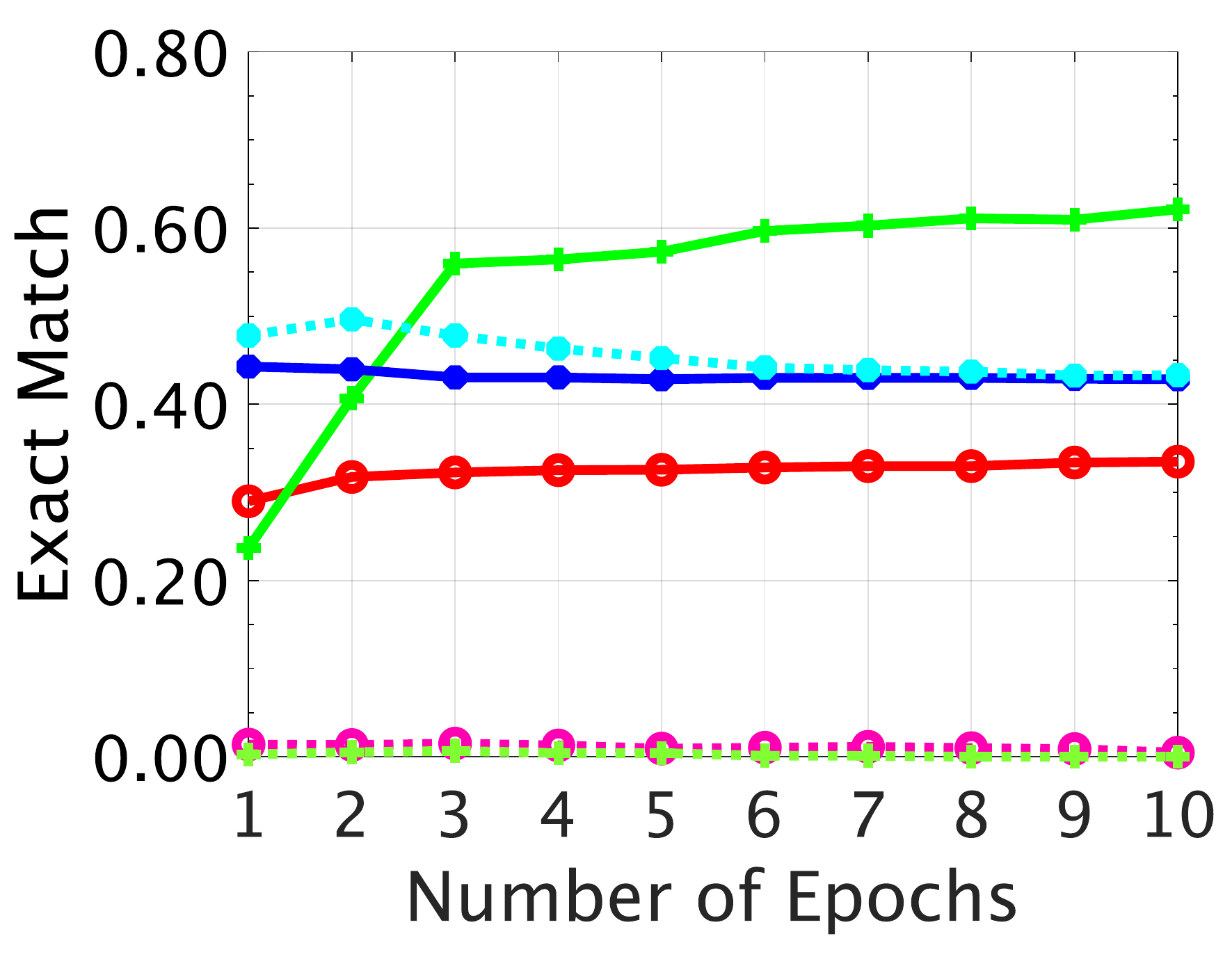}
		\label{fig:multi_plantgo}
	}
	\hfil
	\hspace{-4mm}
	\subfloat[Langlog]{
		\includegraphics[width=1.18in]{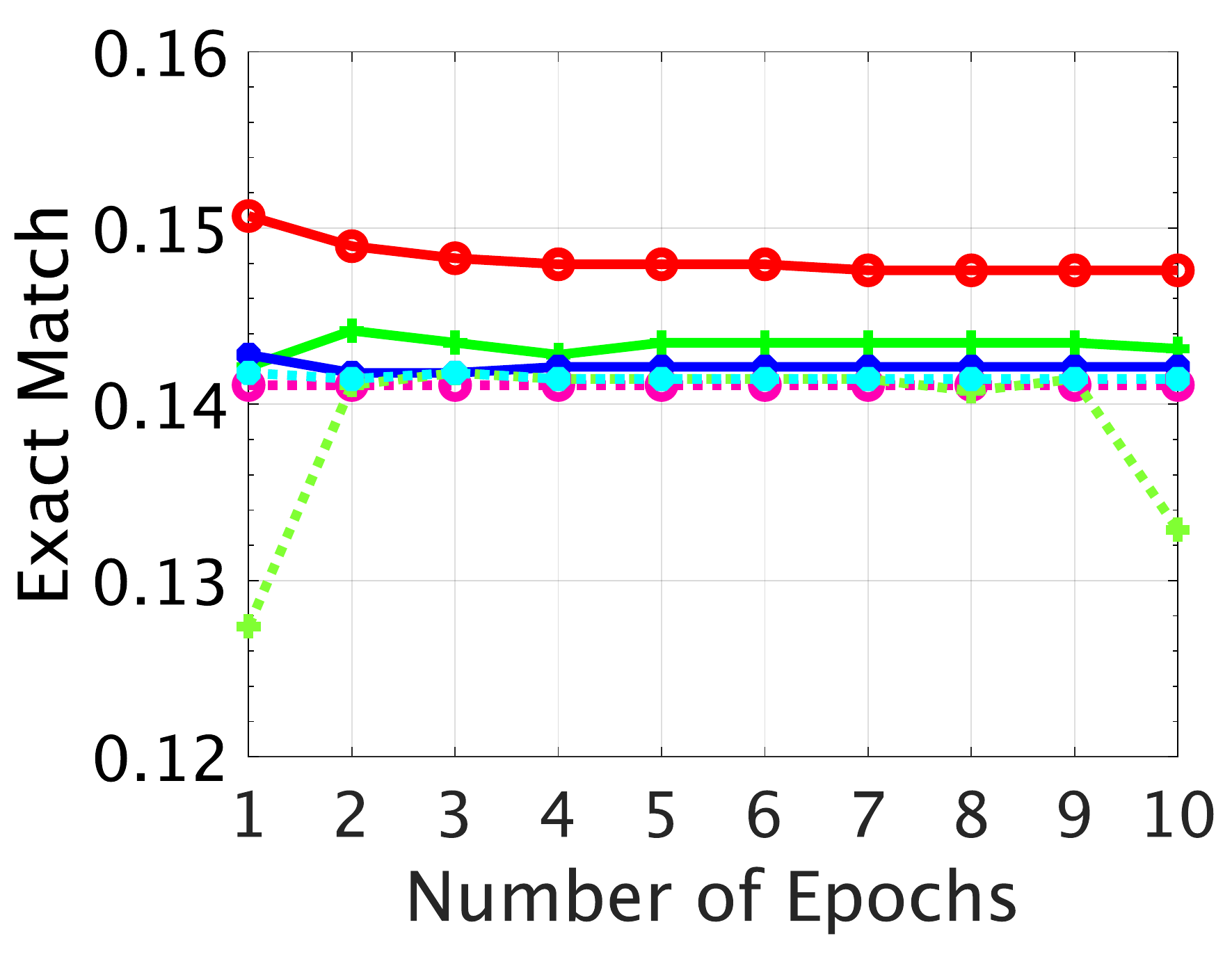}
		\label{fig:multi_langlog}
	}
	%
	\vspace{2mm}
	\includegraphics[width=3.8in]{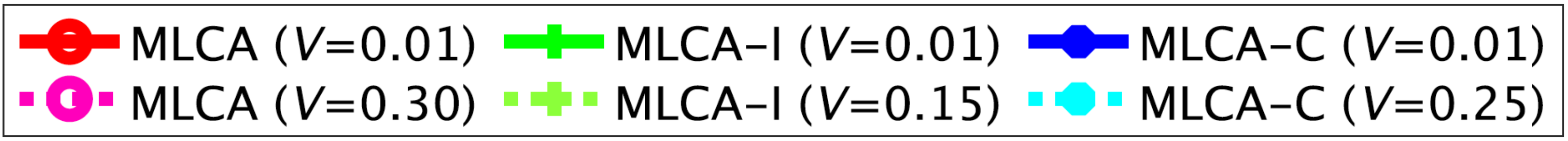}
	\caption{Results of the Exact Match of MLCA, MLCA-I, and MLCA-C with multiple epochs.}
	\label{fig:multiEpochs}
\end{figure*}

\begin{table}[htbp]
	\centering
	\caption{Summary of characteristics of MLCA and its variants}
	\renewcommand{\arraystretch}{1.2}
	\label{tab:characteristicsMLCA}
	\begin{tabular}{cccc}
		\hline\hline
		\multirow{2}{*}{Algorithm} & Similarity& \multicolumn{2}{l}{Classification Performance} \\
		\cline{3-4}
		& Threshold $ V $ & Numerical & Nominal \\
		\hline
		MLCA & Small & Very High & Very High \\
		& Large & Medium & Low \\
		\hline
		MLCA-I & Small & Very High & Low \\
		& Large & Medium & Low \\
		\hline
		MLCA-C & Small & High & High \\
		& Large & Medium & Medium \\
		\hline\hline
	\end{tabular}
\end{table}

We only apply the datasets listed in Table \ref{tab:datasetML} except for the large-scale datasets due to a time-consuming training process. Fig. \ref{fig:multiEpochs} shows results of the Exact Match of MLCA, MLCA-I, and MLCA-C with the learning of the training instances for 1 to 10 epochs. The conditions of this experiment are the same as in Section \ref{sec:quantitative}. 

The following observation is obtained: As the number of epochs increases, the value of the Exact Match increases or remains roughly the same in most cases except for the Birds dataset. In particular, it is effective for categorical data. Therefore, we regard that the multi-epoch learning process is generally beneficial for MLCA, MLCA-I, and MLCA-C.

\subsection{Computational Complexity}
\label{sec:compComplexity}
This section presents the computational complexity of MLCA, MLCA-I, MLCA-C, and comparison algorithms. Specifically, MLCA and its variants are analyzed in detail.

Regarding MLCA, the computational complexity of each process is as follows: For computing a bandwidth of a kernel function in the CIM is $ \mathcal{O}(\frac{n}{\lambda}d) $ (line 5 in Alg. \ref{al:LearnMLCA}), for computing the CIM is $ \mathcal{O}(ndK) $ (line 11 in Alg. \ref{al:LearnMLCA}), for sorting the result of the CIM is $ \mathcal{O}(K\log{K}) $ (line 12 in Alg. \ref{al:LearnMLCA}), and for computing the label probability is $ \mathcal{O}(N_{l}N_{y}) $ (line 24 in Alg. \ref{al:LearnMLCA}). Here, $ n $ is the number of training instances, $ \lambda $ is an interval for adapting $ \sigma $, $ K $ is the number of nodes, $ N_{l} $ is a size of a label set, and $ N_{y} $ is the predefined number of neighbor nodes for $ y_{k_{1}} $. Thus, the total computational complexity of MLCA is $ \mathcal{O}(  \frac{n}{\lambda}d + ndK + K\log{K} + N_{l}N_{y} ) $.

In terms of MLCA-I and MLCA-C, the difference of a training process is only in $ \mathrm{CIM}^{\text{I}} $ and $ \mathrm{CIM}^{\text{C}} $ which is defined in (\ref{eq:defcim_Individual}) and  (\ref{eq:defcim_Cluster}), respectively. Since $ \mathrm{CIM}^{\text{I}} $ considers an individual attribute of a training instance separately, it takes $ \mathcal{O}(nd^{2}K) $. $ \mathrm{CIM}^{\text{C}} $ applies a clustering approach to attributes of a training instance every $ \lambda $ instances. Thus, $ \mathcal{O}( \frac{n}{\lambda}(\frac{n}{\lambda}d + ndK + K\log{K}) ) $ is additionally required. As a result, the computational complexity of MLCA-I and MLCA-C are $ \mathcal{O}(  \frac{n}{\lambda}d + nd^{2}K + K\log{K} + N_{l}N_{y} ) $ and $ \mathcal{O}( (\frac{n}{\lambda} + 1)( \frac{n}{\lambda}d + ndK + K\log{K} ) + \frac{n}{\lambda}N_{l}N_{y} ) $, respectively.

Table \ref{tab:compComplexity} summarizes the computational complexity of MLCA, MLCA-I, MLCA-C, and comparison algorithms. Here, variables in the computational complexity of comparison algorithms are as follows:

{\bf MCIC}: $ T $ is a time period defined as $ T = \frac{1}{\lambda} \log( \frac{\beta_{\mu}}{(\beta_{\mu}-1)} + 1 ) $. Here, $ n $ is the number of instances, $ \lambda $ and $ \beta_{\mu} $ are the parameters of MCIC. $ K_{\text{p}} $ and $ K_{\text{o}} $ are the number of mature and immature clusters of MCIC, respectively. 

{\bf MuENL}: $ N_{l} $ is the size of a label set, $ n $ is the number of instances, and $ d $ is the dimension of instances. Note that the complexity of MuENL in Table \ref{tab:compComplexity} shows a pairwise label ranking classifier, not including a label incremental learning process.

{\bf mlODM}: $ n $ is the number of instances, $ N_{l} $ is the dimension of a current relevant label, and $ I $ is the number of iterations of an optimization process.

{\bf GLOCAL}: $ n $ is the number of instances, $ n_{m} $ is the number of instances of a partitioned training instances, $ k $ is a rank of a label matrix which satisfies $ k < N_{l} $.

{\bf MLSA-$ k $NN}: $ d $ is the dimension of a training instance, $ N_{l} $ is a size of a label set, $ m_\text{max} $ is the maximum size of the window, and $ m_\text{min} $ is the minimum size of the window.

{\bf ML-$ k $NN}: $ n $ is the number of training instances, $ d $ is the dimension of a training instance, $ k $ is the number of nearest neighbors of $ k $-NN, and $ N_{l} $ is a size of a label set. Here, $ \mathcal{O}( n^{2}d ) $ is for $ k $-NN computation, and $ \mathcal{O}( nkN_{l} ) $ is for the label probability computation.

With respect to the computational complexity of each algorithm, MLSA-$ k $NN and MCIC shows their superior computational efficiency than the other algorithms. MLCA, MLCA-I, MLCA-C, and ML-$ k $NN have moderate computational efficiency. These algorithms do not dramatically increase the computational complexity even when the number of instances and labels in a dataset is large. In contrast, MuENL, mlODM, and GLOCAL have a polynomial complexity in terms of the number of instances or the size of a label set. Therefore, we consider that these algorithms are time-consumimg and could not generate a predictive model in a valid time, especially for the large-scale datasets with a large number of training instances and labels.

\begin{table}[htbp]
	\vspace{2mm}
	\centering
	\caption{Summary of computational complexity}
	\renewcommand{\arraystretch}{1.2}
	\label{tab:compComplexity}
	\scalebox{0.97}{
		\begin{tabular}{llc}
			\hline\hline
			Algorithm & Computational Complexity & Ref.  \\
			\hline
			MLCA  & $ \mathcal{O}(  \frac{n}{\lambda}d + ndK + K\log{K} + N_{l}N_{y} ) $ & ---  \\
			MLCA-I  & $ \mathcal{O}(  \frac{n}{\lambda}d + nd^{2}K + K\log{K} + N_{l}N_{y} ) $ & ---   \\
			MLCA-C  & $ \mathcal{O}( (\frac{n}{\lambda} + 1)( \frac{n}{\lambda}d + ndK + K\log{K} ) + \frac{n}{\lambda}N_{l}N_{y} ) $ & --- \\
			\hline
			MCIC  & $ \mathcal{O}(n +  \frac{n}{T}(K_{\text{p}} + K_{\text{o}}) ) $ & ---  \\
			MuENL  & $ \mathcal{O}( ndN_{l}^{2} ) $ & \cite{zhu18a}   \\
			mlODM  & $ \mathcal{O}( nN_{l}^{2}I ) $ & \cite{tan20}  \\
			GLOCAL  & $ \mathcal{O}( n^{2} + n_{m}^{2} + kn ) $  & \cite{zhu20a} \\
			MLSA-$ k $NN  & $ \mathcal{O}( m_{\text{max}}d+m_{\text{max}}N_{l}+m_{\text{max}} \log_{2}\frac{m_\text{max}}{m_\text{min}} ) $  & \cite{roseberry21}  \\
			ML-$ k $NN  & $ \mathcal{O}( n^{2}d + nkN_{l} ) $ & \cite{skryjomski19}   \\
			\hline\hline
		\end{tabular}
	}
\end{table}

The training and testing time on a CPU are summarized in Tables 5 and 6 of the supplementary file.

\section{Concluding Remarks}
\label{sec:conclusion}
This paper proposed a multi-label classification algorithm capable of continual learning by extending our preliminary research \cite{masuyama20a}, namely MLCA. In addition, two variants of MLCA were proposed by modifying the calculation method of the CIM, namely MLCA-I and MLCA-C. The proposed algorithms consist of two components: The CIM-based ART and the Bayesian approach for label probability computation. Because both components can deal with a situation where new training instances and corresponding labels are sequentially provided, the proposed algorithms can realize continual learning. The results of extensive experiments from qualitative and quantitative perspectives showed that MLCA has competitive classification performance to other well-known algorithms while maintaining the continual learning ability. Furthermore, the results also showed that the performance of MLCA can be enhanced by modifying the calculation method of the CIM.

The ability to adapt to concept drift \cite{lu18} and to handle mixed numerical and categorical data \cite{ahmad19} are significant factors for clustering-based algorithms capable of continual learning. A future research topic is to examine the performance of the proposed algorithms under concept drift and to improve them. It is also an important future research topic to modify the proposed algorithms for mixed datasets with both numerical and categorical attributes.


%

%

  \section*{Acknowledgment}
This research was supported by Ministry of Education, Culture, Sports, Science and Technology - JAPAN (MEXT) Leading Initiative for Excellent Young Researchers (LEADER). The Universiti Malaya Impact-oriented Interdisciplinary Research Grant Programme (IIRG) - IIRG002C-19HWB, Universiti Malaya Covid-19 Related Special Research Grant (UMCSRG) CSRG008-2020ST from University of Malaya. National Natural Science Foundation of China (Grant No. 61876075), Guangdong Provincial Key Laboratory (Grant No. 2020B121201001), the Program for Guangdong Introducing Innovative and Enterpreneurial Teams (Grant No. 2017ZT07X386), The Stable Support Plan Program of Shenzhen Natural Science Fund (Grant No. 20200925174447003), Shenzhen Science and Technology Program (Grant No. KQTD2016112514355531).

\ifCLASSOPTIONcaptionsoff
  \newpage
\fi



\bibliographystyle{IEEEtran}
\bibliography{myref}

\begin{thebibliography}{10}
\providecommand{\url}[1]{#1}
\csname url@samestyle\endcsname
\providecommand{\newblock}{\relax}
\providecommand{\bibinfo}[2]{#2}
\providecommand{\BIBentrySTDinterwordspacing}{\spaceskip=0pt\relax}
\providecommand{\BIBentryALTinterwordstretchfactor}{4}
\providecommand{\BIBentryALTinterwordspacing}{\spaceskip=\fontdimen2\font plus
\BIBentryALTinterwordstretchfactor\fontdimen3\font minus
  \fontdimen4\font\relax}
\providecommand{\BIBforeignlanguage}[2]{{%
\expandafter\ifx\csname l@#1\endcsname\relax
\typeout{** WARNING: IEEEtran.bst: No hyphenation pattern has been}%
\typeout{** loaded for the language `#1'. Using the pattern for}%
\typeout{** the default language instead.}%
\else
\language=\csname l@#1\endcsname
\fi
#2}}
\providecommand{\BIBdecl}{\relax}
\BIBdecl

\bibitem{parisi19}
G.~I. Parisi, R.~Kemker, J.~L. Part, C.~Kanan, and S.~Wermter, ``Continual
  lifelong learning with neural networks: A review,'' \emph{Neural Networks},
  vol. 113, pp. 57--71, 2019.

\bibitem{zhang18}
J.~Zhang, C.~Li, D.~Cao, Y.~Lin, S.~Su, L.~Dai, and S.~Li, ``Multi-label
  learning with label-specific features by resolving label correlations,''
  \emph{Knowledge-Based Systems}, vol. 159, pp. 148--157, 2018.

\bibitem{han19}
H.~Han, M.~Huang, Y.~Zhang, X.~Yang, and W.~Feng, ``Multi-label learning with
  label specific features using correlation information,'' \emph{IEEE Access},
  vol.~7, pp. 11\,474--11\,484, 2019.

\bibitem{he19}
L.~He, L.~Xie, H.~Shu, and S.~Hu, ``Discrete semi-supervised learning for
  multi-label image classification and large-scale image retrieval,''
  \emph{Multimedia Tools and Applications}, vol.~78, no.~17, pp.
  24\,519--24\,537, 2019.

\bibitem{zhang21}
C.~Zhang and Z.~Li, ``Multi-label learning with label-specific features via
  weighting and label entropy guided clustering ensemble,''
  \emph{Neurocomputing}, vol. 419, pp. 59--69, 2021.

\bibitem{zheng19}
X.~Zheng, P.~Li, Z.~Chu, and X.~Hu, ``A survey on multi-label data stream
  classification,'' \emph{IEEE Access}, vol.~8, pp. 1249--1275, 2019.

\bibitem{zhang07}
M.-L. Zhang and Z.-H. Zhou, ``{ML-KNN}: A lazy learning approach to multi-label
  learning,'' \emph{Pattern Recognition}, vol.~40, no.~7, pp. 2038--2048, 2007.

\bibitem{zhu18a}
Y.~Zhu, K.~M. Ting, and Z.-H. Zhou, ``Multi-label learning with emerging new
  labels,'' \emph{IEEE Transactions on Knowledge and Data Engineering},
  vol.~30, no.~10, pp. 1901--1914, 2018.

\bibitem{furao06}
S.~Furao and O.~Hasegawa, ``{An incremental network for on-line unsupervised
  classification and topology learning},'' \emph{Neural Networks}, vol.~19,
  no.~1, pp. 90--106, 2006.

\bibitem{shen08}
F.~Shen and O.~Hasegawa, ``{A fast nearest neighbor classifier based on
  self-organizing incremental neural network},'' \emph{Neural Networks},
  vol.~21, no.~10, pp. 1537--1547, 2008.

\bibitem{masuyama18}
N.~Masuyama, C.~K. Loo, and F.~Dawood, ``{Kernel {B}ayesian {ART} and
  {ARTMAP}},'' \emph{Neural Networks}, vol.~98, pp. 76--86, 2018.

\bibitem{masuyama19a}
N.~Masuyama, C.~K. Loo, and S.~Wermter, ``A kernel {B}ayesian adaptive
  resonance theory with a topological structure,'' \emph{International Journal
  of Neural Systems}, vol.~29, no.~5, p. 1850052 (20 pages), 2019.

\bibitem{masuyama19b}
N.~Masuyama, C.~K. Loo, H.~Ishibuchi, N.~Kubota, Y.~Nojima, and Y.~Liu,
  ``Topological clustering via adaptive resonance theory with information
  theoretic learning,'' \emph{IEEE Access}, vol.~7, pp. 76\,920--76\,936, 2019.

\bibitem{masuyamaFTCA}
N.~Masuyama, N.~Amako, Y.~Nojima, Y.~Liu, C.~K. Loo, and H.~Ishibuchi, ``Fast
  topological adaptive resonance theory based on correntropy induced metric,''
  in \emph{2019 IEEE Symposium Series on Computational Intelligence (SSCI)},
  2019, pp. 2215--2221.

\bibitem{liu07}
W.~Liu, P.~P. Pokharel, and J.~C. Pr{\'\i}ncipe, ``{Correntropy: Properties and
  applications in non-{G}aussian signal processing},'' \emph{IEEE Transactions
  on Signal Processing}, vol.~55, no.~11, pp. 5286--5298, 2007.

\bibitem{mclachlan19}
G.~J. McLachlan, S.~X. Lee, and S.~I. Rathnayake, ``Finite mixture models,''
  \emph{Annual Review of Statistics and its Application}, vol.~6, pp. 355--378,
  2019.

\bibitem{lloyd82}
S.~Lloyd, ``{Least squares quantization in {PCM}},'' \emph{IEEE Transactions on
  Information Theory}, vol.~28, no.~2, pp. 129--137, 1982.

\bibitem{kohonen82}
T.~Kohonen, ``{Self-organized formation of topologically correct feature
  maps},'' \emph{Biological Cybernetics}, vol.~43, no.~1, pp. 59--69, 1982.

\bibitem{fritzke95}
B.~Fritzke, ``{A growing neural gas network learns topologies},''
  \emph{Advances in Neural Information Processing Systems}, vol.~7, pp.
  625--632, 1995.

\bibitem{carpenter88}
G.~A. Carpenter and S.~Grossberg, ``{The ART of adaptive pattern recognition by
  a self-organizing neural network},'' \emph{Computer}, vol.~21, no.~3, pp.
  77--88, 1988.

\bibitem{marsland02}
S.~Marsland, J.~Shapiro, and U.~Nehmzow, ``{A self-organising network that
  grows when required},'' \emph{Neural Networks}, vol.~15, no.~8, pp.
  1041--1058, 2002.

\bibitem{grossberg87}
S.~Grossberg, ``{Competitive learning: From interactive activation to adaptive
  resonance},'' \emph{Cognitive Science}, vol.~11, no.~1, pp. 23--63, 1987.

\bibitem{tan14}
S.~C. Tan, J.~Watada, Z.~Ibrahim, and M.~Khalid, ``Evolutionary fuzzy {ARTMAP}
  neural networks for classification of semiconductor defects,'' \emph{IEEE
  Transactions on Neural Networks and Learning Systems}, vol.~26, no.~5, pp.
  933--950, 2014.

\bibitem{matias18}
A.~L. Matias and A.~R.~R. Neto, ``On{ARTMAP}: A fuzzy {ARTMAP}-based
  architecture,'' \emph{Neural Networks}, vol.~98, pp. 236--250, 2018.

\bibitem{matias21}
A.~L. Matias, A.~R.~R. Neto, C.~L.~C. Mattos, and J.~P.~P. Gomes, ``A novel
  fuzzy {ARTMAP} with area of influence,'' \emph{Neurocomputing}, vol. 432, pp.
  80--90, 2021.

\bibitem{carpenter91b}
G.~A. Carpenter, S.~Grossberg, and D.~B. Rosen, ``{Fuzzy {ART}: Fast stable
  learning and categorization of analog patterns by an adaptive resonance
  system},'' \emph{Neural Networks}, vol.~4, no.~6, pp. 759--771, 1991.

\bibitem{vigdor07}
B.~Vigdor and B.~Lerner, ``{The {B}ayesian {ARTMAP}},'' \emph{IEEE Transactions
  on Neural Networks}, vol.~18, no.~6, pp. 1628--1644, 2007.

\bibitem{wang19}
L.~Wang, H.~Zhu, J.~Meng, and W.~He, ``Incremental local distribution-based
  clustering using {B}ayesian adaptive resonance theory,'' \emph{IEEE
  Transactions on Neural Networks and Learning Systems}, vol.~30, no.~11, pp.
  3496--3504, 2019.

\bibitem{da20}
L.~E.~B. da~Silva, I.~Elnabarawy, and D.~C. Wunsch~II, ``Distributed dual
  vigilance fuzzy adaptive resonance theory learns online, retrieves
  arbitrarily-shaped clusters, and mitigates order dependence,'' \emph{Neural
  Networks}, vol. 121, pp. 208--228, 2020.

\bibitem{zhang13}
M.-L. Zhang and Z.-H. Zhou, ``A review on multi-label learning algorithms,''
  \emph{IEEE Transactions on Knowledge and Data Engineering}, vol.~26, no.~8,
  pp. 1819--1837, 2013.

\bibitem{boutell04}
M.~R. Boutell, J.~Luo, X.~Shen, and C.~M. Brown, ``Learning multi-label scene
  classification,'' \emph{Pattern Recognition}, vol.~37, no.~9, pp. 1757--1771,
  2004.

\bibitem{tsoumakas10}
G.~Tsoumakas, I.~Katakis, and I.~Vlahavas, ``Random {\it k}-labelsets for
  multilabel classification,'' \emph{IEEE Transactions on Knowledge and Data
  Engineering}, vol.~23, no.~7, pp. 1079--1089, 2010.

\bibitem{liu16}
H.~Liu, X.~Wu, and S.~Zhang, ``Neighbor selection for multilabel
  classification,'' \emph{Neurocomputing}, vol. 182, pp. 187--196, 2016.

\bibitem{clare01}
A.~Clare and R.~D. King, ``Knowledge discovery in multi-label phenotype data,''
  in \emph{Principles of Data Mining and Knowledge Discovery}, 2001, pp.
  42--53.

\bibitem{osojnik17}
A.~Osojnik, P.~Panov, and S.~D{\v{z}}eroski, ``Multi-label classification via
  multi-target regression on data streams,'' \emph{Machine Learning}, vol. 106,
  no.~6, pp. 745--770, 2017.

\bibitem{elisseeff01}
A.~Elisseeff and J.~Weston, ``A kernel method for multi-labelled
  classification,'' in \emph{14th International Conference on Neural
  Information Processing Systems: Natural and Synthetic}, vol.~14, no.~7, 2001,
  pp. 681--687.

\bibitem{zhang16_multi}
N.~Zhang, S.~Ding, and J.~Zhang, ``Multi layer {ELM-RBF} for multi-label
  learning,'' \emph{Applied Soft Computing}, vol.~43, pp. 535--545, 2016.

\bibitem{zhang20}
H.~Zhang, J.~Yang, G.~Jia, S.~Han, and X.~Zhou, ``{ELM-MC}: multi-label
  classification framework based on extreme learning machine,''
  \emph{International Journal of Machine Learning and Cybernetics}, pp. 1--14,
  2020.

\bibitem{zhu18}
Y.~Zhu, J.~T. Kwok, and Z.-H. Zhou, ``Multi-label learning with global and
  local label correlation,'' \emph{IEEE Transactions on Knowledge and Data
  Engineering}, vol.~30, no.~6, pp. 1081--1094, 2018.

\bibitem{tan20}
Z.-H. Tan, P.~Tan, Y.~Jiang, and Z.-H. Zhou, ``Multi-label optimal margin
  distribution machine,'' \emph{Machine Learning}, vol. 109, no.~3, pp.
  623--642, 2020.

\bibitem{roseberry21}
M.~Roseberry, B.~Krawczyk, Y.~Djenouri, and A.~Cano, ``Self-adjusting {\it k}
  nearest neighbors for continual learning from multi-label drifting data
  streams,'' \emph{Neurocomputing}, vol. 442, pp. 10--25, 2021.

\bibitem{colombini17}
G.~G. Colombini, I.~B.~M. de~Abreu, and R.~Cerri, ``A self-organizing map-based
  method for multi-label classification,'' in \emph{2017 International Joint
  Conference on Neural Networks (IJCNN)}, 2017, pp. 4291--4298.

\bibitem{benyettou17}
A.~Benyettou, Y.~Bennani, A.~Bendahmane, and G.~Cabanes, ``Semisupervised
  multi-label classification through topological active learning,''
  \emph{International Journal on Communications Antenna and Propagation (I. Re.
  CAP)}, vol.~7, no.~3, pp. 222--232, 2017.

\bibitem{boulbazine18}
S.~Boulbazine, G.~Cabanes, B.~Matei, and Y.~Bennani, ``Online semi-supervised
  growing neural gas for multi-label data classification,'' in \emph{2018
  International Joint Conference on Neural Networks (IJCNN)}, 2018, pp. 1--8,
  doi: 10.1109/IJCNN.2018.8\,489\,776.

\bibitem{nguyen19}
T.~T. Nguyen, M.~T. Dang, A.~V. Luong, A.~W.-C. Liew, T.~Liang, and J.~McCall,
  ``Multi-label classification via incremental clustering on an evolving data
  stream,'' \emph{Pattern Recognition}, vol.~95, pp. 96--113, 2019.

\bibitem{carpenter92}
G.~A. Carpenter, S.~Grossberg, N.~Markuzon, J.~H. Reynolds, and D.~B. Rosen,
  ``{Fuzzy {ARTMAP}: A neural network architecture for incremental supervised
  learning of analog multidimensional maps},'' \emph{IEEE Transactions on
  Neural Networks}, vol.~3, no.~5, pp. 698--713, 1992.

\bibitem{benites10}
F.~Benites, F.~Brucker, and E.~Sapozhnikova, ``Multi-label classification by
  {ART}-based neural networks and hierarchy extraction,'' in \emph{The 2010
  International Joint Conference on Neural Networks (IJCNN)}, 2010, pp. 1--9.

\bibitem{benites17}
F.~Benites and E.~Sapozhnikova, ``Improving scalability of {ART} neural
  networks,'' \emph{Neurocomputing}, vol. 230, pp. 219--229, 2017.

\bibitem{yuan17}
L.~X. Yuan, S.~C. Tan, P.~Y. Goh, C.~P. Lim, and J.~Watada, ``Fuzzy {ARTMAP}
  with binary relevance for multi-label classification,'' in
  \emph{International Conference on Intelligent Decision Technologies}, 2017,
  pp. 127--135.

\bibitem{park18}
J.-Y. Park and J.-H. Kim, ``Incremental class learning for hierarchical
  classification,'' \emph{IEEE Transactions on Cybernetics}, vol.~50, no.~1,
  pp. 178--189, 2018.

\bibitem{marriott95}
S.~Marriott and R.~F. Harrison, ``{A modified fuzzy ARTMAP architecture for the
  approximation of noisy mappings},'' \emph{Neural Networks}, vol.~8, no.~4,
  pp. 619--641, 1995.

\bibitem{henderson12}
D.~J. Henderson and C.~F. Parmeter, ``Normal reference bandwidths for the
  general order, multivariate kernel density derivative estimator,''
  \emph{Statistics \& Probability Letters}, vol.~82, no.~12, pp. 2198--2205,
  2012.

\bibitem{tsoumakas11}
G.~Tsoumakas, E.~Spyromitros-Xioufis, J.~Vilcek, and I.~Vlahavas, ``Mulan: A
  java library for multi-label learning,'' \emph{The Journal of Machine
  Learning Research}, vol.~12, pp. 2411--2414, 2011.

\bibitem{bhatia16}
K.~Bhatia, K.~Dahiya, H.~Jain, P.~Kar, A.~Mittal, Y.~Prabhu, and M.~Varma,
  ``The extreme classification repository: {M}ulti-label datasets and code,''
  \url{http://manikvarma.org/downloads/XC/XMLRepository.html}, [Online;
  accessed 20-August-2021].

\bibitem{masuyama20a}
N.~Masuyama, Y.~Nojima, C.~K. Loo, and H.~Ishibuchi, ``Multi-label
  classification based on adaptive resonance theory,'' in \emph{2020 IEEE
  Symposium Series on Computational Intelligence (SSCI)}, 2020, pp. 1913--1920.

\bibitem{demvsar06}
J.~Dem{\v{s}}ar, ``Statistical comparisons of classifiers over multiple data
  sets,'' \emph{Journal of Machine Learning Research}, vol.~7, no.~1, pp.
  1--30, 2006.

\bibitem{zhu20a}
C.~Zhu, D.~Miao, Z.~Wang, R.~Zhou, L.~Wei, and X.~Zhang, ``Global and local
  multi-view multi-label learning,'' \emph{Neurocomputing}, vol. 371, pp.
  67--77, 2020.

\bibitem{skryjomski19}
P.~Skryjomski, B.~Krawczyk, and A.~Cano, ``Speeding up {\it k}-nearest
  neighbors classifier for large-scale multi-label learning on {GPU}s,''
  \emph{Neurocomputing}, vol. 354, pp. 10--19, 2019.

\bibitem{lu18}
J.~Lu, A.~Liu, F.~Dong, F.~Gu, J.~Gama, and G.~Zhang, ``Learning under concept
  drift: {A} review,'' \emph{IEEE Transactions on Knowledge and Data
  Engineering}, vol.~31, no.~12, pp. 2346--2363, 2018.

\bibitem{ahmad19}
A.~Ahmad and S.~S. Khan, ``Survey of state-of-the-art mixed data clustering
  algorithms,'' \emph{IEEE Access}, vol.~7, pp. 31\,883--31\,902, 2019.

\end{thebibliography}

%
%

%

\begin{IEEEbiography}[{\includegraphics[width=1in,height=1.25in,clip,keepaspectratio]{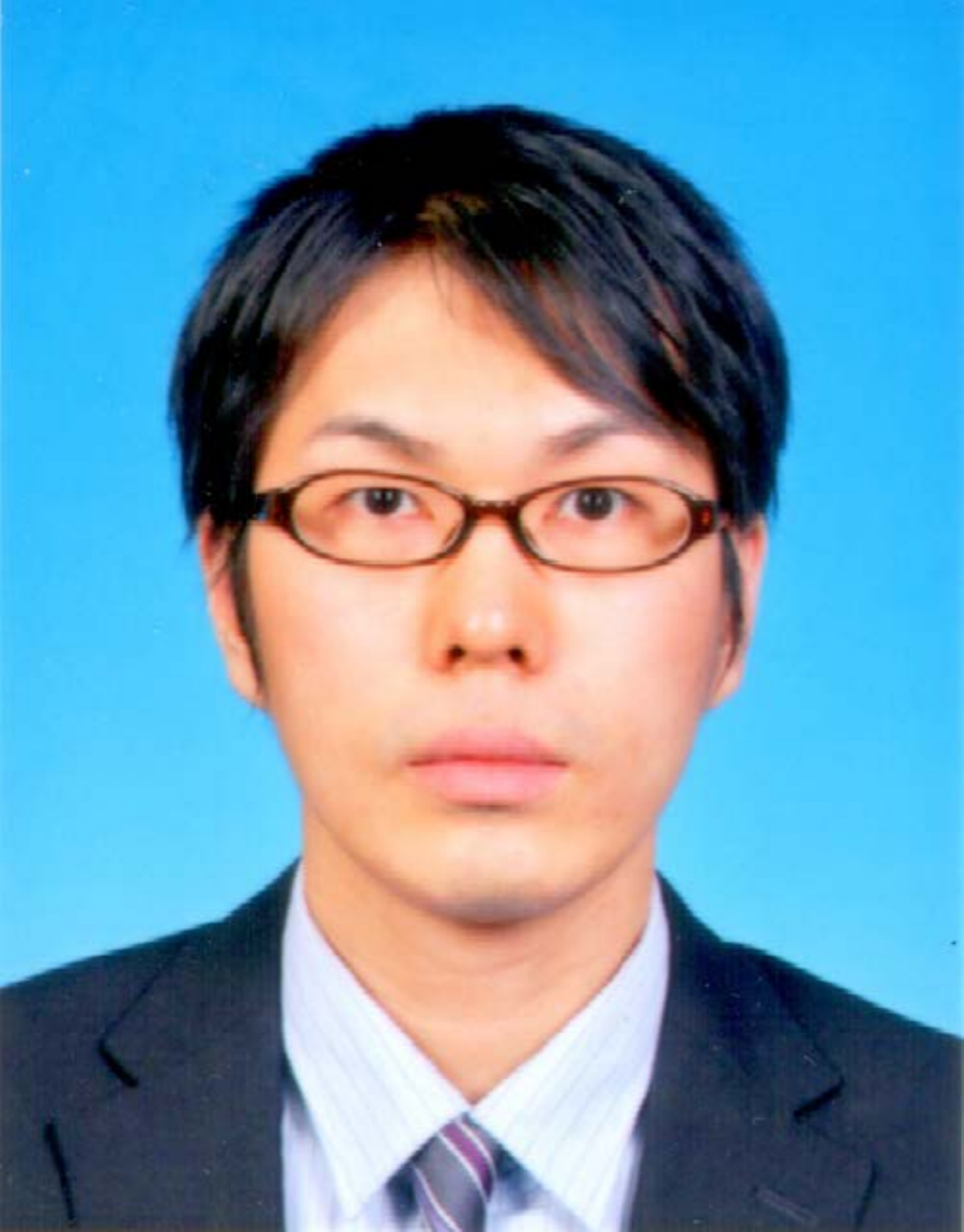}}]{Naoki Masuyama}
	(S'12--M'16) received the B.Eng. degree from Nihon University, Funabashi, Japan, in 2010, the M.E. degree from Tokyo Metropolitan University, Hino, Japan in 2012, and the Ph.D. degree from the Faculty of Computer Science and Information Technology, University of Malaya, Kuala Lumpur, Malaysia, in 2016. 
	
	He is currently an Assistant Professor with the Department of Computer Science and Intelligent Systems, Osaka Prefecture University, Sakai, Japan. 
	
	His current research interests include clustering, data mining, and continual learning.
\end{IEEEbiography}
\vspace{-5mm}
\begin{IEEEbiography}[{\includegraphics[width=1in,height=1.25in,clip,keepaspectratio]{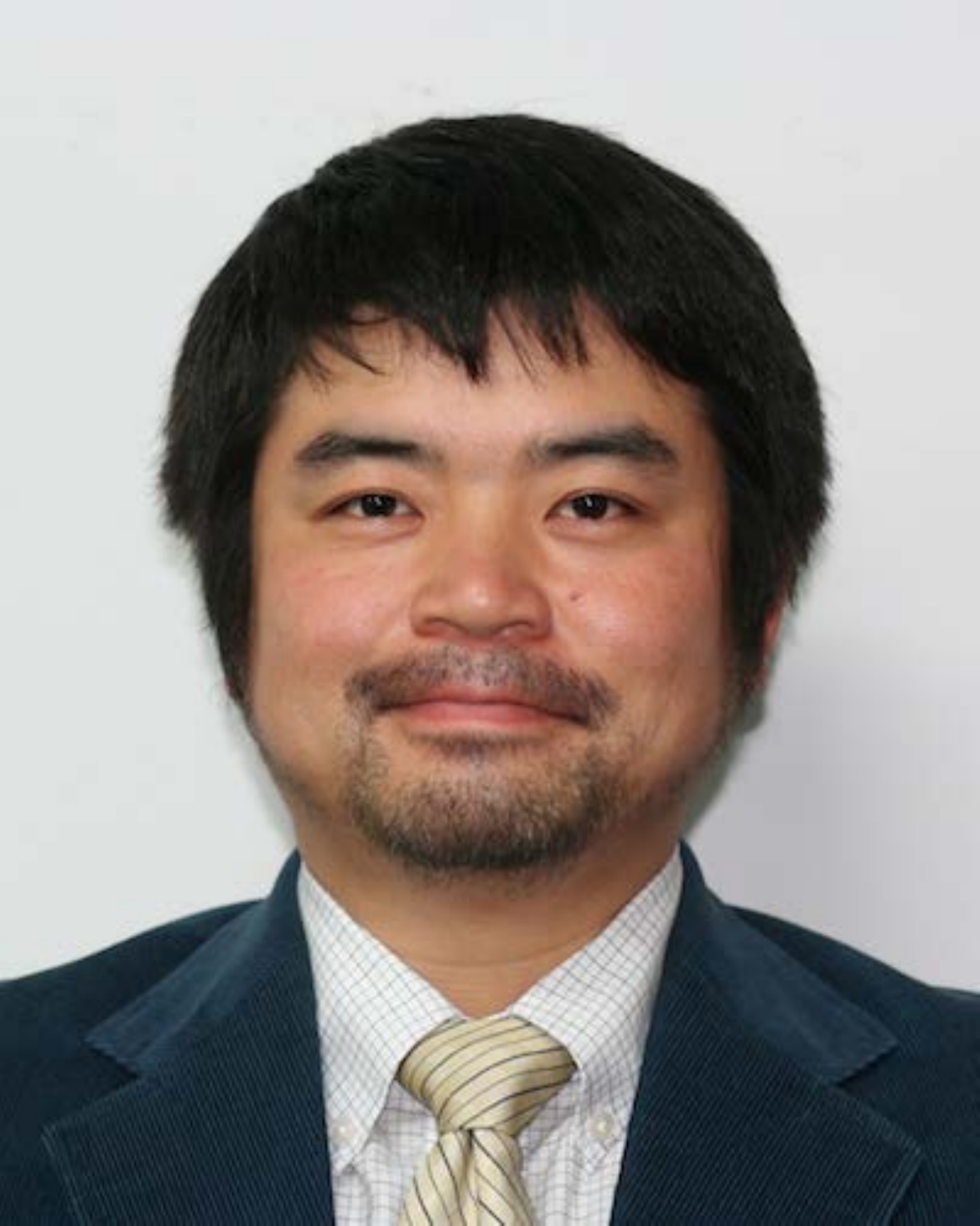}}]{Yusuke Nojima}
	received the B.S. and M.S. Degrees in mechanical engineering from Osaka Institute of Technology, Osaka, Japan, in 1999 and 2001, respectively, and the Ph.D. degree in system function science from Kobe University, Hyogo, Japan, in 2004.
	
	Since 2004, he has been with Osaka Prefecture University, Osaka, Japan, where he is currently a Professor in Department of Computer Science and Intelligent Systems. 
	
	His research interests include evolutionary fuzzy systems, evolutionary multiobjective optimization, and parallel distributed data mining. He was a guest editor for several special issues in international journals. He was a task force chair on Evolutionary Fuzzy Systems in Fuzzy Systems Technical Committee of IEEE Computational Intelligence Society. He was an associate editor of IEEE Computational Intelligence Magazine (2014-2019).
\end{IEEEbiography}
\vspace{-5mm}
\begin{IEEEbiography}[{\includegraphics[width=1in,height=1.25in,clip,keepaspectratio]{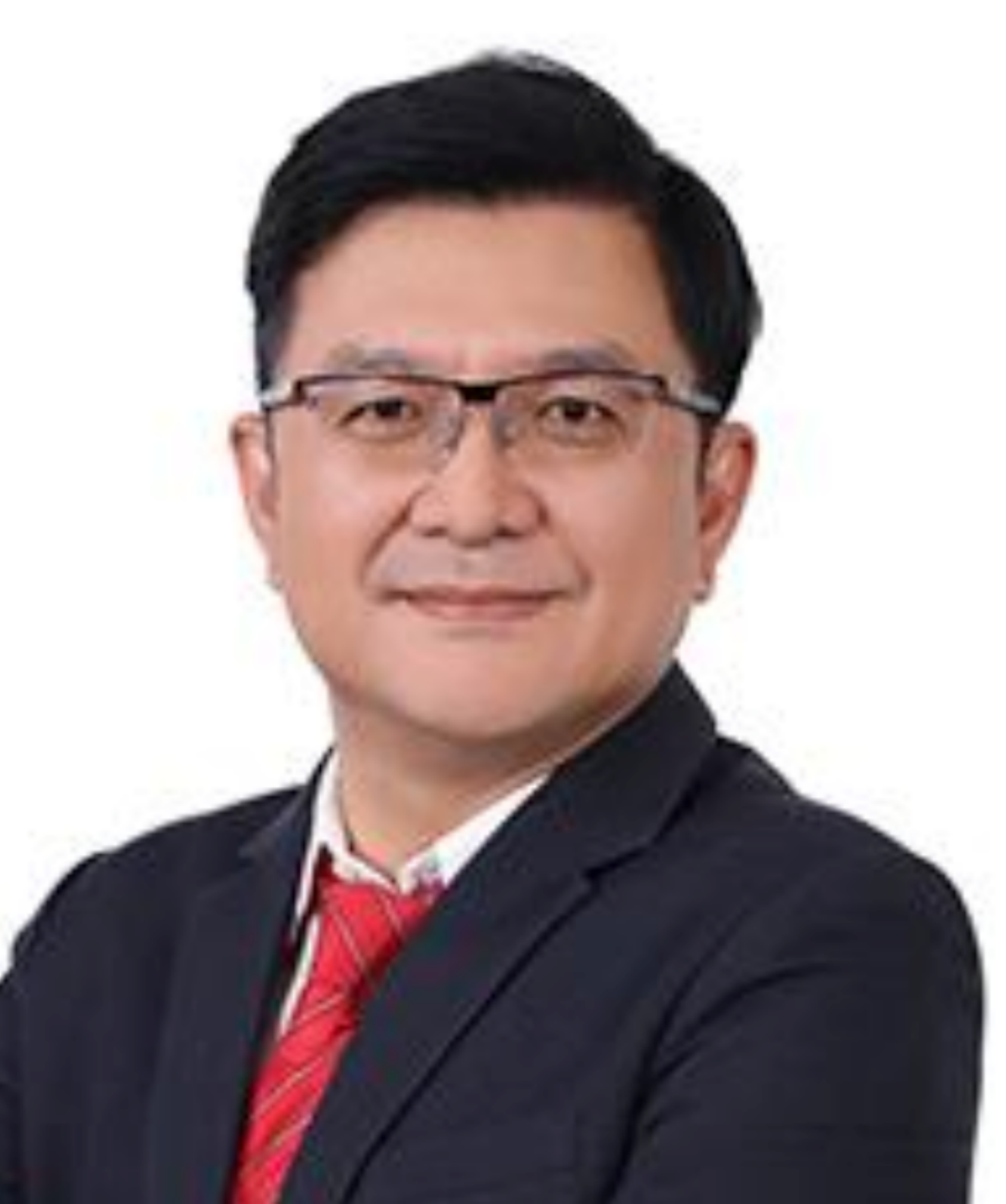}}]{Chu Kiong Loo}
	(SM'14)  holds a Ph.D. (University Sains Malaysia) and B.Eng. (First Class Hons in Mechanical Engineering from the University of Malaya).
	
	He was a Design Engineer in various industrial firms and is the founder of the Advanced Robotics Lab. at the University of Malaya. He has been involved in the application of research into Perus's Quantum Associative Model and Pribram's Holonomic Brain Model in humanoid vision projects. Currently he is Professor of Computer Science and Information Technology at the University of Malaya, Malaysia. He has led many projects funded by the Ministry of Science in Malaysia and the High Impact Research Grant from the Ministry of Higher Education, Malaysia. Loo's research experience includes brain-inspired quantum neural networks, constructivism-inspired neural networks, synergetic neural networks and humanoid research.
\end{IEEEbiography}
\vspace{-5mm}
\begin{IEEEbiography}[{\includegraphics[width=1in,height=1.25in,clip,keepaspectratio]{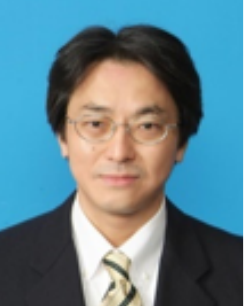}}]{Hisao Ishibuchi}
	(M'93--SM'10--F'14) received the B.S. and M.S. degrees in precision mechanics from Kyoto University, Kyoto, Japan, in 1985 and 1987, respectively, and the Ph.D. degree in computer science from Osaka Prefecture University, Sakai, Osaka, Japan, in 1992.
	
	Since 1987, he has been with Osaka Prefecture University for 30 years. He is currently a Chair Professor with the Department of Computer Science and Engineering, Southern University of Science Technology, Shenzhen, China. His current research interests include fuzzy rule-based classifiers, evolutionary multiobjective optimization, many-objective optimization, and memetic algorithms. 
	
	Dr. Ishibuchi was the IEEE Computational Intelligence Society (CIS) VicePresident for Technical Activities from 2010 to 2013. He was an IEEE CIS AdCom Member from 2014 to 2019, an IEEE CIS Distinguished Lecturer from 2015 to 2017, and an Editor-in-Chief of the IEEE Computational Intelligence Magazine from 2014 to 2019. He is also an Associate Editor of the IEEE TRANSACTIONS ON EVOLUTIONARY COMPUTATION, the IEEE TRANSACTIONS ON CYBERNETICS, and the IEEE ACCESS.
\end{IEEEbiography}

%
%
%




\end{document}